\documentclass{article}

\usepackage{amsmath,amsfonts,amssymb,amsthm,dsfont}
\usepackage{xspace}
\usepackage{fullpage}
\usepackage{graphicx}
\usepackage{subfigure}
\usepackage{epstopdf}
\usepackage[noend]{algpseudocode}
\usepackage{algorithmicx}
\usepackage{algorithm}
\usepackage{float}
\usepackage[space]{grffile}

\newcommand{\ie}{\emph{i.e.}, }
\newcommand{\eg}{\emph{e.g.}, }
\newcommand{\bs}{\mathbf}
\newcommand{\Rbb}{\mathbb{R}}
\newcommand{\bb}{\mathbb}
\newcommand{\cl}{\mathcal}

\newcommand{\EE}{\mathbb{E}}
\newcommand{\VV}{\mathbb{V}}

  \title{Non-parametric PSF estimation from celestial transit solar images using blind deconvolution}
  \author{%
    A. Gonz\'{a}lez\footnote{%
      Institute of Information and
      Communication Technologies, Electronics and Applied Mathematics
      (ICTEAM), Universit\'{e} catholique de Louvain (UCL), B-1348
      Louvain-la-Neuve, Belgium. \url{adriana.gonzalez@uclouvain.be},
      \url{laurent.jacques@uclouvain.be}},\ \ 
    V. Delouille\footnote{%
      Royal Observatory of Belgium, Avenue Circulaire 3, 1180
      Bruxelles, Belgium. \url{v.delouille@oma.be}}\ \ and L. Jacques$^*$}
  \date{\today}

\begin{document}
\maketitle

\begin{abstract}
\noindent\emph{Context:} Characterization of instrumental effects in astronomical imaging is important in order to extract accurate physical information from the observations. The measured image in a real optical instrument is usually represented by the convolution of an ideal image with a Point Spread Function (PSF). Additionally, the image acquisition process is also contaminated by other sources of noise (read-out, photon-counting). The problem of estimating both the PSF and a denoised image is called blind deconvolution and is ill-posed.
\medskip

\noindent\emph{Aims:} We propose a blind deconvolution scheme that relies on image regularization. Contrarily to most methods presented in the literature, our method does not assume a parametric model of the PSF and can thus be applied to any telescope.
\medskip

\noindent\emph{Methods:} Our scheme uses a wavelet analysis prior model on the image and weak assumptions on the PSF. We use observations from a celestial transit, where the occulting body can be assumed to be a black disk. These constraints allow us to retain meaningful solutions for the filter and the image, eliminating trivial, translated and interchanged solutions. Under an additive Gaussian noise assumption, they also enforce noise canceling and avoid reconstruction artifacts by promoting the whiteness of the residual between the blurred observations and the cleaned data. 
\medskip

\noindent\emph{Results:} Our method is applied to synthetic and experimental data. The PSF is estimated for the SECCHI/EUVI instrument using the 2007 Lunar transit, and for SDO/AIA using the 2012 Venus transit. Results show that the proposed non-parametric blind deconvolution method is able to estimate the core of the PSF with a similar quality to parametric methods proposed in the literature. We also show that, if these parametric estimations are incorporated in the acquisition model, the resulting PSF outperforms both the parametric and non-parametric methods.
\end{abstract}

\section{Introduction}
\label{sec:intro}

Deconvolution is an ubiquitous data processing method that arises in a variety of applications, \eg biomedical imaging~\cite{Jefferies:02}, astronomy~\cite{2012A&A...539A.133P,Shearerphdthesis}, remote sensing~\cite{DFXL2011} and video and photo enhancement~\cite{Fergus:2006}.
Formally, this problem amounts to recovering a signal $\bs x$ from \emph{blurred} and \emph{corrupted} observations $\bs y$:
\begin{equation}
\label{eq:blur-model}
\bs y = \Theta(\bs h \otimes \bs x),    
\end{equation}
where $\otimes$ stands for the \emph{convolution} operation between $\bs x$ and some other signal $\bs h$, while $\Theta$ is a general function accounting for some corrupting noise, \eg $\Theta(\bs u) := \bs u + \bs n$ under an additive corruption model with some (Gaussian) noise $\bs n$.

For instance, when a scene is captured by an optical instrument, the observation is a blurred or degraded version of the original image, corrupted by noise and by some effects due to the instrument (motion, out-of-focus, light scattering, ...). In such a case, the imaging system is usually assumed to be well represented by the sensing model (\ref{eq:blur-model}) where the ideal image $\bs x$ is blurred by a Point Spread Function (PSF) $\bs h$. Implicitly, the PSF describes the response of the imaging device to a Dirac point source, \ie the impulse response of the instrument, while $\Theta$ can distinguish different sources of noise contaminating the image: read-out, photon counting, multiplicative and compression noise, among others.

If the true PSF can be determined in advance, then it is possible to recover the original, undistorted image by convolving the acquired image with the inverse PSF. This is the so called \lq known-PSF deconvolution.' In most practical applications, however, finding the true PSF is impossible and an approximation must be made. As the acquired image is corrupted by various sources of noise, both the PSF and a denoised version of the image should be recovered together. This process is known as \lq blind deconvolution.'

Since there are infinite combinations of image and filter that are compatible with the distorted observations, the blind deconvolution problem is severely ill-posed. One way to reduce the number of unknowns is to introduce a forward (parametric) model of the PSF. For instance, Oliveira et al.~\cite{OFB2007} have modeled the motion blur by a straight line, where the number of unknowns is reduced to two: the line length and angle. Babacan et al.~\cite{4689325} have modeled smoothly varying PSFs by using a simultaneous autoregressive prior, where the only parameter to estimate is the variance of a Gaussian function. However, such methods are limited to specific applications as they can only recover the expected model of the PSF, excluding the estimation of a generic non-parametric PSF. Furthermore, due to the ill-posedness of the blind deconvolution problem, a slight mismatch between the specified model and the true PSF can lead to poorly deconvolved images.

Blind-deconvolution is also very sensitive to the noise present in the observations. While some works~\cite{Ayers:88, 2006SoPh..239..531G} have used fast and efficient techniques such as a simple inverse filtering to recover both the PSF and the undistorted image from the blurred observations, these methods present low tolerance to noise~\cite{Fish:95}. 

Robust blind deconvolution methods, on the other hand, optimize a data fidelity term, which is based on the acquisition model, stabilized by some additional regularization terms. If the observations are corrupted by an additive Gaussian noise, the deconvolution problem can be solved by a regularized least-squares (LS) minimization~\cite{5226594}.  In the presence of Poisson noise, the problem can be formulated using the Kullback-Leibler (KL) divergence as the data fidelity term~\cite{Fish:95, 2012A&A...539A.133P,0266-5611-29-6-065017}, which represents a more complex function to minimize. To avoid dealing with the KL divergence, the Poisson corrupted data can also be handled through a Variance Stabilization Transform (VST)~\cite{4738431,2012ApJ...749L...8S, Shearerphdthesis}. The VST provides an approximated Gaussian noise distributed data and thus allows working with a regularized LS formulation. 

In this work, blind deconvolution is used to recover both the PSF and the undistorted image from blurred observations acquired by solar extreme ultraviolet (EUV) telescopes. We use  
the proximal alternating minimization method recently proposed by Attouch et al.~\cite{Attouch:2010:PAM:1836121.1836131} in the context of additive Gaussian noise. This algorithm allows handling LS problems for a large amount of regularization functionals. The advantage of this method over usual alternating minimization approaches, \eg Fish et al.~\cite{Fish:95, 5226594}, is that it provides theoretical convergence guarantees and it is general enough to include a wide variety of prior information.
To our knowledge, it is the first time that blind deconvolution is solved using the proximal algorithm of Attouch et al.~\cite{Attouch:2010:PAM:1836121.1836131}.

In optical telescopes, the PSF originates from various instrumental effects, such as: optical aberrations (spherical, astigmatism, coma), diffraction (produced by, \eg an entrance filter mesh), scattering (from, \eg micro-roughness of the mirrors resulting in long-range diffuse illumination), charge spreading, etc. All these effects `spread' light, \ie incident photons which would otherwise focus to a single point of the focal place, may get detected at a location that is a bit shifted or even far away. This does affect different types of measurements, such as the photometry of fine features, see, \eg~\cite{2009ApJ...690.1264D}.

The PSF of solar telescopes is usually modeled using pre-flight instrument specifications~\cite{1995SoPh..157..141M, aia_psf}. After building a specific model for a given instrument, few parameters need to be estimated in order to recover the PSF. For instance, Gburek et al.~\cite{2006SoPh..239..531G} fitted the central pixels of the TRACE PSF to a Moffat function, while DeForest et al.~\cite{2009ApJ...690.1264D} modeled the long-range effect of the TRACE PSF as the sum of a measured diffraction pattern with a circularly symmetric scattering profile. The PSFs from the four EUVI instrument channels on STEREO-B were studied by~Shearer et al.~\cite{2012ApJ...749L...8S,Shearerphdthesis}, where the long-range scattering effect was assumed to follow a parametric piece-wise power-law model. A similar method was used to estimate the PSF of the SWAP instrument on board the PROBA2 satellite~\cite{2013ApJ...777...72S}. Finally, Poduval et al.~\cite{2013ApJ...765..144P} provided a semi-empiric model for the Atmospheric Imaging Assembly (AIA) onboard SDO, similar to the one used by DeForest et al.~\cite{2009ApJ...690.1264D} for TRACE. In all these works, the complete set of parameters that were fitted are in the range of five to eleven values, just a few compared to the resolution of the PSF (millions of pixels).

\begin{figure}[ht]
  \centering
	\includegraphics[width=0.7\textwidth]{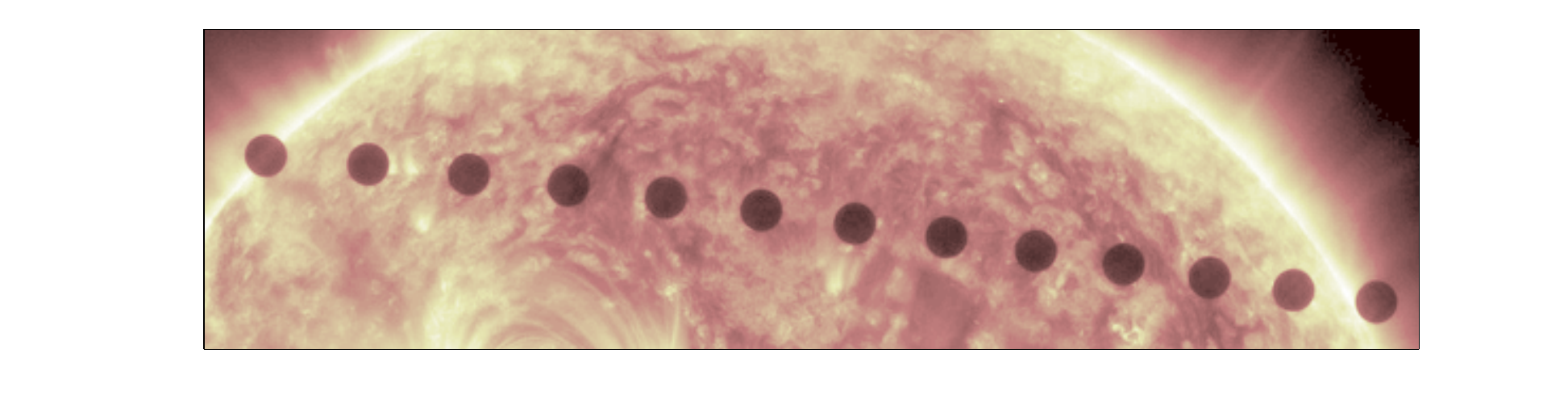}
  \caption{Accumulation of observations of the Venus transit captured by SDO/AIA on June 5th - 6th~2012}
 \label{fig:Transit}
\end{figure}

In addition to building a parametric model of the PSF using the specifications of the instruments, some works~\cite{2009ApJ...690.1264D,2012ApJ...749L...8S, 2013ApJ...765..144P} have used the information from a celestial transit to help inferring the PSF of a given instrument. A celestial transit, shown in Figure~\ref{fig:Transit}, is an astronomical event where the Moon or a planet move across the disk of the sun, hiding a part of it, as seen from the telescope. Since the EUV emission of transiting celestial bodies is zero, any apparent emission is known to be caused by the instrument. Therefore, this type of event provides strong prior information that allows the regularization of the blind deconvolution problem.

Let us note that all these parametric methods present the disadvantage of being specific for a given instrument and depending on a good characterization of the PSF, which is not always possible~\cite{2014SoPh..289.1043M}.

\subsection{Contribution}

Central to our work is the information provided by the transit of a celestial object whose apparent boundary on the recorded image can be predicted with high (sub-pixel) accuracy. This is typically the case for the observation of Moon or Venus transits for which ephemeris allows us to precisely know their apparent boundaries at the recording time, provided of course that \emph{(i)} the small variations of the object topography around a perfect disk (\eg mountains, craters) are smaller than a pixel width, and \emph{(ii)}, that such a transitting object has no atmosphere that could blur its apparent boundary (\eg as for an Earth transit). As will be clear below, these hypotheses have been respected for at least two cases: for the instrument SECCHI/EUVI during the 2007 Lunar transit and for SDO/AIA during the 2012 Venus transit. For these two situations, and for any other transit respecting the conditions above, we know a priori the values for a set of pixels in the image and also their exact location. Our work uses such information as constraint in the function to be optimized for blindly deconvolving images. Notice that our approach could also be applied to other situations where pixel values and exact location are known a priori (\eg dark calibration patterns in Computer Vision applications).

Contrarily to previous works in solar physics, the blind deconvolution method demonstrated in this paper is not based on a specific model of the PSF, but infers it from the observed data. This allows the method to be used for estimating the PSF of any instrument provided it has a prior information on a set of pixel values. Since no model is imposed on the PSF, the large number of unknowns makes the problem computationally intractable. We thus focus on the estimation of the PSF core, which is defined as the central pixels encompassing at least $99\%$ of the total PSF energy. Note that this thresholding follows the radial energy characterization of the SDO/AIA PSF in the paper of Poduval et al.~\cite{2013ApJ...765..144P} and a similar study of the available PSFs of SECCHI/EUVI (\eg~\cite{2012ApJ...749L...8S}). The PSF core accounts for charge spreading, optical aberrations, and diffraction effects. Note that the same algorithm would be able to handle a parametric model of the PSF provided a precise instrument characterization is available. For simplicity, we consider an average additive Gaussian noise model that gathers all sources of noise present in the observation. A detailed discussion on how to handle more general noise models is provided in Section~\ref{sec:discussion}.

The proposed method is first tested through simulated realistic scenarios. Results show the ability to estimate a given PSF with high quality. We also show the importance of considering multiple transit observations in order to provide a better conditioning to the filter estimation problem.

We validate our method on AIA and EUVI observations using solar transits. The validation of the recovered filters is based on the reduction of the celestial body apparent emissions. The estimated PSFs were also tested for images containing active regions. Results show that, when recovering the PSF core, the proposed non-parametric scheme can achieve similar results to parametric methods. Also, we consider the case where the parametric PSF is incorporated in the imaging model as an additional filter convolving the actual image~\cite{2012ApJ...749L...8S,Shearerphdthesis}. We show that the resulting parametric/non-parametric PSF provides better results than both parametric and non-parametric PSFs estimated independently.
Let us note that, since parametric methods are based on the physics of the instrument, they tend to provide a more precise PSF containing both the core and the diffraction peaks. However, when the telescope presents unknown or unexpected behavior, an accurate model of the filter cannot be built. In such cases, we require non-parametric models, as the one presented in this paper, which due to their flexibility can handle better some instrument's properties.

\subsection{Outline}

In Section~\ref{sec:Model} we describe the discrete forward model and the noise estimation. In Section~\ref{sec:BID} we present the formulation of the blind deconvolution problem based on the image and filter prior information. Section~\ref{sec:AM} describes the alternating minimization method used for estimating the filter and the image. It briefly presents the method, including the initialization, the parameter estimation, the stopping criteria and the numerical reconstruction. In Section~\ref{sec:NBID} we present a non-blind deconvolution method that is used for the experimental validation. In Section~\ref{sec:Results} we present the results on both synthetic and experimental data for the AIA and EUVI telescopes. Finally, Section~\ref{sec:discussion} presents a discussion of the obtained results, providing some perspectives for future research work.

\subsection{Notations}

We denote by $\bb N$, $\bb Z$ and $\bb R$ the sets of natural, integer and real numbers, respectively. The set of the non-negative real numbers is denoted by $\bb R_+$. Most of domain dimensions are denoted by capital roman letters, \eg $M, N, \ldots$ Vectors and matrices are associated with bold symbols, \eg $\bs \Phi \in \Rbb^{M\times N}$ or $\bs u \in \Rbb^M$, while lowercase light letters are associated with scalar values. The $i^{\rm th}$ component of a vector $\bs u$ reads either $u_i$ or $(\bs u)_i$, while the vector $\bs u_i$ may refer to the $i^{\rm th}$ element of a vector set. The vector of ones in $\Rbb^D$ is denoted by $\bs 1_D = (1, \cdots, 1)^\mathtt{T}$. The set of indices in $\Rbb^D$ is $[D] = \{1, \cdots, D\}$. The support of a vector $\bs u \in \Rbb^D$ is defined as $\textrm{supp} \ \bs u = \{ i \in [D] : u_i \neq 0 \}$. The cardinality of a set $\cl C$, measuring the number of elements of the set, is denoted by $|\cl C|$. The convolution between two vectors $\bs u,\bs v \in \Rbb^D$ for some dimension $D \in \bb N$ is denoted equivalently by $\bs u \otimes \bs v = \bs v \otimes \bs u$. We denote $\Gamma^0(\cl V)$ the class of proper, convex and lower-semicontinuous functions from a finite dimensional vector space $\cl V$ (\eg $\bb R^D$) to $(-\infty,+\infty]$~\cite{CombettesPesquet2011}. The (convex) indicator function $\imath_{\cl C}(\bs x)$ of the set $\cl C$ is equal to $0$ if $\bs x\in \cl C$ and $+\infty$ otherwise. The 2-D discrete delta function, denoted as $\bs \delta_0(m,n)$, is equal to 1 if $m = n = 0$, and $0$ otherwise. The transposition of a matrix $\bs \Phi$ reads $\bs \Phi^\mathtt{T}$. For any $p\geq 1$, the $\ell_p$-norm of $\bs u$ is $\|\bs u\|_p = \left(\sum_i |u_i|^p\right)^{1/p}$. The Frobenius norm of $\bs \Phi$ is given by $\|\bs \Phi\|_F^2 = \sum_i \sum_j |\phi_{ij}|^2$. The Gaussian distribution of mean $0$ and variance $\sigma^2$ is denoted by $\cl N(0,\sigma^2)$.

\section{Problem statement}
\label{sec:Model}

The telescope imaging process can be mathematically modeled as the instrument's PSF $\bs h$ convolving the true image $\bs x$, \ie $\bs h \otimes \bs x$. The convolution operation, represented by $\otimes$, consists of integrating portions of the actual scene weighted by the PSF. In the following, we consider a discrete setting where the convolution integration is represented by a discrete sum.

\subsection{Discrete model}

Let us consider that, during the transit, the EUV telescope acquires a set of $P$ images containing a celestial body. In order to limit the computation time, the effective Field-of-View (FoV) of each observation is restricted to an image patch centered on the transiting celestial body with a size several times bigger than the body apparent diameter. The effects of this FoV truncation will be discussed in detail in Section~\ref{sec:BID} and Section~\ref{sec:Results}. In our work, all $n \times n$ images are represented as $N = n^2$ vectors so that linear transformations of images are identifiable with matrices. Each observed patch is a collection of values gathered in a vector $\bs y_j \in \Rbb^N$, with $j=1,\cdots,P$. The observed values are modeled as a \lq true' image $\bs x_j \in \Rbb^N$ convolved by the instrument's PSF ($\bs h \in \Rbb^N$). The PSF is assumed to be spatially and temporally invariant, thus is the same for each observed patch. For simplicity, we consider that all sources of noise present in the observation can be gathered in an average noise model, which is assumed to be additive, white and Gaussian. The observations are then assumed to be affected by an Additive White Gaussian Noise (AWGN), $\bs n_j \in \Rbb^{N}$, with $(\bs n_j)_i \sim_{\rm iid} \cl N(0,\sigma^2)$. The acquisition process of the telescope is thus modeled as follows:
\begin{equation}
\label{eq:discrete_model_vector}
	\bs y_j = \bs h \otimes \bs x_j + \bs n_j.
\end{equation}
This model assumes that the observed images $\bs y_j$ are uncompressed and have been corrected for CCD effects (dark current removal, flat fielding, despiking). In the case of SDO/AIA, this corresponds to level 1 data processing.

In 1-D, the discrete circular convolution $\sum_{k=1}^n u_k g_{i-k+1}$ of a vector $\bs u \in \bb R^n$ with a filter $\bs g \in \bb R^n$ can always be described by the product of $\bs u$ with a circulant matrix 
\begin{equation}
\bs \Phi(\bs g) 
\begin{pmatrix}
g_1&g_2&\cdots&g_n\\
g_n&g_1&\cdots&g_{n-1}\\
\vdots&&&\vdots\\
g_2&g_3&\cdots&g_1  
\end{pmatrix}
\in \Rbb^{n \times n},
\end{equation}
\ie a matrix where every row is a right cyclic shift of the \emph{kernel} $\bs g$~\cite{CIT-006}. 
In 2-D and in particular for the model (\ref{eq:discrete_model_vector}), the convolution of $\bs x_j$ by $\bs h$ can still be represented by the multiplication of $\bs x_j$ by matrix $\bs \Phi(\bs h) \in \bb R^{N \times N}$ that is now \emph{block-circulant with circulant blocks}, \ie a matrix made of $n \times n$ blocks, each block being a $n\times n$ circulant matrix representing the action of one row of the 2-D filter $\bs h$ in~(\ref{eq:discrete_model_vector}).

If we gather the $P$ \lq true' images in a single matrix $\bs X = (\bs x_1, \cdots, \bs x_P) \in \Rbb^{N \times P}$, the acquisition model can be transformed into the following matrix form:
\begin{equation}
\label{eq:discrete_model_matrix}
	\bs Y = \bs \Phi(\bs h) \ \bs X + \bs N,
\end{equation}
with $\bs Y, \bs N \in \Rbb^{N \times P}$ two matrices that similarly gather the $P$ observed images and their noises, respectively.

\subsection{Noise estimation}
\label{sec:noise_estim}

For the sake of simplicity and to keep fast numerical methods, the model~(\ref{eq:discrete_model_matrix}) implicitly considers an additive noise model. A way to generalize our method to Poisson noise would be to include a VST~\cite{ansc,4738431,2012ApJ...749L...8S} in the acquisition model in (\ref{eq:discrete_model_matrix}), both on the observations $\bs Y$ and on the convolution result $\bs \Phi(\bs h) \ \bs X$ (see Section~\ref{sec:discussion} for more details).

Under the AWGN assumption, the energy of the residual noise is known to be bounded using the Chernoff-Hoeffding bound~\cite{doi:10.1080.01621459.1963.10500830}:
\begin{equation}
\label{eq:epsilon}
\| \ \bs Y \ - \ \bs \Phi(\bs h) \ \bs X \ \|_F^2 = \| \ \bs N \ \|_F^2 < \varepsilon^2 := \sigma^2 \left(NP + c\sqrt{NP}\right),
\end{equation}
with high probability for $c = \cl O (1)$. A first guess of the noise variance $\sigma^2$ can be estimated using the Robust Median Estimator ($\sigma_{RME}^2$)~\cite{donoho01091994}, which is based on the assumption that the noise is an additional high frequency component in the observed signal. More details on this variance estimation are provided in Section~\ref{sec:ExpData}.

Let us note that the variance could also be obtained from the physical noise characteristics using, for instance, dark areas in the images to estimate the dark noise. However, such computations may not consider all the noise sources present in the observations, providing an underestimation of the actual variance. 

In this work, we prefer to adopt another strategy. Since~(\ref{eq:epsilon}) considers an average noise model and since~(\ref{eq:discrete_model_matrix}) is an AWGN approximation of the actual data noise corruption, we aim to find an adaptive value of $\sigma$ that optimizes the AWGN assumption, \ie which minimizes somehow the proximity of this simple model to the true corruptions. Section~\ref{sec:ExpData} describes in detail this adaptive strategy and demonstrates its advantage over the fixed choice $\sigma = \sigma_{RME}$.

\section{Blind deconvolution problem}
\label{sec:BID}

In this section, we aim at reconstructing both $\bs X$ and $\bs h$ from the noisy observations $\bs Y$. Since the data is corrupted by an AWGN, we can estimate the image and the filter using a least squares minimization, \ie by finding the values that minimize the energy of the noise:
\begin{equation}
\label{eq:LS}
	\textrm{min}_{\tilde{\bs X}, \tilde{\bs h}} \ \textstyle\frac{1}{2} \| \ \bs Y - \bs \Phi (\tilde{\bs h}) \ \tilde{\bs X} \ \|^2_F, 
\end{equation}
with $\tilde{\bs X} \in \Rbb^{N \times P}$ and $\tilde{\bs h} \in \Rbb^{N}$.

The blind deconvolution problem in (\ref{eq:LS}) is ill-posed and can have infinite possible solutions. By regularizing this problem we aim to reduce the amount of possible solutions to those that are meaningful.
In the coming sections, we first describe the prior information and constraints on the image and the filter and then use this information to formulate a regularized blind deconvolution problem.

\subsection{Prior information on the image}
\label{sec:ImagePrior}
In the following, we present in detail the available prior information on the image candidate.

\paragraph*{\bf (a) Zero disk intensity:} Each image of the transit contains the observation of a celestial body, Venus or the Moon. For each observation $\bs y_j$, we know that the celestial bodies' EUV emission is zero and therefore can be represented as a black disk of constant radius $R$ and center $c_j$, both being known from external astronomical observations:~\cite{2012SoPh..275....3P} for SDO/AIA and (J.-P. Wuelser \emph{private communication}) for SECCHI/EUVI. The set of pixels of $\bs x_j$ inside the disk is denoted by $\Omega_j$. In the minimization problem we can set to zero the pixels of the $j^{\rm th}$ image that are inside the set $\{\Omega_j\}$, \ie $X_{ij} = (\bs x_j)_i = 0$ for all $i \in \Omega_j$. This prior information is crucial in the regularization of the blind deconvolution problem as it allows us to remove the issue of interchangeability between the filter and the image. Furthermore, it prevents ``oppositely'' translated solutions of the image and the filter, a problem occurring because the convolution process is blind to such translations, \ie $\bs h \otimes \bs x_j = \cl T_{\bs a} \cl T_{- \bs a} (\bs h \otimes \bs x_j) = (\cl T_{\bs a} \bs h) \otimes (\cl T_{- \bs a}\bs x_j)$, with $\cl T_{\bs a}$ a translation by $\bs a \in \bb Z^2$.

\paragraph*{\bf (b) Analysis-based sparsity model:} As commonly done with ill-posed inverse problems associated to image restoration tasks~\cite{1994A&A...288..342S,382009,5414556}, we regularize our method with a 2-D wavelet prior. Compared to a frequency representation, as obtained by the Fourier transform, a 2-D wavelet transform allows to represent images with a \emph{multi-resolution} scheme~\cite{Mallat:2008:WTS:1525499}, \ie with a linear combination of localized \emph{wave atoms}, with variable sizes and locations, whose number is much smaller than the initial number of pixels. Most inverse problems in imaging are regularized using a synthesis wavelet prior, \ie promoting the synthesis of the estimated image with as few `wavelets' as possible (sparsity criterion)~\cite{1994A&A...288..342S,382009}. More recently, better inversion results have been obtained using \emph{analysis}-based sparsity models~\cite{Carrillo21102012,Sudhakar2014}, which rather promote sparse \emph{projections} of the estimated image over a redundant system of wave-functions, \ie a \emph{dictionary}. This set can be much larger than an orthonormal basis and hence it can efficiently capture much more different image features.

Following this recent trend, we decide to adopt such an analysis prior that, in opposition to the synthesis framework, also allows us to work directly on the image domain without increasing the dimensionality of the optimization problem. As a dictionary, we use the system associated to the Undecimated Discrete Wavelet Transform (UDWT)~\cite{Starck:2010:SIS:1830428}. The UDWT can also be seen as the union of all translations of an orthonormal DWT. Conversely to the DWT, this makes the UDWT translation invariant, \ie it enables an efficient characterization of all image features whatever their location~\cite{49062880,5420029}, hence providing a better image reconstruction than the traditional DWT.

In this context, we assume that, if we represent the image candidate on a redundant wavelet dictionary $\bs \Psi \in \Rbb^{N \times W}$ made of $W$ vectors in $\Rbb^N$, the wavelet coefficients are sparse, \ie the coefficients matrix $\bs \Psi^\mathtt{T} \bs X$ has few important values and its $\ell_1$-norm (computed over all its entries) is expected to be small. However, the wavelet coefficients whose support touches the boundary of the Moon or Venus disk are not sparse. Hence, we do not consider in this prior model those coefficients that are affected by the occulting body. Also, we follow a common practice in the field which removes the (unsparse) scaling coefficients~\cite{Mallat:2008:WTS:1525499} from the $\ell_1$-norm computation. 
The set of \emph{detail} coefficients that are not affected by the black disk is denoted by $\Theta$, with $|\Theta|=Q$. See Section~\ref{sec:SynthData} for further details on how to compute the set $\Theta$. The matrix $\bs S_\Theta \in \Rbb^{Q \times W}$ is the corresponding selection operator that extracts from any coefficients vector $\bs u \in \Rbb^W$ only its components indexed by $\Theta$. The global prior information can thus be exploited by promoting a small $\ell_1$-norm on the wavelet coefficients belonging to $\Theta$. The rationale of this prior is to enforce noise canceling and avoid artifacts in the reconstructed image.

\paragraph*{\bf (c) Image non-negativity:} Note that the image corresponds to a measure of a photon emission process, therefore, its pixel values must be non-negative everywhere.

\subsection{Constraints on the PSF}
\label{sec:FilterPrior}

Since the filter is not known a priori and it changes from one instrument to another, we are interested in adding soft constraints that are common to most solar EUV telescopes. We are not aiming at reconstructing all of the PSF's components but rather at estimating its core, which is responsible for most of the diffused light.

We first consider that, since the PSF corresponds to an observation of a point, it is non-negative everywhere. Additionally, by assuming that the amount of light entering in the instrument is preserved, the $\ell_1$-norm of the filter candidate must be equal to one. This is explained by taking the sum over the pixels of one observed patch in (\ref{eq:discrete_model_matrix}), which is equal to $\bs 1^\mathtt{T} \bs y_j = \bs 1^\mathtt{T} \bs \Phi (\bs h) \bs x_j$ in a noiseless process. Since the filter is non-negative, taking its $\ell_1$-norm equal to one is equivalent to $\bs 1^\mathtt{T} \bs \Phi (\bs h) = \bs 1^\mathtt{T}$, which results in $\bs 1^\mathtt{T} \bs y_j = \bs 1^\mathtt{T} \bs x_j$, \ie the light is preserved. Therefore, the filter candidate must belong to the Probability Simplex \cite{OPT-003}, defined as \linebreak $\cl PS = \{ \bs h : h_i \geq 0, \| \ \bs h \ \|_1 = 1 \}$.

One important assumption in our proposed method is that the filter candidate is of size \linebreak $(2b+1)\times(2b+1)$, for any $b \in \bb N$, and is centered in the spatial origin of the discrete grid. This means that the filter candidate has a limited support inside a set $\Gamma$ of size $(2b+1)^2$, \ie the filter has only important values in the central pixels and is negligible beyond the considered support. This allows us to work with a patch (only a part of the observation) and not with the complete FoV of the instrument in order to reduce the computation time. 

EUV images have a high dynamic range. Hence, due to long-range effects, a given pixel value can be affected by the value of another high intensity pixel located as far as 100 arcsec away for SDO/AIA~\cite{2013ApJ...765..144P} and 1500 arcsec away for SECCHI/EUVI \cite{2012ApJ...749L...8S}. In such cases, despite a PSF that can rapidly decay far from its center, the high intensity pixel can induce a long-range effect that is not taken into account by a filter with truncated support. Let us define the complementary set of $\Gamma$ as $\Gamma^\mathtt{C} = [N] \backslash \Gamma$, with $|\Gamma^\mathtt{C}| = N - (2b + 1)^2$. We assume that the actual PSF is composed by two different filters: the PSF core with support on $\Gamma$, denoted by $\bs h_\Gamma$, which accounts for short-range effects, and another one, denoted $\bs h_{\Gamma^\mathtt{C}}$ with support on $\Gamma^\mathtt{C}$, accounting for long-range effects, \ie $\bs h = \bs h_\Gamma + \bs h_{\Gamma^\mathtt{C}}$. An accurate estimation of the long-range PSF is out of the scope of this work. Its effect inside the disk of the celestial body is modeled as a constant $\mu$, \ie $\left(\bs h_{\Gamma^\mathtt{C}} \otimes \bs x_j \right)_i \approx \mu, \ \forall i \in \Omega_j$ (see Appendix~\ref{sec:long_range_psf} for more details). For a given patch, the acquisition model in (\ref{eq:discrete_model_vector}) can then be approximated as: $ \bs y_j = (\bs h_\Gamma + \bs h_{\Gamma^\mathtt{C}}) \otimes \bs x_j = \bs h_\Gamma \otimes \bs x_j + \mu \bs 1_N.$ In this work, we aim at estimating only the PSF core, which is called $\bs h$ hereafter. The effect of the long-range PSF, denoted by $\mu$, is computed in a preprocessing stage by the average value inside the center of several observed transit disks (see Section \ref{sec:ExpData}). This simple estimation is motivated by the presence of a systematic intensity background at the center of each observed disk transit. This one is quite independent of the Sun activity around such disk and its mean intensity (few tens of DNs) is also far above the estimated noise level (few DNs). In a future work we plan to replace this rough evaluation by a joint estimation of the value of $\mu$ with the filter and the image (see Section~\ref{sec:discussion} for some discussions).

Notice that our method could allow the addition of other convex constraints, \eg if we know that the filter is sparse or that $0 \leq h_i \leq g_i$, for some upper bound $g_i$ on $h_i$ such as a specific power law decay. However, we will not consider this possibility here as we want to stay as agnostic as possible on the properties of the filter to reconstruct.

\subsection{Final Formulation}

From Sections \ref{sec:ImagePrior} and \ref{sec:FilterPrior}, we can formulate the following regularized blind deconvolution problem:
\begin{align}
\label{eq:BID2}
	\textrm{min}_{\tilde{\bs X}, \tilde{\bs h}} & \quad \rho \ \| \ \bs S_\Theta \ \bs \Psi^\mathtt{T} \ \tilde{\bs X} \ \|_1 + \textstyle\frac{1}{2} \| \ \bs Y - \ \bs \Phi (\tilde{\bs h}) \ \tilde{\bs X} - \ \mu \bs 1_N \bs 1_P^\mathtt{T} \ \|^2_F \ \\ \nonumber
		\textrm{s.t.} & \quad (\tilde{\bs x}_j)_i = 0 \ \textrm{if} \ i \in \Omega_j; \ (\tilde{\bs x}_j)_i \geq 0 \ \textrm{otherwise} \quad \quad \quad \ \ \\ \nonumber
		&  \quad \tilde{\bs h} \in \cl PS; \ \textrm{supp} \ \tilde{\bs h} = \Gamma \quad \quad \quad \quad \quad \quad \quad \quad \quad \quad
\end{align}
with $\tilde{\bs X} \in \Rbb^{N \times P}$, $\tilde{\bs h} \in \Rbb^N$ and $\mu \bs 1_N \bs 1_P^\mathtt{T}$ cancels the long-range part of the filter. The regularization parameter, denoted by $\rho$, controls the trade-off between the sparsity of the image projection in a wavelet dictionary and the fidelity to the observations. This essential parameter is estimated in Appendix~\ref{sec:param}. 

It is important to note that the prior information related to the darkness of the occulting body ($(\tilde{\bs x}_j)_i = 0$ if $i \in \Omega_j$), which is crucial in the regularization of the blind deconvolution problem, is taken into account in the first constraint in (\ref{eq:BID2}).

The problem of estimating the image and the filter in (\ref{eq:BID2}) can also be written as
\begin{equation}
\label{eq:BID3}
	\left\{\bs X^*, \bs h^*\right\} = \textrm{argmin}_{\tilde{\bs X}, \tilde{\bs h}} \ \ \cl L\left(\tilde{\bs X}, \tilde{\bs h}\right) \quad \textrm{s.t.} \quad \tilde{\bs X} \in \cl P_0, \ \ \tilde{\bs h} \in \cl D,
\end{equation}
where $\cl L\left(\tilde{\bs X}, \tilde{\bs h}\right) = \rho \ \| \ \bs S_\Theta \ \bs \Psi^\mathtt{T} \ \tilde{\bs X} \ \|_1 + \textstyle\frac{1}{2} \| \ \bs Z - \bs \Phi (\tilde{\bs h}) \ \tilde{\bs X} \ \|^2_F$ is the objective function, with $ \bs Z~=~\bs Y - \mu \bs 1_N \bs 1_P^\mathtt{T}$ the modified observations. The convex sets $\cl P_0$ and $\cl D$ are defined as follows: \linebreak $\cl P_0 = \{ \bs U \in \Rbb_+^{N\times P} \ : \ (\bs u_j)_i = 0 \ \textrm{if} \ i \in \Omega_j \}$; $\cl D = \{ \bs v \in \Rbb_+^{N} \ : \ \| \ \bs v \ \|_1 = 1, \textrm{supp} \ \bs v = \Gamma \}$. 

\section{Proximal Alternating Minimization}
\label{sec:AM}

In this section, we describe the alternating minimization method used for solving (\ref{eq:BID3}) and hence estimating the image and the PSF from the noisy observations. 

The problem defined in (\ref{eq:BID3}) is non-convex with respect to both $\bs X$ and $\bs h$, but it is convex with respect to $\bs X$ (resp. $\bs h$) if $\bs h$ (resp. $\bs X$) is known. Such a problem is usually solved iteratively, by estimating the image and the filter alternately~\cite{5226594}. The general behavior of this type of algorithm is as follows: 
$$ \left( \bs X^{(k)}, \bs h^{(k)} \right) \rightarrow \left(\bs X^{(k+1)}, \bs h^{(k)} \right) \rightarrow \left(\bs X^{(k+1)}, \bs h^{(k+1)} \right).$$

It has been shown in~\cite{5226594} that this algorithm is able to provide fairly good results. However, there are not theoretical guarantees for its convergence. Recently, \cite{Attouch:2010:PAM:1836121.1836131} proposed a proximal alternating minimization algorithm where cost-to-move functions are added to the common alternating algorithm. These functions are quadratic costs that penalize variations between two consecutive iterations. They guarantee the algorithm convergence provided some mild conditions are met on the regularity of the objective function $\cl L\left(\bs X, \bs h \right)$ and the constraints in (\ref{eq:BID3}). The algorithm iterates as follows:  
\begin{align}
\label{eq:alternating_attouch} 
		\bs X^{(k+1)} & = \textrm{argmin}_{\tilde{\bs X}} \ \cl L\left(\tilde{\bs X}, \ \bs h^{(k)}\right) \ + \textstyle\frac{\lambda_x^{(k)}}{2} \| \ \tilde{\bs X} - \bs X^{(k)} \ \|_F^2, \\ \nonumber
		\bs h^{(k+1)} & = \textrm{argmin}_{\tilde{\bs h}} \ \cl L\left(\bs X^{(k+1)}, \tilde{\bs h}\right) + \textstyle\frac{\lambda_h^{(k)}}{2} \| \ \tilde{\bs h} - \bs h^{(k)}  \ \|_2^2,
\end{align}
where $\lambda_x$ and $\lambda_h$ are the cost-to-move penalization parameters that control the variations between two consecutive iterations. Section \ref{sec:lambda} describes how these parameters are tuned.

The non-convexity of (\ref{eq:BID3}) prevents attaining a global minimizer. However, it can be shown that, since the objective function $\cl L\left(\bs X, \bs h\right)$ and the constraints in (\ref{eq:BID3}) meet the conditions stated in~\cite{Attouch:2010:PAM:1836121.1836131}, the algorithm (\ref{eq:alternating_attouch}) converges to a critical point of the problem. Later in this section we describe how to stop the iterations so that the vicinity of a critical point of (\ref{eq:BID3}) is reached.

\subsection{Cost-to-move parameters}
\label{sec:lambda}

The presence of cost-to-move parameters ($\lambda_x$, $\lambda_h$) different than zero ensures the convergence of the algorithm (\ref{eq:alternating_attouch}) to a critical point of $\left(\bs X, \bs h\right)$~\cite{Attouch:2010:PAM:1836121.1836131}. However, their values are not crucial in the reconstruction results. In the following, we describe the tuning criteria used in this work which is based on the works of Puy and Vandergheynst~\cite{puy:robust}. The iterations start with high values of $\lambda_x$ and $\lambda_h$, keeping the image and the filter estimations closer to the initial values. When the number of iterations $k$ increases, the filter and the image estimates become more and more accurate and the parameters $\lambda_x$ and $\lambda_h$ can be progressively decreased.

\subsection{Stopping criteria}
\label{sec:stopping-criteria}

It is important to find an automatic criterion for stopping the iterations in (\ref{eq:alternating_attouch}) when the solution reaches the vicinity of a critical point of (\ref{eq:BID3}). A usual stopping criterion is based on the quality of the reconstruction with respect to the Ground Truth image and filter, which are in general unavailable. As proposed by Almeida and Figueiredo~\cite{6497608}, we use the spectral characteristics of the noise to analyze the quality of the estimation. This allows stopping the minimization when a high quality solution has been obtained. Let us define each residual image at iteration $k$ as the difference between the observed image and the blurred estimate: 
\begin{equation}
\label{eq:residual}
	\bs R^{(k)} = \bs Z - \bs \Phi(\bs h^{(k+1)}) \ \bs X^{(k+1)} = \left( \bs r_1^{(k)}, \cdots, \bs r_P^{(k)} \right). 
\end{equation}

Since the observation model is degraded by an additive white noise, we know that the residual image is spectrally white if the estimation at iteration $k$ has a good quality, otherwise the residual contains structured artifacts. Therefore, the iterations can be stopped when the residual is spectrally white, \ie when the 2-D autocorrelation $C_{\bs r_j\bs r_j}(m,n)$ of each residual image $\bs r_j^{(k)}$ is approximately the function $\delta_0(m,n)$. Let us note that, for the computation of the 2-D autocorrelation function, each residual vector $\bs r_j^{(k)}$ needs to be previously transformed into a matrix of $n \times n$ pixels and normalized to zero mean and unit variance.

Considering a $(2L+1)\!\times\!(2L+1)$ window, we compute the distance between the autocorrelation and the Kronecker function $\delta_0$ as follows:
\begin{equation}
\label{eq:MR}
	\cl M(\bs r_j^{(k)}) = - \textstyle\sum_{(m,n)=(-L,-L)}^{(L,L)} \left(\delta_0(m,n) - C_{\bs r_j\bs r_j}(m,n)\right)^2 = - \textstyle\sum_{\substack{(m,n)=(-L,-L)\\(m,n) \neq (0,0)}}^{(L,L)} \left(C_{\bs r_j\bs r_j}(m,n)\right)^2,
\end{equation}
which is higher for whiter residuals. As suggested by Almeida and Figueiredo~\cite{6497608}, in our experiments we have used $L = 4$. We then average over the measures obtained for each residual image:
\begin{equation}
\label{eq:MRP}
	\cl M(\bs R^{(k)}) = \textstyle\frac{1}{P} \sum_{j=1}^P \cl M(\bs r_j^{(k)}).
\end{equation}

In practice, we observe that $\cl M(\bs R^{(k)})$ has large negative values at the beginning of the iterations when the image and the filter are not properly estimated. Then, the value of $\cl M(\bs R^{(k)})$ increases with $k$ until it reaches a maximum close to zero. Finally, it starts to decrease again after the algorithm has converged to the best estimations for a fixed value of $\rho$ and starts overfitting the noise. Therefore, the iterations can be stopped when the maximum of the whiteness measure ($\cl M$) is reached. A similar behavior has also been observed in~\cite{6497608}.

\subsection{Numerical Reconstruction}

The proximal alternating minimization algorithm is summarized in Algorithm~\ref{alg:alternatemin}. This algorithm requires to set the initial values of the image, the filter and the regularization parameter. The initialization of the image and the filter is discussed in Appendix~\ref{sec:init_ImageFilter}, and the regularization parameter is tuned iteratively as explained in Appendix~\ref{sec:param}.

\begin{algorithm}[ht]
\newcommand{\sComment}[1]{\vspace{1mm}\Statex\quad\ \emph{\underline{#1}}:\vspace{1mm}}
  \caption{Proximal Alternating Minimization Algorithm
    \label{alg:alternatemin}}
  \begin{algorithmic}[1]
    \Require{\parbox[t]{10cm}{$\bs X^{(1)} = \bs X_0$; $\bs h^{(1)} = \bs h_0$; $\lambda_x^{(1)}\!= \lambda_h^{(1)}\!= \rho = \rho_0$; $\Delta = 0.75$; MaxIter =20}\vspace{2mm}}
		\For{$k = 1$ to MaxIter}
			\sComment{1st step, image estimation}
				\State $\bs X^{(k+1)} = \textrm{argmin}_{\tilde{\bs X} \in \cl P_0} \ \cl L\left(\tilde{\bs X}, \bs h^{(k)}, \rho \right) + \frac{\lambda_x^{(k)}}{2} \| \ \tilde{\bs X} - \bs X^{(k)} \ \|_F^2$
			\sComment{2nd step, filter estimation}
				\State $\bs h^{(k+1)} = \textrm{argmin}_{\tilde{\bs h} \in \cl D} \ \cl L\left(\bs X^{(k+1)}, \tilde{\bs h}\right) + \frac{\lambda_h^{(k)}}{2} \| \ \tilde{\bs h} - \bs h^{(k)} \ \|_2^2$
           	\sComment{Parameters update}
				\State $\lambda_x^{(k+1)} = \lambda_x^{(k)} \Delta$
				\State $\lambda_h^{(k+1)} = \lambda_h^{(k)} \Delta$
			\sComment{Compute residuals and whiteness measure}
				\State $\bs R^{(k)} = \bs Z - \bs\Phi(\bs h^{(k+1)}) \ \bs X^{(k+1)}$
				\State Compute $\cl M(\bs R^{(k)})$ using (\ref{eq:MRP})
                 \sComment{Stop when residual is spectrally whiter}
				\If{ $ \cl M^{(k+1)} < \cl M^{(k)}$ }\quad  \rm break.
				\EndIf
		\EndFor
		\State Return $\bs X^{(k)}$ and $\bs h^{(k)}$.
  \end{algorithmic}
\end{algorithm}

The first step in Algorithm \ref{alg:alternatemin} estimates the image by minimizing a sum of three convex functions with the image constrained to the set $\cl P_0$. This constraint can be handled by adding the convex indicator function on the set, \ie $\imath_{\cl P_0}(\bs X)$. The resulting optimization, containing a sum of two smooth and two non-smooth convex functions, can be solved using the primal-dual algorithm of Chambolle and Pock~\cite{Chambolle2011} summarized in Appendix \ref{sec:appendix_CP}.

The second step in Algorithm \ref{alg:alternatemin} estimates the filter by minimizing a sum of two smooth convex functions with the filter constrained to the set $\cl D$. The optimization problem can be solved using the Accelerated Proximal Gradient (APG) Method~\cite{OPT-003}. This algorithm is able to minimize a sum of a non-smooth and a smooth convex function, which allows us to handle the constraint as a convex indicator function on the set, \ie $\imath_{\cl D}(\bs h)$. The APG algorithm is briefly described in Appendix \ref{sec:appendix_APG}. 

The algorithms used to solve the minimization problems described above were chosen because they are well suited to solve each problem optimally. Since the cost functions are different in each step, the same algorithm could not be used to optimally find a solution of both problems. The presence of two non-smooth functions in the first step, prevents the use of the APG algorithm. For the second step, the algorithm of Chambolle and Pock~\cite{Chambolle2011} could be used to solve the optimization problem, however, this algorithm presented a slower convergence than APG, which is optimal when the optimized cost contains a differentiable function.

Let us note that, in the numerical reconstruction, the convolution operator is implemented using the Fast Fourier Transform (FFT). This operation assumes that the boundaries of the image are periodic, a condition that is far from reality. In actual imaging, no condition can be assumed on the boundary pixels, \ie they cannot be assumed to be zero, neither periodic, nor reflexive. Not taking care of the boundaries conditions would introduce ringing artifacts. Based on the work of Almeida and Figueiredo~\cite{6502713}, in the numerical experiments we consider unknown boundary conditions, \ie, the pixels belonging to the image border are not observed. To take this into account, the problem formulation needs to be appropriately modified as discussed in Appendix \ref{sec:appendix_ubc}.

\section{Filter validation through non-blind deconvolution}
\label{sec:NBID}

This section is dedicated to the validation of the filter estimated using Algorithm~\ref{alg:alternatemin}. Since the true emissions of the Moon and Venus are zero, the deconvolution with a correct filter should remove all the celestial bodies' apparent emissions. Therefore, the validation process consists of taking an image from a transit observation that was not used for filter estimation (\ie $\bs y_j \notin \{\bs y_1, \cdots, \bs y_P\}$), deconvolving this image using the estimated filter $\bs h$, and verifying that, in the reconstructed image $\bs x_j^*$, the pixels inside the celestial body disk are zero.

To obtain this reconstructed image, we formulate a least squares minimization problem regularized using the prior information available on the image, that is, that the image is non-negative and has a sparse representation on a suitable wavelet basis $\bs \Psi \in \Rbb^{W \times N}$. The proposed non-blind deconvolution method reads as follows:
\begin{align}
\label{eq:NBID}
	\bs x_j^* = \textrm{argmin}_{\tilde{\bs x}_j} & \ \rho \ \| \ \bs S_\Delta \ \bs \Psi^\mathtt{T} \ \tilde{\bs x}_j \ \|_1 + \textstyle\frac{1}{2} \| \ \bs z_j - \bs h \otimes \tilde{\bs x}_j \ \|^2_2 \\ \nonumber
	\textrm{s.t.} & \ \tilde{\bs x}_j \in \cl P 
\end{align}
where $\bs S_\Delta \in \Rbb^{S \times W}$ is the selection operator of the set $\Delta$, with $|\Delta| = S$, which contains the detail coefficients. The convex set $\cl P$ is defined as $\cl P = \left\{ \bs u \in \Rbb^{N} : u_i \geq 0 \right\}$. The non-blind image estimation is solved using an extension of the primal-dual algorithm of Chambolle and Pock~\cite{Chambolle2011} (see Section~\ref{sec:appendix_CP} for more details). Since this algorithm is able to minimize a sum of three non-smooth convex functions, the constraint can be handled as a convex indicator function. Notice that other algorithms could be used such as the Generalized Forward Backward algorithm~\cite{raguet2013generalized}. 

\section{Experiments}
\label{sec:Results}

In this section, we first present some synthetic results that allow us to validate the effectiveness of the proposed blind deconvolution method to recover the correct filter. Later we present experimental results on images taken by the SDO/AIA and SECCHI/EUVI telescopes.

All algorithms were implemented in MATLAB and executed on a 3.2 GHz Intel i5-650 CPU with 3.7 GiB of RAM, running on a 64 Bit Ubuntu 14.04 LTS operating system.

\subsection{Synthetic Data}
\label{sec:SynthData}

\begin{figure}[ht]
  \centering
	\hfill \includegraphics[height=3.75cm]{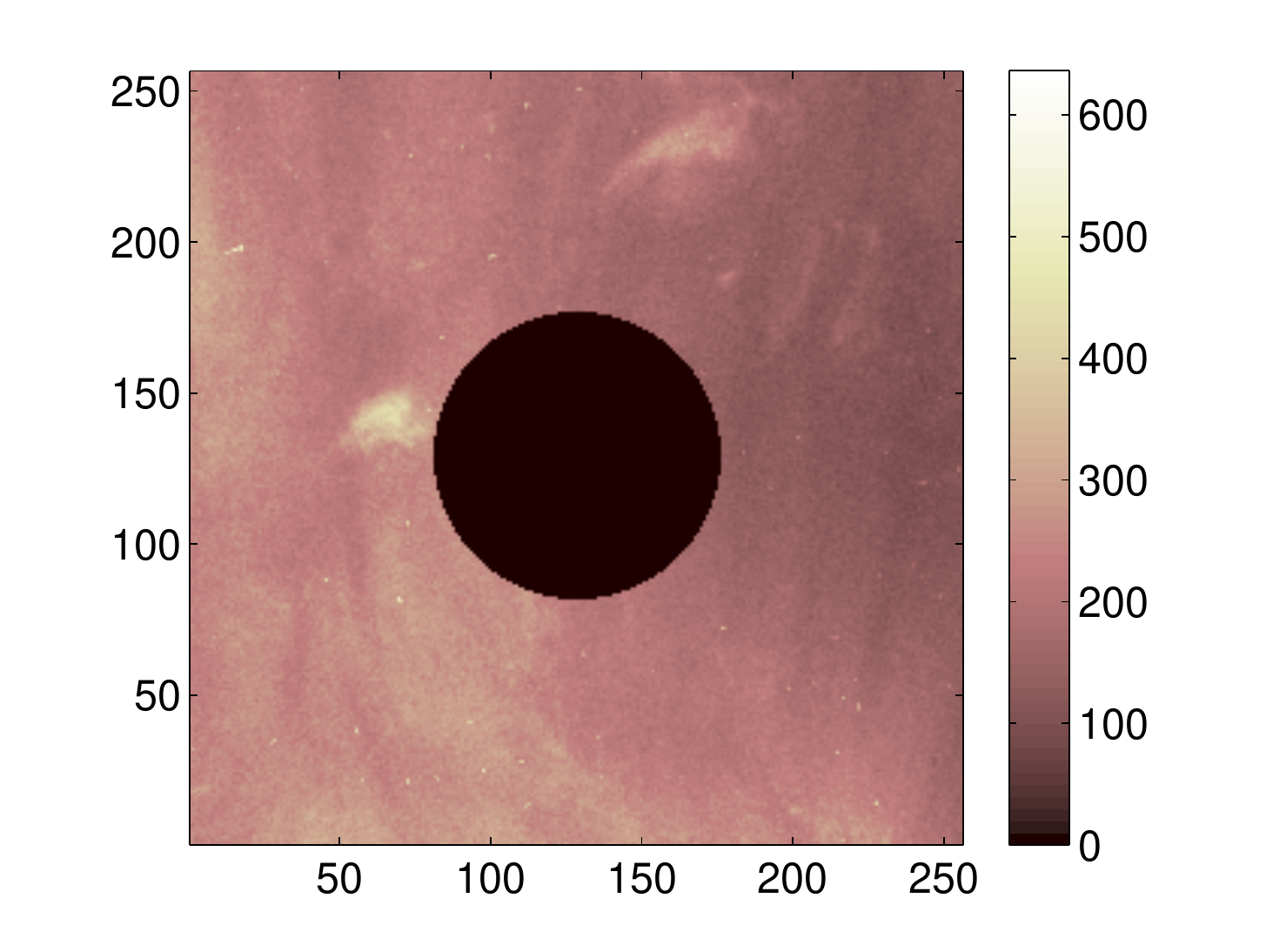} \hfill
	\includegraphics[height=3.75cm]{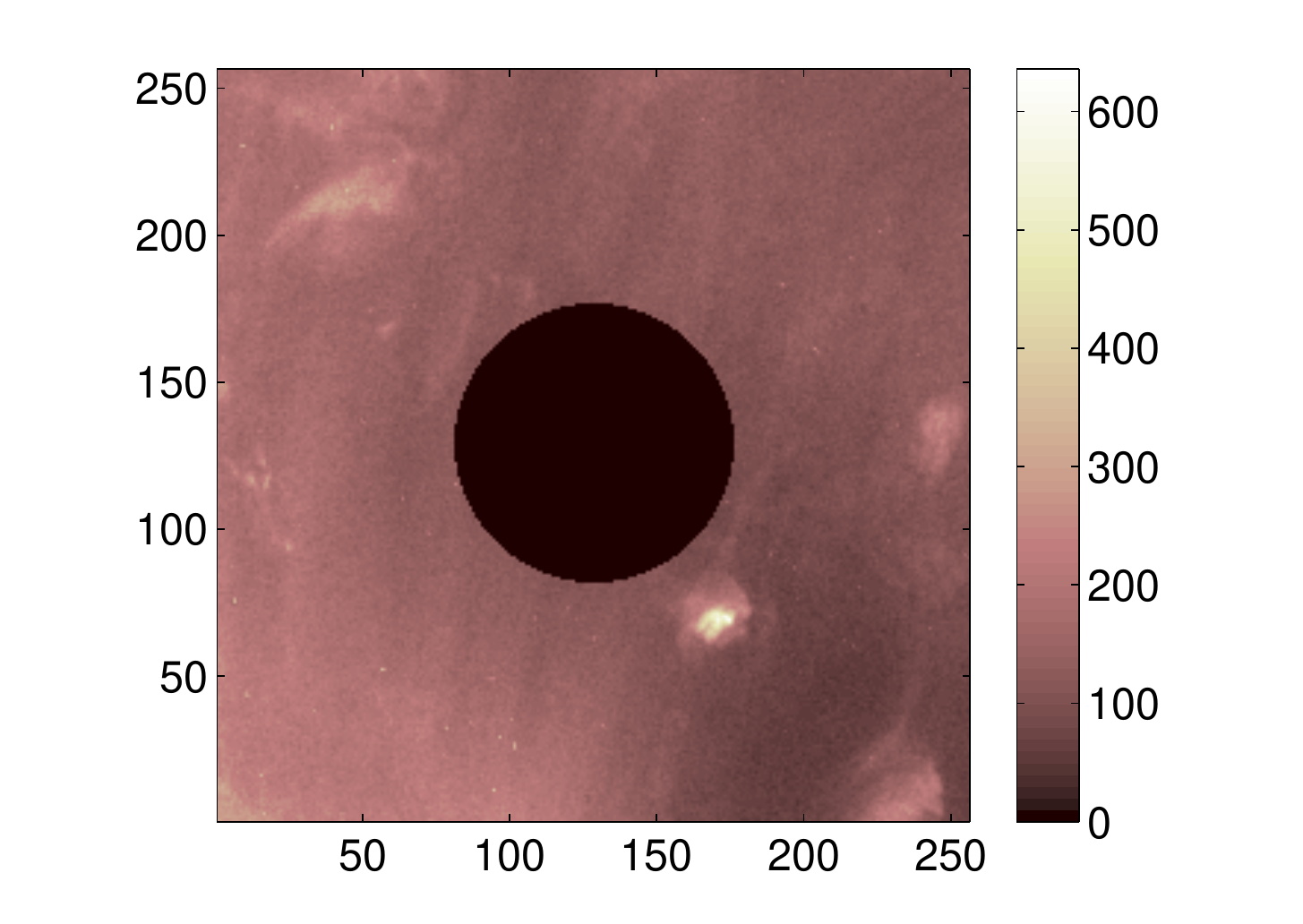} \hfill
	\includegraphics[height=3.75cm]{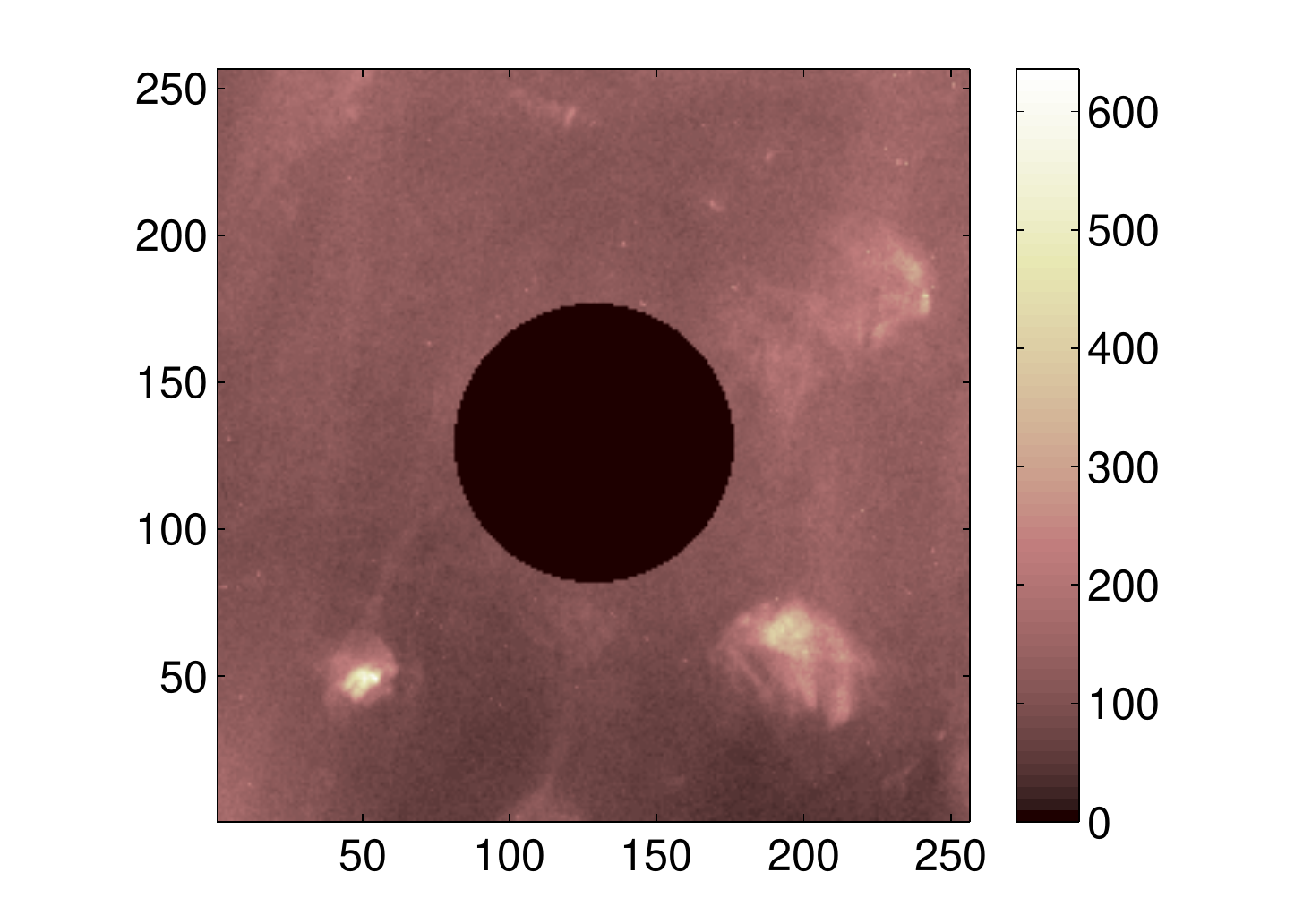}
  \caption{Realistic images: (left) first image $\bs x_1$, (center) second image $\bs x_2$ and (right) third image $\bs x_3$ from the simulated transit.}
 \label{fig:SyntheticImages}
\end{figure}

Three synthetic images are selected to test the reconstruction ($P = 3$). They are defined on a \linebreak $256\!\times\!256$ pixel grid ($N = 256^2$). To simulate actual images from the celestial transit, we take an image of the Sun (the image observed by SDO/AIA at 00:00 UT on June 6th 2012) and select three different cutouts of $N$ pixels each. The center of the cutouts are arbitrarily selected such that the corresponding regions are sufficiently distant from each other and correspond to a sample of a full Venus trajectory. In each image, we add a black disk of radius $R = 48$ pixels, centered at the pixel $[129, \ 129]$, which represents the celestial body. Figure~\ref{fig:SyntheticImages} depicts the images from the simulated transit. 

During the solar transit, the position and size of the celestial body are known at all moments. However, when this event is imaged, the coordinates of the center represented in the discrete image grid may contain an error of maximum one pixel. This uncertainty is simulated in our synthetic experiments by considering the sets $\Omega_j$ and $\Theta$ to be defined using disks of radius $r_\Omega$ and $r_\Theta$, respectively, with one pixel difference with respect to the actual object's radius, \ie $r_\Omega = R - 1$ and $r_\Theta = R + 1$.

\begin{figure}[ht]
  \centering
	\hfill \includegraphics[height=4.1cm]{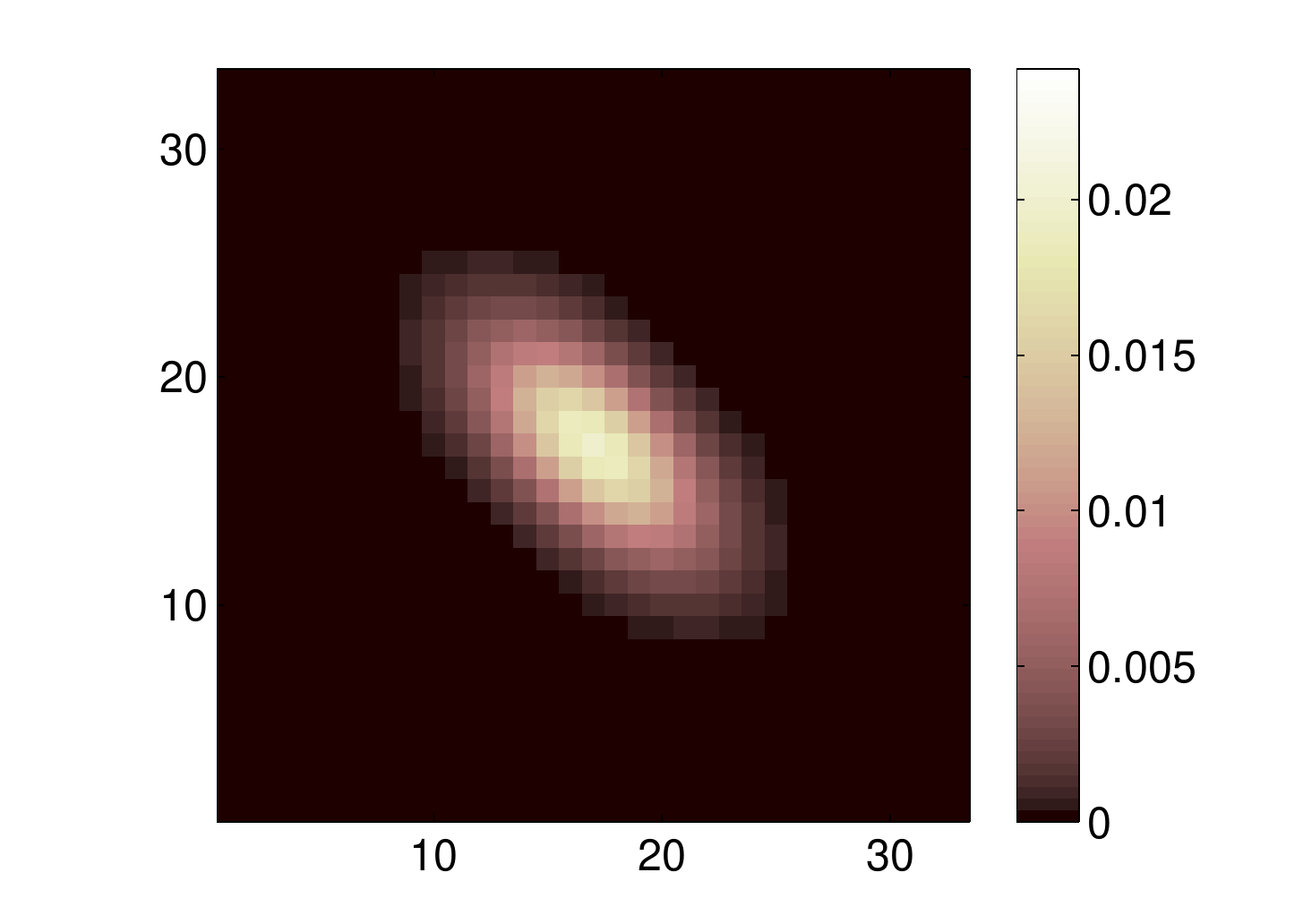} \hfill
	\includegraphics[height=4.1cm]{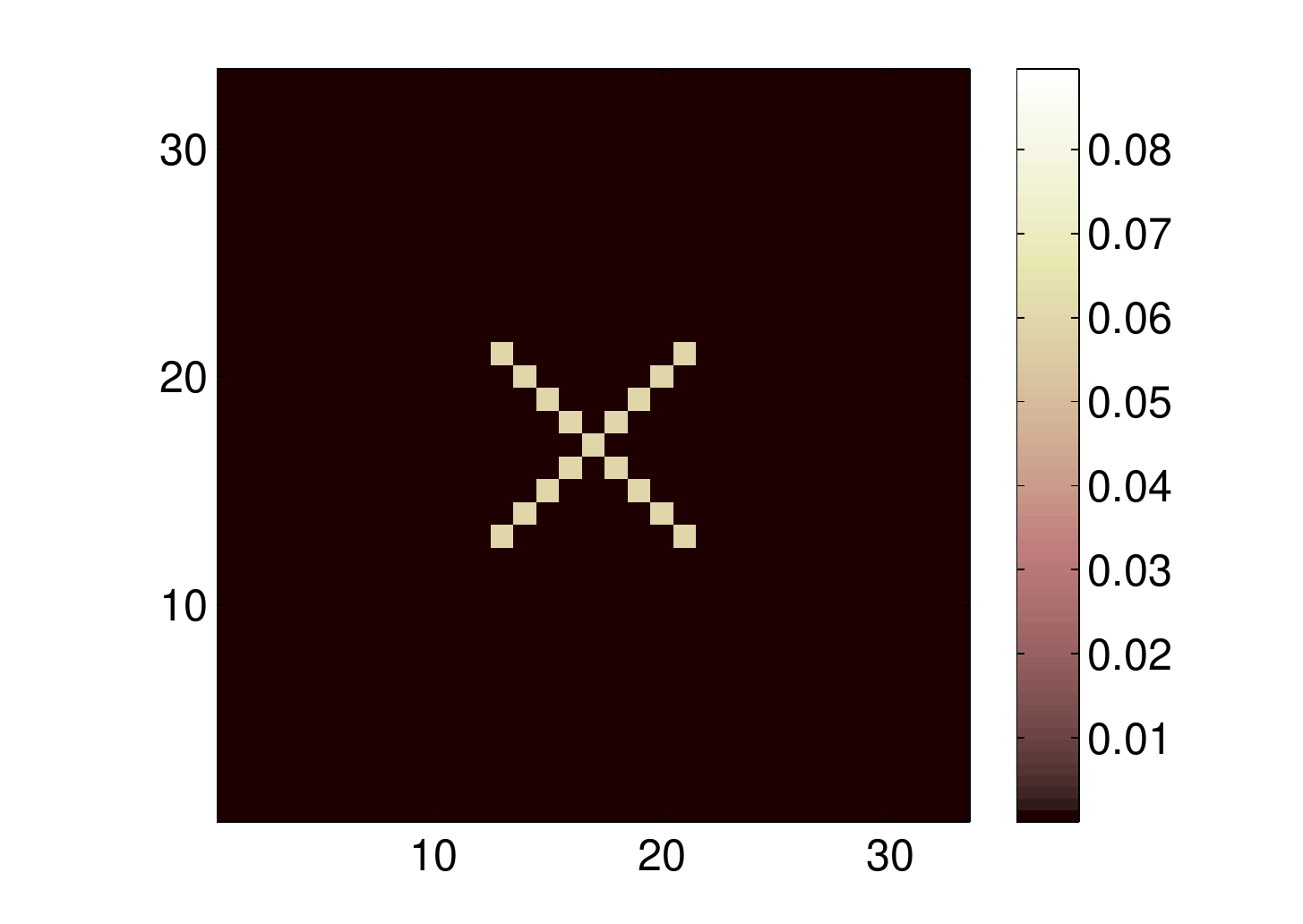} \hfill
  \caption{Realistic filters in a $(2b+1)\!\times\!(2b+1)$ window with $b=16$. (left) Anisotropic Gaussian Filter. (right) X Filter.}
 \label{fig:SyntheticFilters}
\end{figure}

Two kinds of discrete filters are selected and they have a limited support on a $33\!\times\!33$ pixel grid ($b = 16$). The first filter is simulated by an anisotropic Gaussian function with standard deviation of 2 pixels horizontally and 4 pixels vertically, rotated by 45 degrees (see Figure~\ref{fig:SyntheticFilters}-(left)). The second filter is simulated by a X shape (see Figure~\ref{fig:SyntheticFilters}-(right)), hereafter called the X filter. These functions help to demonstrate the capacity of the proposed method to recover not only diffusion filters such as the anisotropic Gaussian, but also diffraction filters such as the X example.
Let us note that, in these synthetic experiments, there is no need to test the long-range assumption on the PSF.

The measurements are simulated according to (\ref{eq:discrete_model_vector}), where additive white Gaussian noise $\bs N \sim_{\rm iid}~\cl N(0,\sigma^2)$ is added in order to simulate a realistic scenario. The noise variance ($\sigma^2$) is set such that the blurred signal to noise ratio ${\rm BSNR} = 10 \log_{10} \left(\textrm{var} \left(\bs \Phi(\bs h) \ \bs X \right) / \sigma^2\right)$ is equal to 30 dB, which corresponds to a realistic BSNR in the actual observations.

During the experiments, the wavelet transformation used for the sparse image representation is the redundant Daubechies wavelet basis with two vanishing moments and three levels of decomposition~\cite{Mallat:2008:WTS:1525499}. Other mother wavelets and more vanishing moments can be considered, however, due to the dictionary redundancy, the choice does not have a significant impact on the reconstruction results. Higher levels can also be considered but the computational time is notably higher and the reconstruction quality does not significantly improve. 

As discussed in Section~\ref{sec:ImagePrior}, the wavelet coefficients whose support touches the boundary of the occulting body are not sparse as they spread over all the scales of the wavelet transform. Thus, we need to define the set $\Theta$ containing the coefficients that are not affected by the disk. To determine this set, we first generate a set of images with constant background that contain disks of radius $r_\Theta$ centered as the celestial body. All the elements inside the disks are set as random values with a standard uniform distribution ($\cl U(0,1)$). We then compute the wavelet coefficients of this image on the basis $\bs \Psi$. Finally, the set $\Theta$ is composed by the indexes of the \textit{detail} coefficients that are equal to zero.

The reconstruction quality of $\bs X^*$ with respect to the true image $\bs X^{GT}$ is measured using the increase in SNR
(ISNR) defined as $ {\rm ISNR}\ =\ 20\log_{10} \| \ \bs Y - \bs X^{GT} \ \|_F/\| \ \bs
X^* - \bs X^{GT} \ \|_F$. The reconstruction quality of $\bs h^*$ with respect to the true filter $\bs h^{GT}$ is measured using the reconstruction SNR (RSNR), where $ {\rm RSNR}\ =\ 20\log_{10} \| \ \bs h^{GT} \ \|_2/\| \ \bs h^{GT} - \bs h^* \ \|_2.$

Let us first analyze the behavior of the algorithm when we increase the number of observations $P$ used for the reconstruction. Table \ref{tab:SynthResults_Gauss2} presents the reconstruction quality of the results using the blind deconvolution problem for the anisotropic Gaussian filter of Figure~\ref{fig:SyntheticFilters}-(left). The results correspond to an average over four trials and they are presented for $P$ = 1, 2 and 3 observations. For $P = 2, 3$ we use a warm start by initializing the algorithm with the results obtained for $P-1$.

\begin{table}[ht]
	\centering
	\begin{tabular}{ l  c  c  c c }
		\hline \hline
		$P$ & ISNR $\bs X$ [dB] & RSNR $\bs h$ [dB] & $\cl M(\bs R)$ \\
		\hline
		1 & 1.27 & 9.63 & -0.25\\
		\hline
		2 & 1.33 & 13.07 & -0.16\\
		\hline
		3 & 1.36 & 14.05 & -0.03\\
		\hline
	\end{tabular}
	\caption{Reconstruction quality for different number of simulated transit observations using the synthetic anisotropic Gaussian filter.}
  \label{tab:SynthResults_Gauss2}
\end{table}

We notice that the reconstruction quality of the filter significantly improves as the number of observations $P$ increases. The whiteness measure $\cl M(\bs R)$ presents the same behavior, which indicates that the residual image is whiter when more observations are considered. This is explained by an increase of the available information in the filter reconstruction problem for the same number of unknowns.

Regarding the numerical complexity, the algorithm requires an average of five iterations on the value of $\rho$ and a total time of 1.5 hours. When the number of iterations on $\rho$ increases, we observed that the estimated residual energy is closer to the actual noise energy.

\begin{figure}[ht]
\centering
	\begin{tabular}{c c c}
		\includegraphics[height=3.75cm]{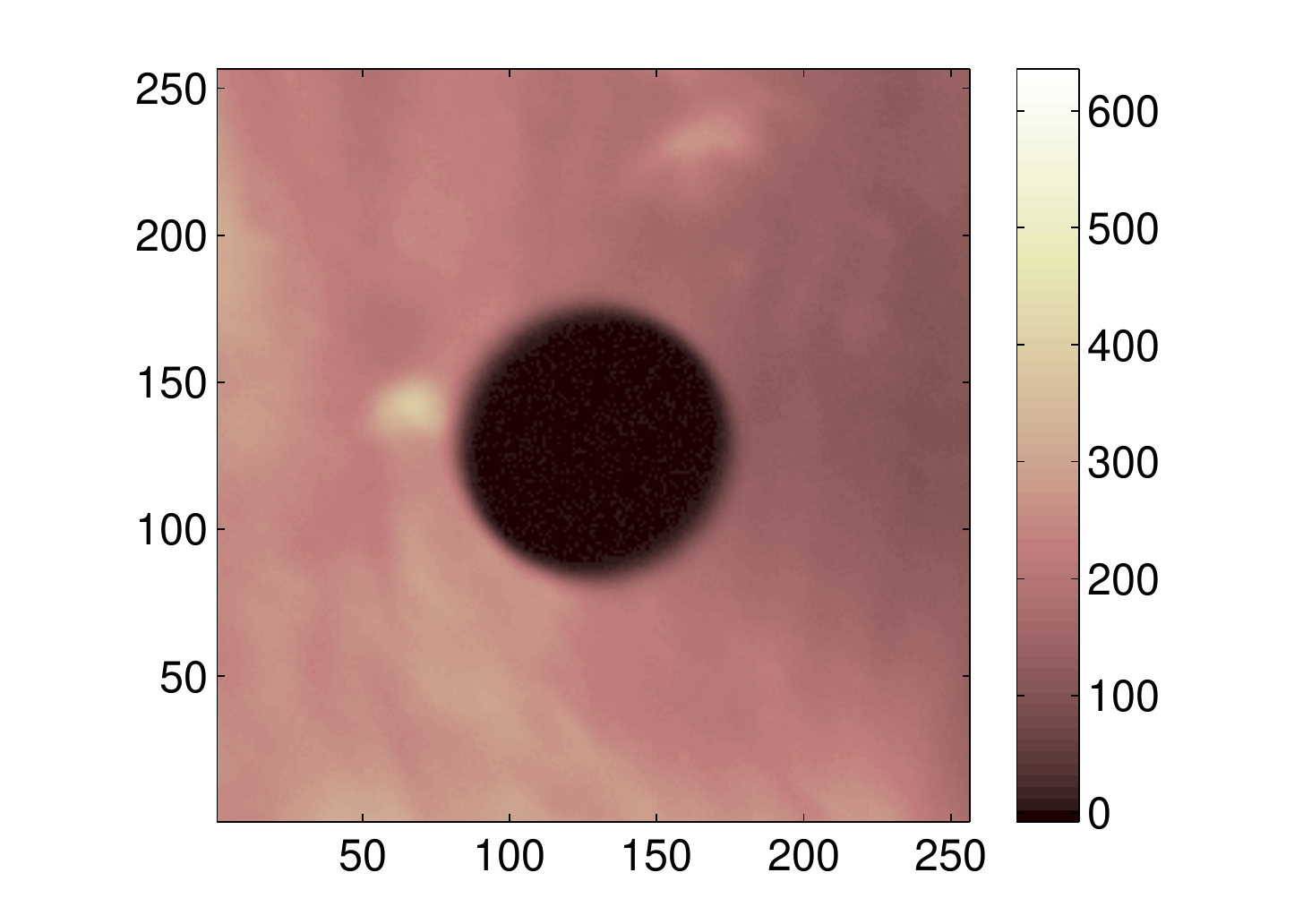} &
		\includegraphics[height=3.75cm]{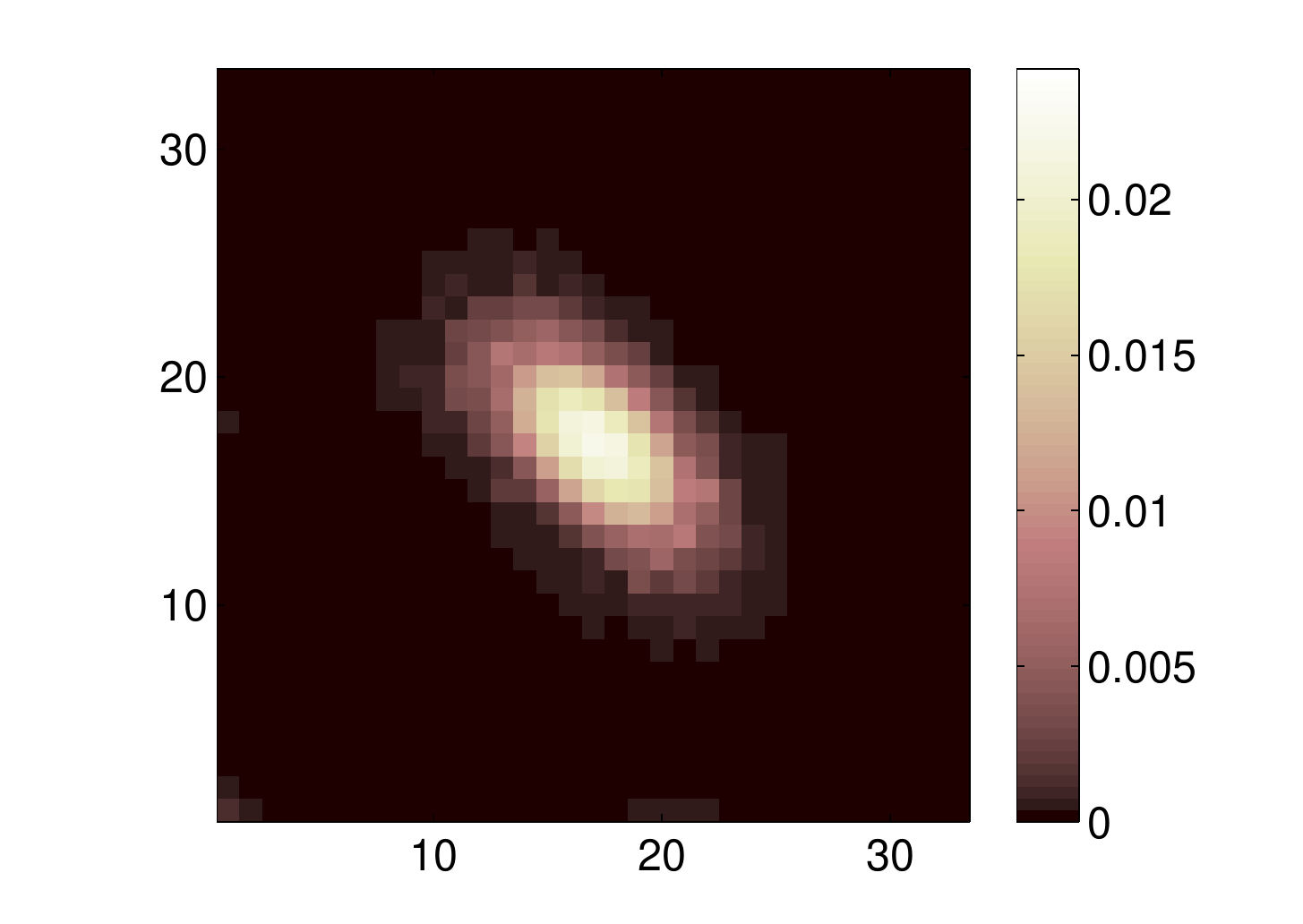} &
		\includegraphics[height=3.75cm]{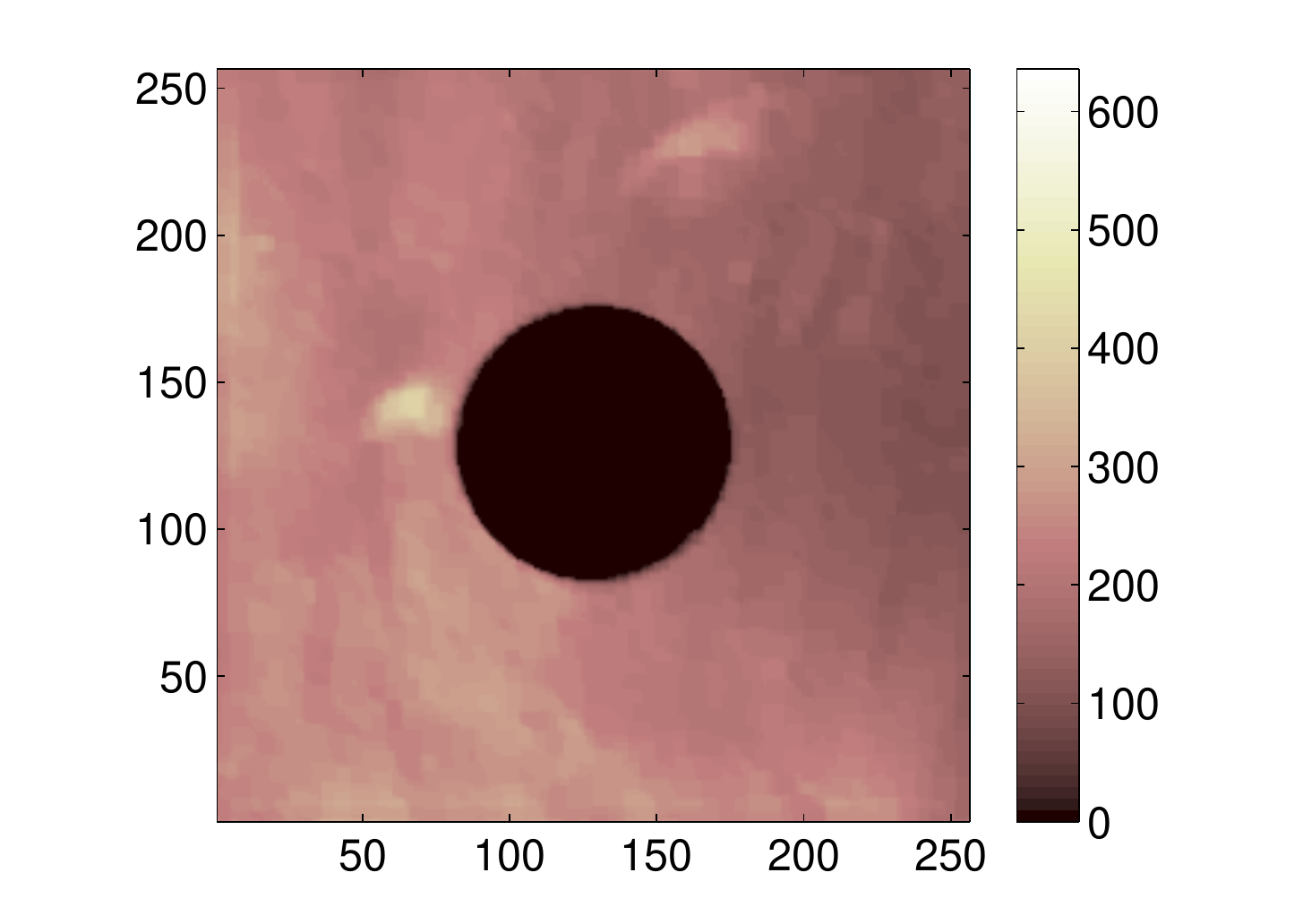}
	\end{tabular}
\caption{Results for the anisotropic Gaussian filter and $P = 3$. (left) Noisy observation of image~1 ($\bs y_1$). (center) Filter Reconstruction ($\bs h^*$) using blind deconvolution, RSNR = 14.05 dB. (right) Image Reconstruction ($\bs x_1^*$) using non-blind deconvolution, ISNR = 1.83 dB.}
 \label{fig:SynthResults_Gauss2_P3}
\end{figure}

Let us now show some reconstruction results of the different filters and how they reproduce the true zero pixel values inside the object. Figure~\ref{fig:SynthResults_Gauss2_P3}-(left) depicts the first noisy observed patch and Figure~\ref{fig:SynthResults_Gauss2_P3}-(center) presents the resulting PSF when reconstructing the anisotropic Gaussian filter for $P = 3$. We can observe how the algorithm is able to estimate the filter with a high quality, even when no hard constraints are introduced on the filter shape. The reconstruction quality can be further improved if stronger conditions are assumed on the filter. 

Figure~\ref{fig:SynthResults_Gauss2_P3}-(right) presents the reconstructed image using the estimated filter from Figure~\ref{fig:SynthResults_Gauss2_P3}-(center) in the non-blind deconvolution of Section \ref{sec:NBID}. Note that the majority of the pixels inside the estimated disk are zero, except for some numerical errors. To quantify this validation, we use as measurement the disk intensity, \ie the sum of the pixel values inside the disk. We compare the ratio between the disk intensity for the deconvolved image, denoted by $\cl S_X$, and the disk intensity for the observed patch, denoted by $\cl S_Y$. We obtained a disk intensity ratio of $\cl S_X/ \cl S_Y = 8.16 \times 10^{-2}$. This shows the effectiveness of the estimated filter to recover the true zero emissions inside the disk.

\begin{figure}[ht]
\centering
	\begin{tabular}{c c c}
		\includegraphics[height=3.75cm]{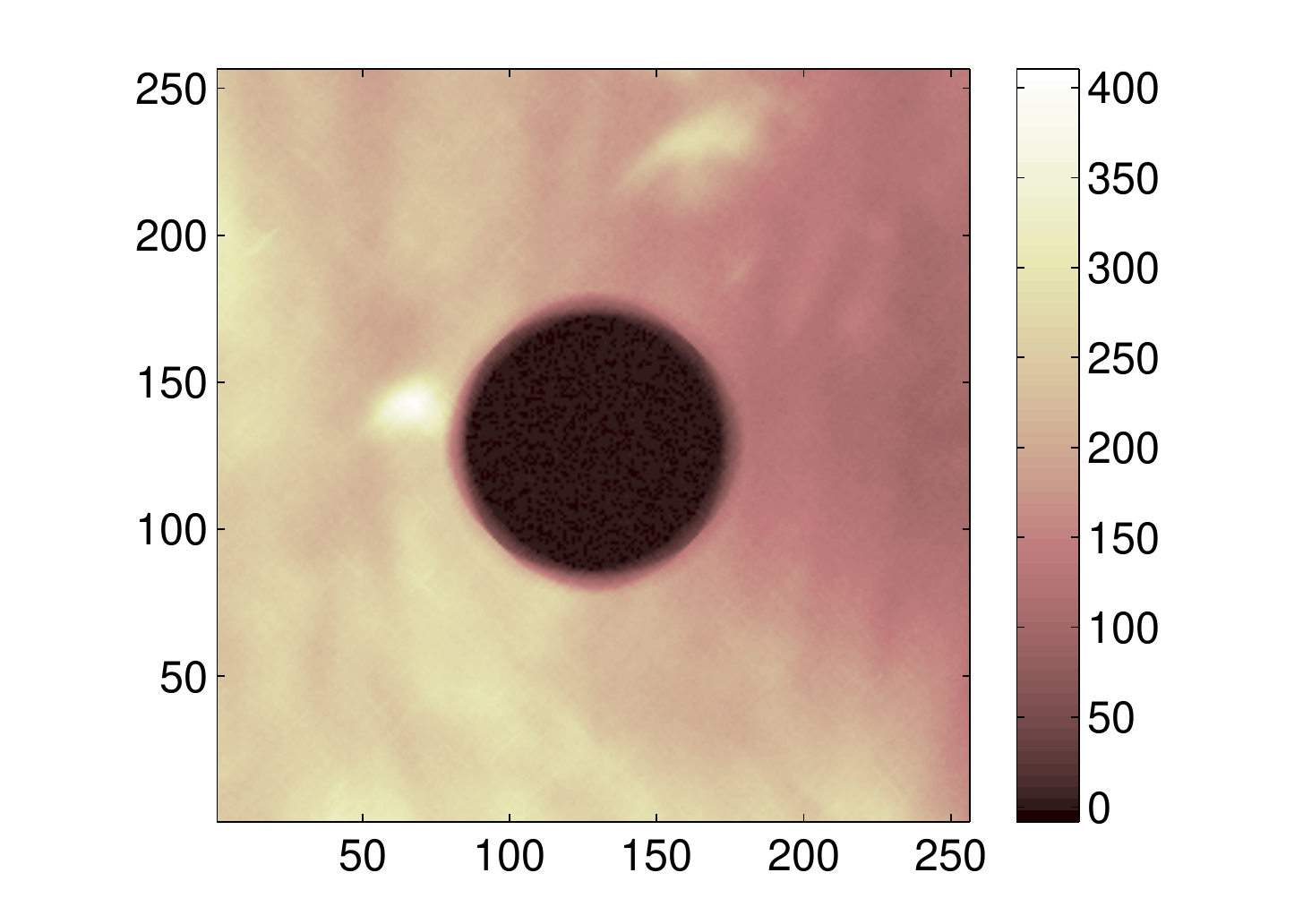} &
		\includegraphics[height=3.75cm]{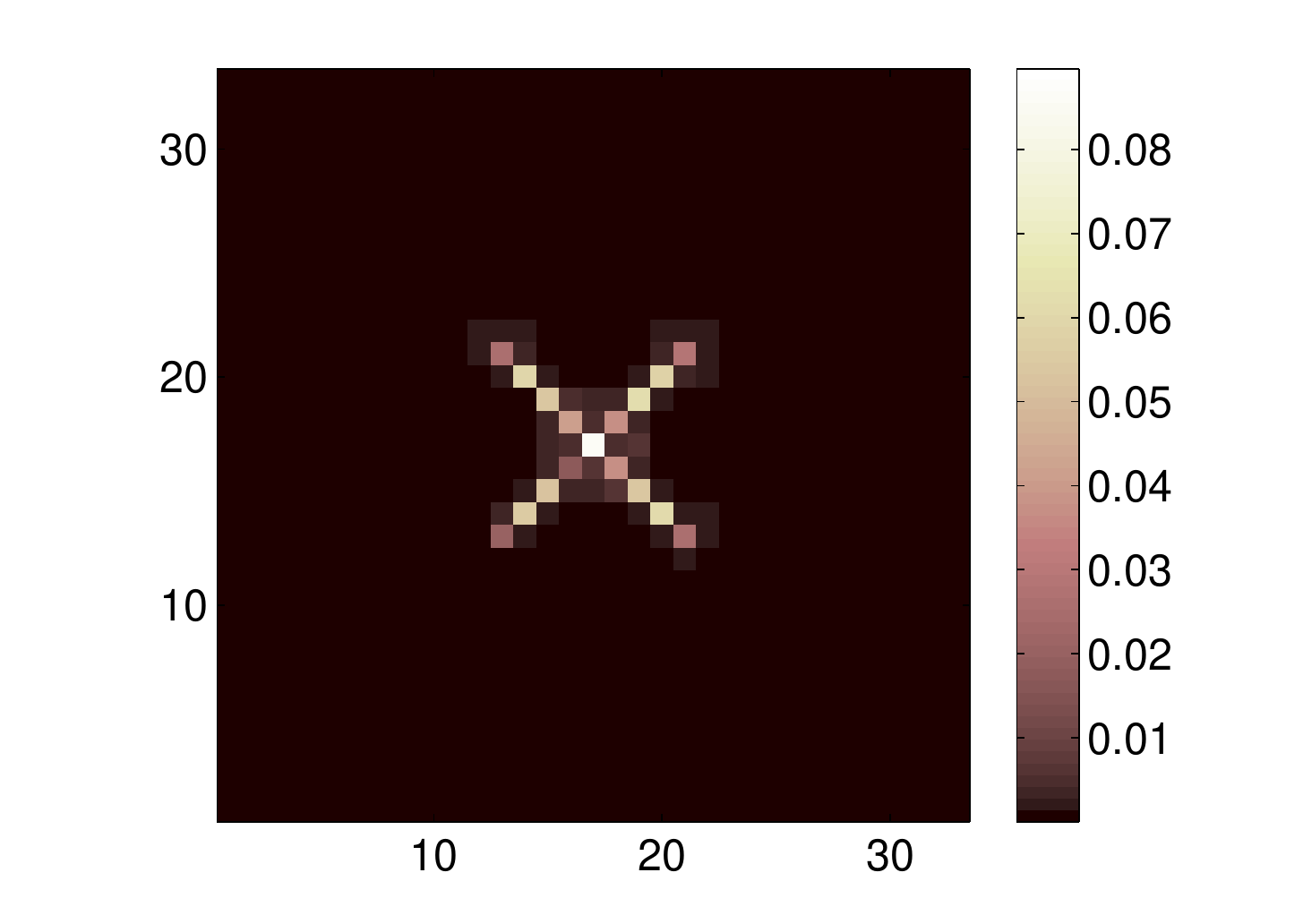} &
		\includegraphics[height=3.75cm]{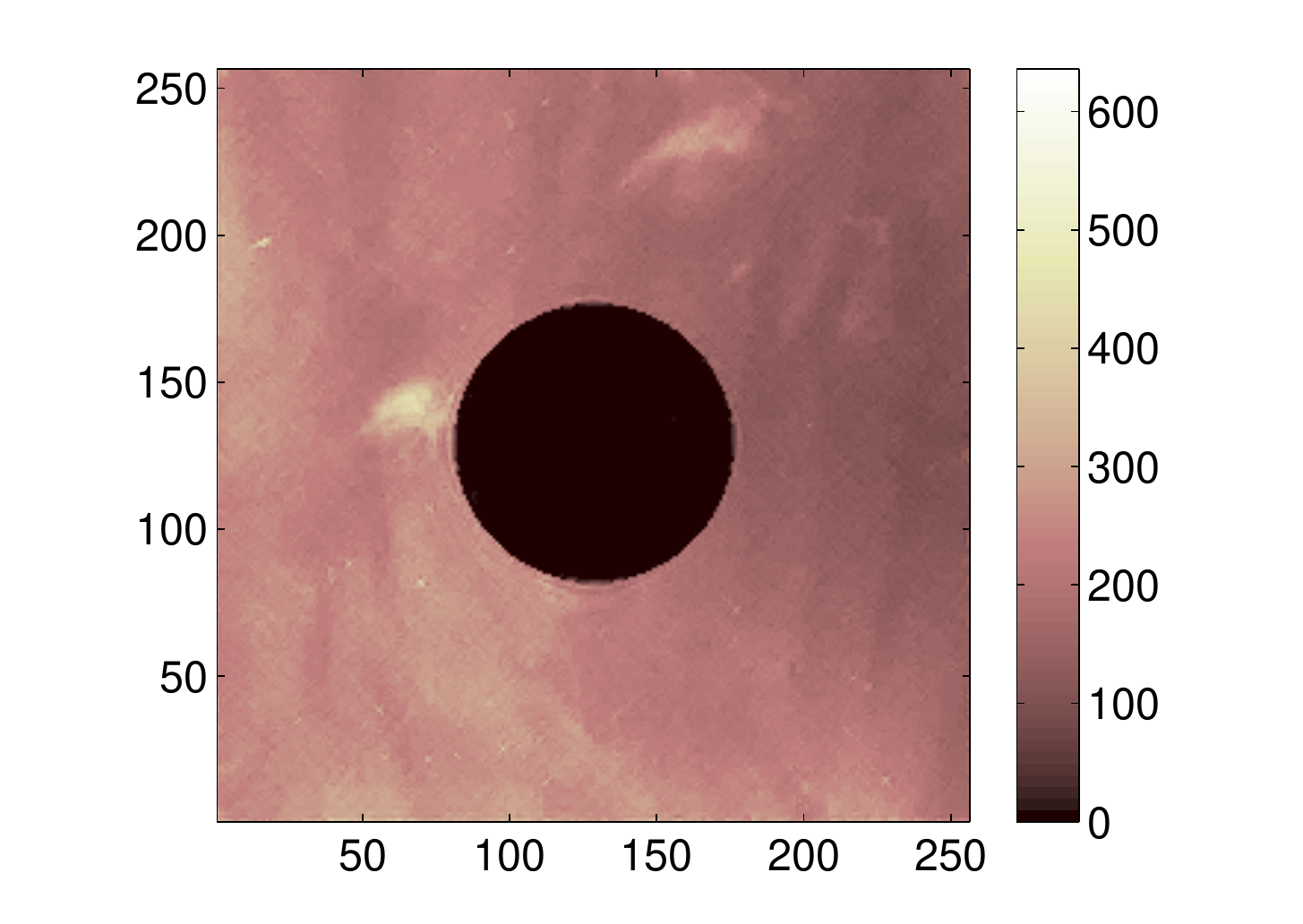}
	\end{tabular}
\caption{Results for the X filter and $P = 3$. (left) Noisy observation of image~1 ($\bs y_1$). (center) Filter Reconstruction ($\bs h^*$) using blind deconvolution, RSNR = 7.52 dB. (right) Image Reconstruction ($\bs x_1^*$) using non-blind deconvolution, ISNR = 2.55 dB.}
 \label{fig:SynthResults_X_P3}
\end{figure}

Figure~\ref{fig:SynthResults_X_P3} depicts the results for $P=3$ observations using the X filter of Figure~\ref{fig:SyntheticFilters}-(right). Let us note that the reconstruction quality is lower than in the case of the Gaussian filter because the X filter is more complex and hence it is harder to estimate without extra information on its shape. Nevertheless, since it causes less diffusion on the observation, the image reconstruction quality is better. When comparing the values inside the disk between the observations and the estimated images, we observed that the disk intensity ratio is $\cl S_X/ \cl S_Y = 2.03 \times 10^{-2}$, which validates the zero emissions inside the disk.

Finally, we also considered a case (not displayed here) where the observation is only corrupted by noise. As expected, the filter estimated by the algorithm is a high quality Kronecker delta function with RSNR = 66 dB. This result shows the capability of the algorithm to recover highly localized filters of one pixel radius. 

\subsection{Experimental Data}
\label{sec:ExpData}

The proposed blind deconvolution method was tested on two experimental sets of data. A first data set corresponds to the observations of the Venus transit on June 5th - 6th 2012 by the Atmospheric Imaging Assembly (AIA)~\cite{2012SoPh..275...17L} on board the Solar Dynamics Observatory (SDO). The second data set corresponds to the observations of the Moon transit on February 25th 2007 by the Extreme Ultraviolet Imager (EUVI)~\cite{2008SSRv..136...67H}, a part of the Sun Earth Connection Coronal and Heliospheric Investigation (SECCHI) instrument suite on board the STEREO-B spacecraft. These images are compressed using the RICE algorithm~\cite{AIAKeywords}, which is lossless~\cite{Pence2009} and hence does not introduce additional errors in the PSF estimation. 

In the following, we present for each telescope the results obtained when reconstructing the filter using the blind deconvolution approach. Then, we validate the obtained filters with transit images that were not used for the PSF estimation. Finally, we demonstrate how the obtained filters work when deconvolving non-transit images. All results are compared with previously estimated PSFs.

\subsubsection{SDO/AIA - Venus transit}

We consider three $4096 \times 4096$ level 1 images from the transit ($P = 3$) recorded by the 19.3~nm channel of AIA. The filter is assumed to have a limited support inside a $129 \times 129$ pixel grid ($b = 64$), which allows encompassing 99\% of the energy of previously estimated PSFs. The presence of a long-range PSF was verified in the observations by analyzing the pixels inside the disk of Venus. To estimate this effect, we considered a total of 10~patches and computed the mean intensity value on a disk of radius 10~pixels inside the disk of Venus. The obtained value $\mu~=~43.3$ DN was removed from the observations to estimate the filter using the blind deconvolution scheme.

The radius of the Venus disk is known to be 49~pixels and hence, the disk represents a small area inside the high resolution image. Since the blind deconvolution takes benefit mainly from the black disk in the transit images, we select a patch of size $N = 256^2$ centered on the Venus disk. As explained in Appendix~\ref{sec:appendix_ubc}, our method considers unknown border pixels based on the filter support (completely defined by $b$ after removing the value of $\mu$ from the observed patch). Therefore, cropping the observed image does not have any effects on the reconstruction results. Furthermore, considering smaller observations keeps the optimization problem computationally tractable.

As explained in Section~\ref{sec:noise_estim}, we use an \emph{adaptive} noise estimation strategy to optimize the AWGN assumption in~(\ref{eq:discrete_model_matrix}). For this, we start from $\sigma = \sigma_{RME}$, the value estimated by the Robust Median Estimator (RME). Then, the blind deconvolution method is performed for different values of $\sigma$ that are multiples of $\sigma_{RME}$. Finally, we take the value of $\sigma$ that optimizes the whiteness of the residual $\cl M(\bs R)$ as defined in~(\ref{eq:MRP}), \ie  the residual in~(\ref{eq:residual}) only contains white noise without any remaining of the signal features. The variance $\sigma_{RME}^2$ is computed as follows~\cite{donoho01091994}:

\begin{equation}
\label{eq:RME}
	\sigma_{RME}^2 = \frac{\textrm{med } | \ \bs \alpha \ |}{0.6745},
\end{equation}
where $\bs \alpha $ is a vector containing the finest scale wavelet coefficients of the observed vector $\bs y_j$, \ie $\bs \alpha = \bs S_\Lambda \bs \Psi^\mathtt{T} \bs y_j$, with $\bs S_\Lambda \in \bb R^{N \times W}$ the selection operator of the set $\Lambda$ that contains the finest scale wavelet coefficients. This estimation resulted in $\sigma_{RME} =$ 2.81 DN.

\begin{figure}[ht]
\centering
	\begin{tabular}{c c c}
		\includegraphics[height=4cm]{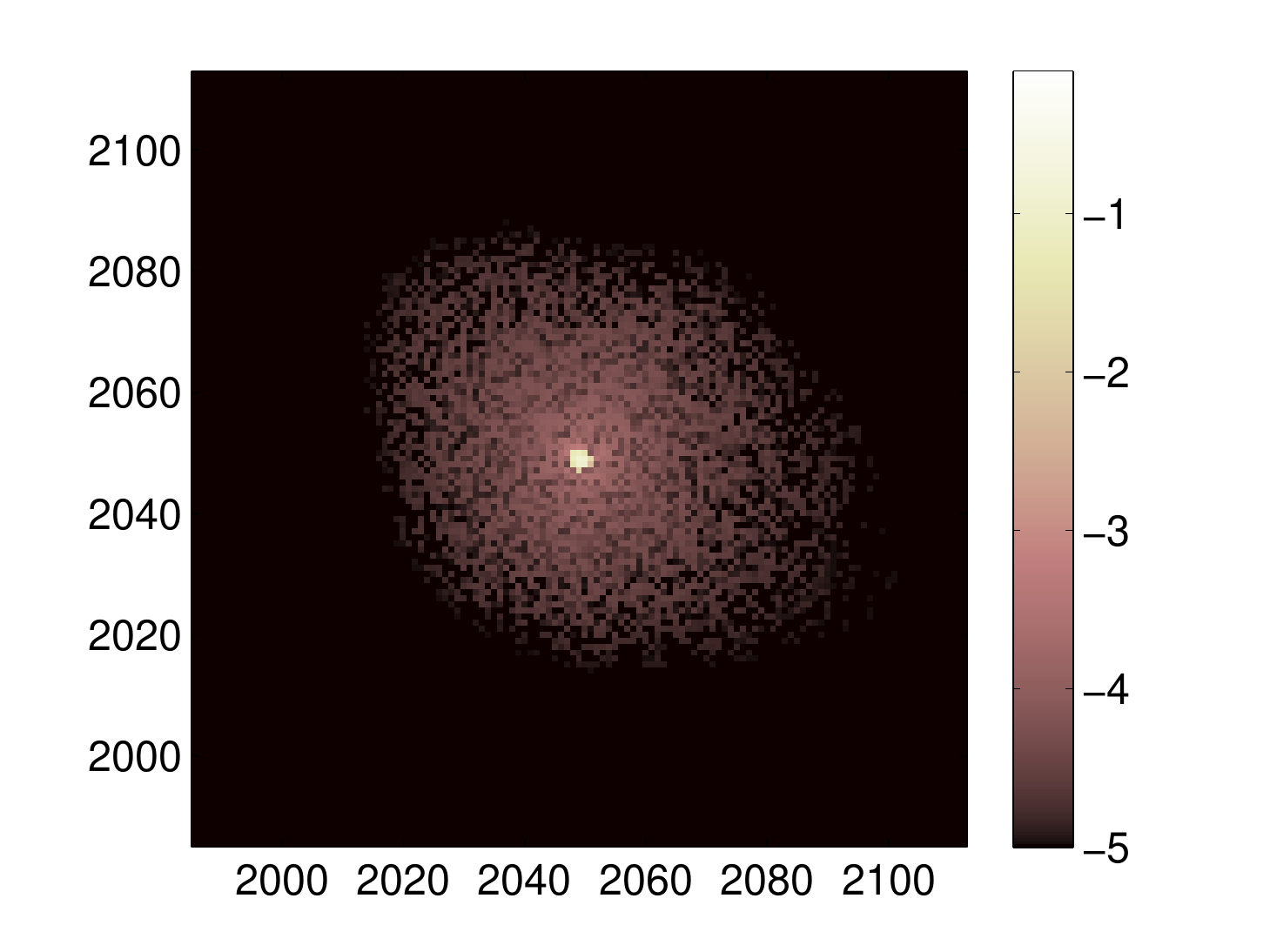} &
		\includegraphics[height=4cm]{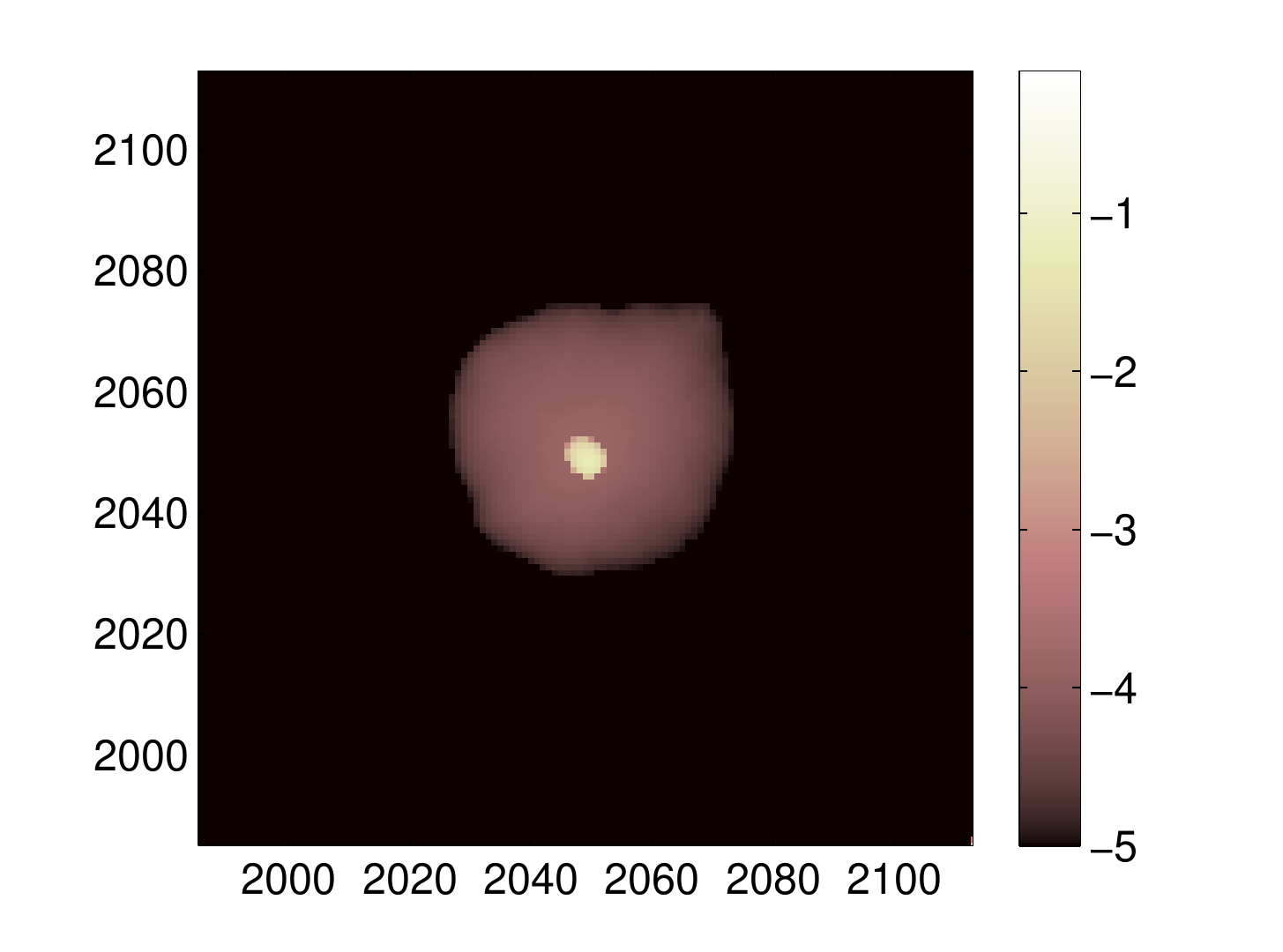} &
		\includegraphics[height=4cm]{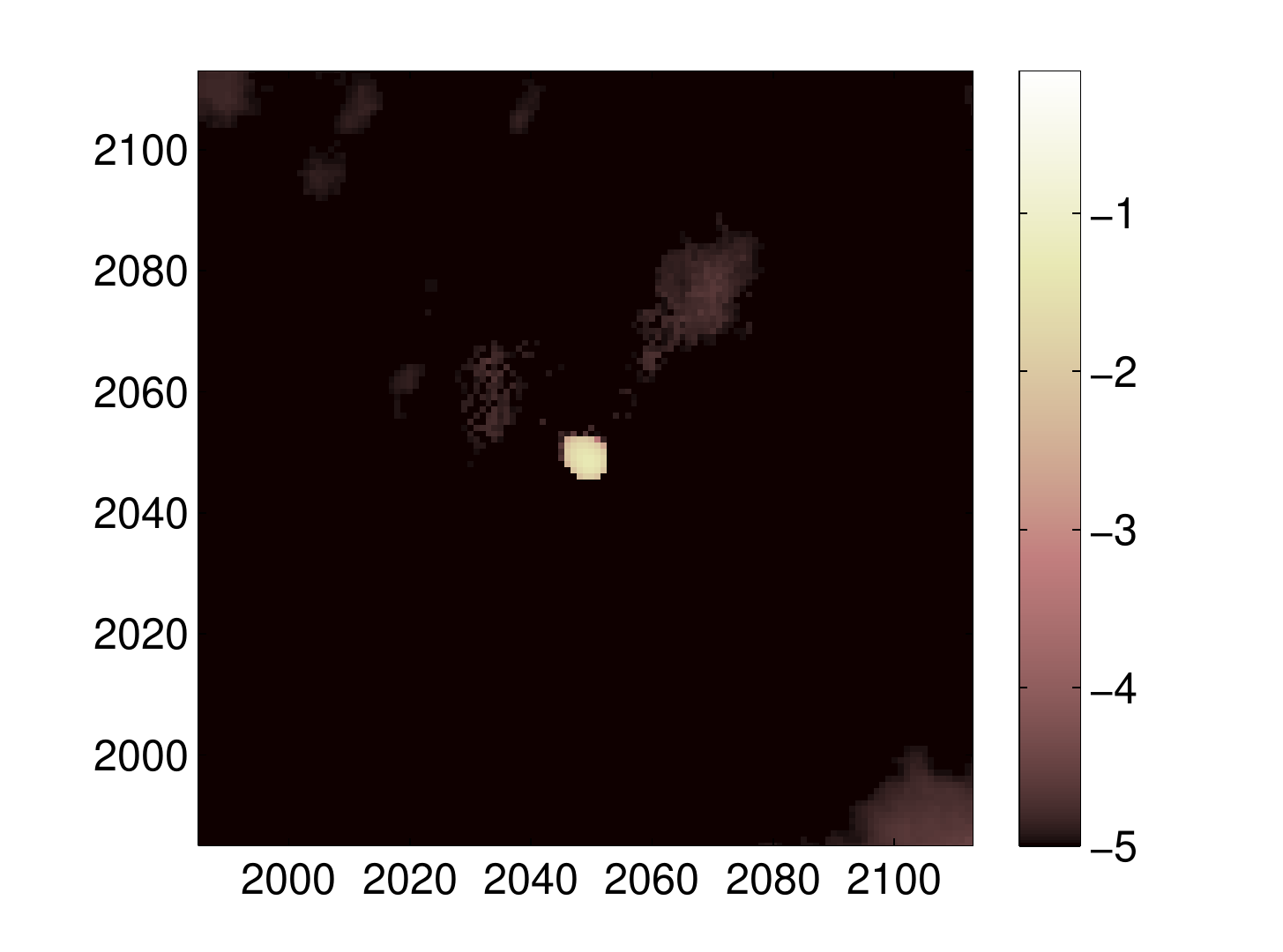} \\
		$\sigma\!=\!\sigma_{RME}$, $\cl M(\bs R) = -0.29$ & $\sigma\!=\!2 \sigma_{RME}$, $\cl M(\bs R) = \textbf{-0.15}$ & $\sigma\!=\!3 \sigma_{RME}$, $\cl M(\bs R) = -0.18$
	\end{tabular}
\caption{Logarithm of the non-parametric filters estimated for the SDO/AIA telescope taken inside the set $\Gamma$. The filters are calculated for several values of $\sigma$ providing different residual whiteness (measured by $\cl M(\bs R)$).}
\label{fig:ExpResults_AIA_Filter_severalSigma}
\end{figure}

Figure~\ref{fig:ExpResults_AIA_Filter_severalSigma} shows the resulting PSFs for different values of $\sigma$. We can observe that the whitest residual is obtained for $\sigma = 2\sigma_{RME} =$ 5.62 DN and we thus select this value for our non-parametric PSF estimate $\bs h_{\rm np}^*$ depicted in Figure~\ref{fig:ExpResults_AIA_Filter_severalSigma}-(center). Let us note that, if the noise is underestimated, we reconstruct part of the noise in the PSF and image, and the resulting PSF is too noisy (see Figure~\ref{fig:ExpResults_AIA_Filter_severalSigma}-(left)). Oppositely, if the noise is overestimated, the algorithm provides the trivial solution, \ie the PSF tends to a discrete delta and the image to the observations (see Figure~\ref{fig:ExpResults_AIA_Filter_severalSigma}-(right)).

\begin{figure}[ht]
\centering
	\begin{tabular}{c c c}
		\includegraphics[height=4cm]{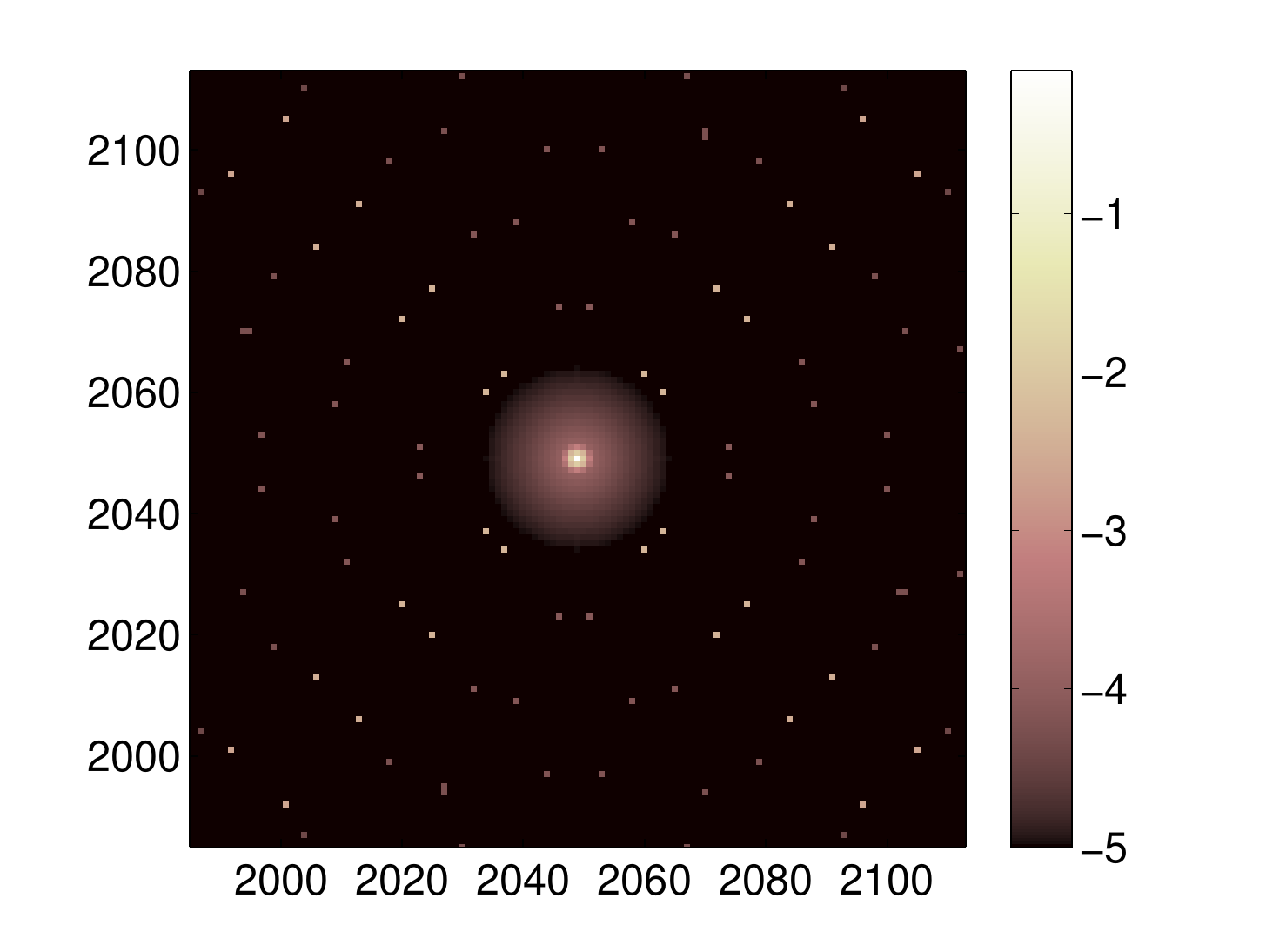} & 
		\includegraphics[height=4cm]{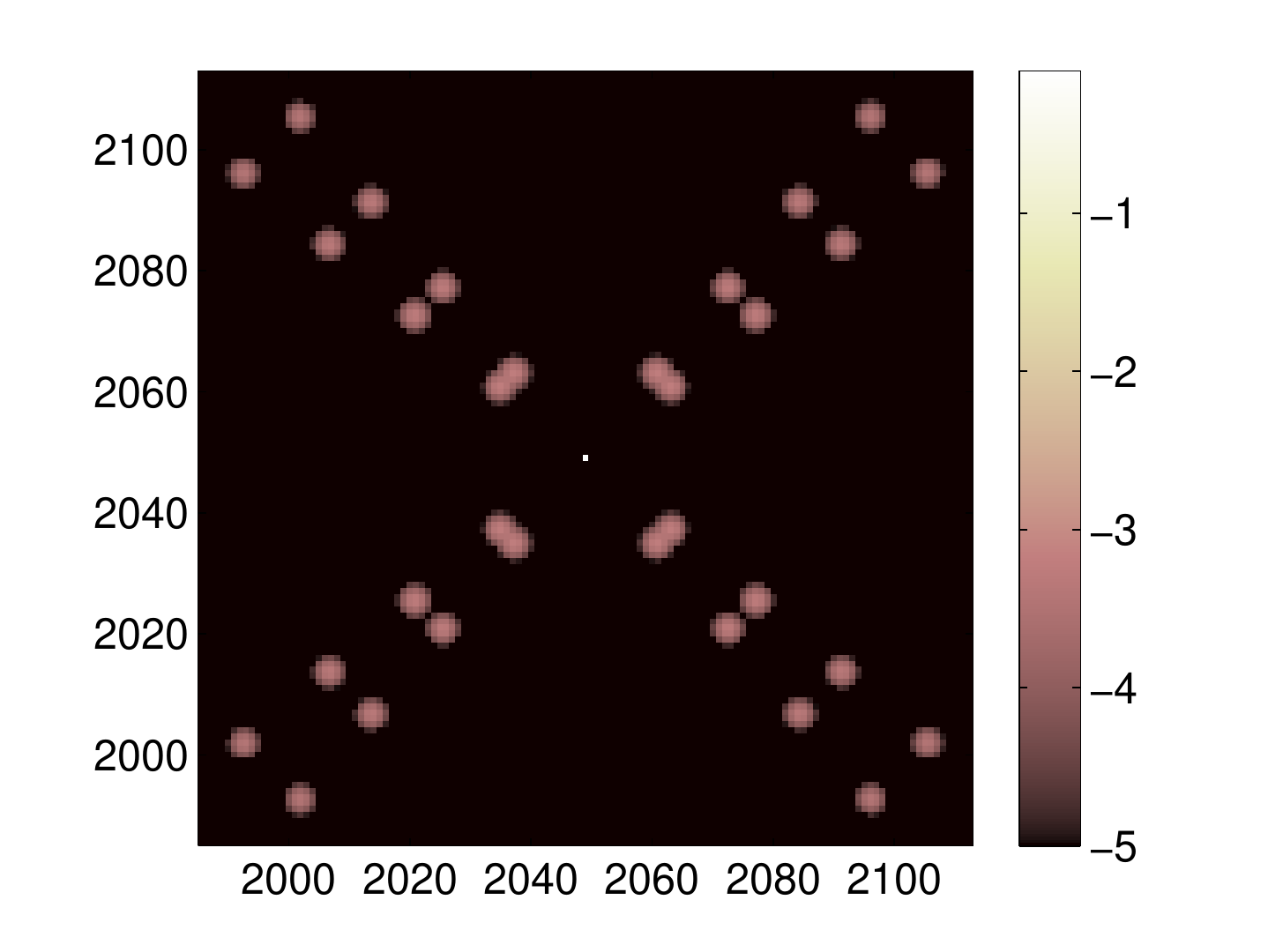} &
		\includegraphics[height=4cm]{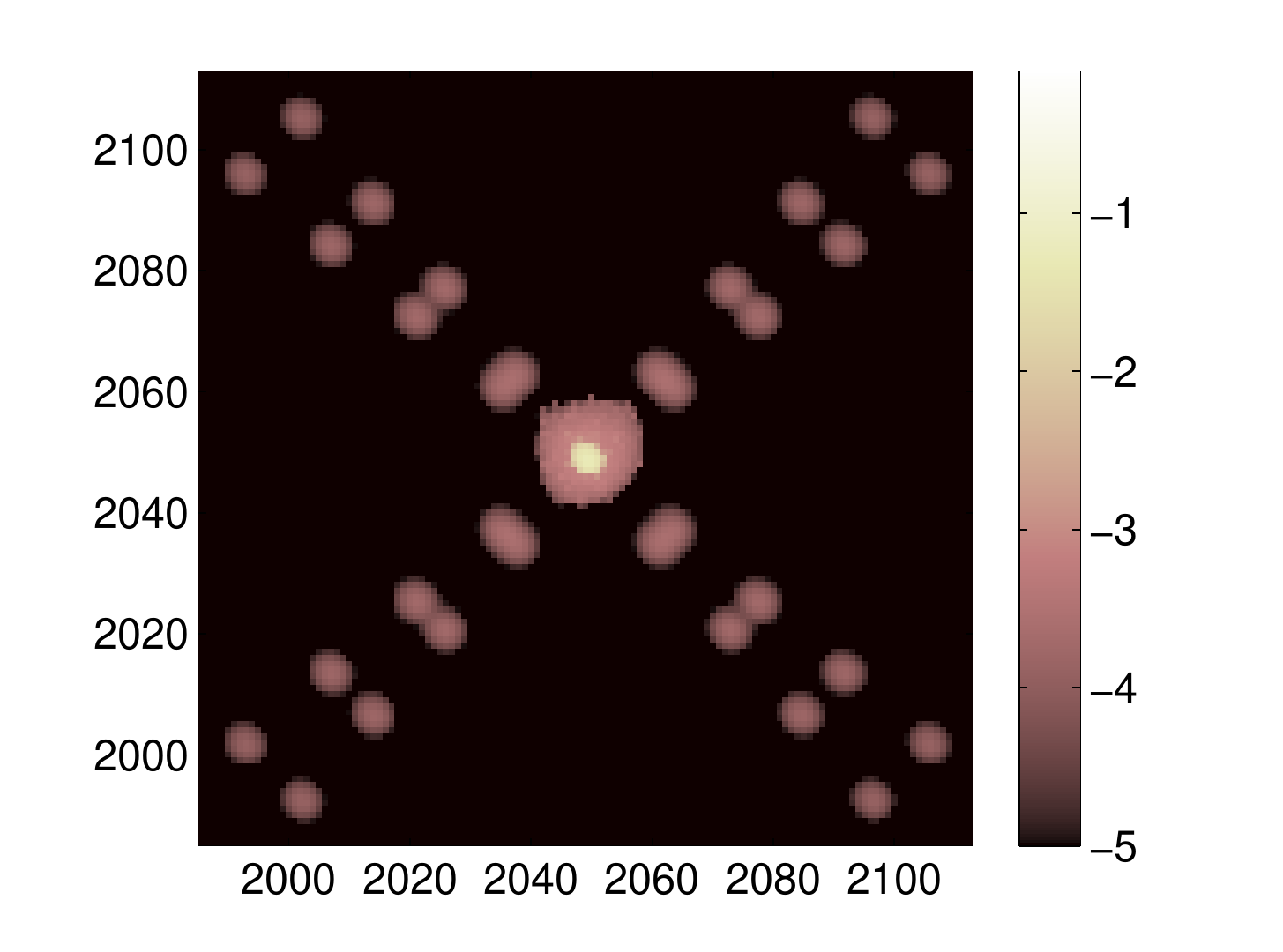}
	\end{tabular}
\caption{Logarithm of filters for the SDO/AIA telescope taken inside the set $\Gamma$. (left) Parametric PSF estimated by Poduval et al.~\cite{2013ApJ...765..144P}, \ie $\bs h_{\rm p_1}$. (center) Diffraction mesh pattern modeled using the parameters estimated by Poduval et al.~\cite{2013ApJ...765..144P}, \ie $\bs h_{\rm p_2}$. (right) Combined parametric/non-parametric filters using the parametric PSF shown on the center, \ie $\bs h^*_{{\rm p_2-np}}$.}
\label{fig:ExpResults_AIA_Filter_Poduval}
\end{figure}

The estimated non-parametric PSF in Figure~\ref{fig:ExpResults_AIA_Filter_severalSigma}-(center) is compared with the parametric PSF estimated by Poduval et al.~\cite{2013ApJ...765..144P}, depicted in Figure~\ref{fig:ExpResults_AIA_Filter_Poduval}-(left). This PSF, denoted $\bs h_{\rm p_1}$, was obtained by fitting a parametric model based on the optical characterization of the telescope. We observe that the obtained non-parametric filter is highly localized in the center of the discrete grid and presents a limited support. Let us note that the observation noise level prevents the estimation of the diffraction peaks present in $\bs h_{\rm p_1}$. These can only be obtained by constraining the shape of the filter in the reconstruction process, as implicitly done by parametric deconvolution methods.

To solve the blind deconvolution problem, the algorithm requires an average of four iterations on the value of $\rho$ and a total time of 8 hours on our computer. In order to reduce the computation time, some functions such as the proximity operators may be computed in parallel, as many of the convex functions on which they are defined are separable~\cite{OPT-003}. The computation time can be further reduced if, for instance, the algorithms are implemented in C instead of MATLAB.

In order to obtain a more accurate PSF estimation containing the diffraction peaks, let us incorporate a parametric PSF inside the acquisition model in (\ref{eq:discrete_model_vector}) by considering a combined parametric/non-parametric PSF defined as $\bs h^*_{\rm p-np} = \bs h_{\rm p} \otimes \bs h^*_{\rm np}$~\cite{2012ApJ...749L...8S, Shearerphdthesis}. The parametric part is obtained by considering only the mesh diffraction components in the PSF estimated by Poduval et al.~\cite{2013ApJ...765..144P}. This modified PSF, denoted $\bs h_{\rm p_2}$, is depicted in Figure~\ref{fig:ExpResults_AIA_Filter_Poduval}-(center). The non-parametric part ($\bs h^*_{\rm np}$) is estimated using the proposed blind deconvolution approach. Figure~\ref{fig:ExpResults_AIA_Filter_Poduval}-(right) presents the resulting parametric/non parametric PSF, \ie $\bs h^*_{\rm p_2-np}$.

We can observe that the resulting parametric/non-parametric PSF is a corrected version of the parametric model from Figure~\ref{fig:ExpResults_AIA_Filter_Poduval}-(center). Its central shape is also more diffused similarly to what is observed in the parametric PSF estimated by Poduval et al.~\cite{2013ApJ...765..144P}.  

\begin{figure}[ht]
\centering
	\begin{tabular}{c c c}
		\includegraphics[height=4cm]{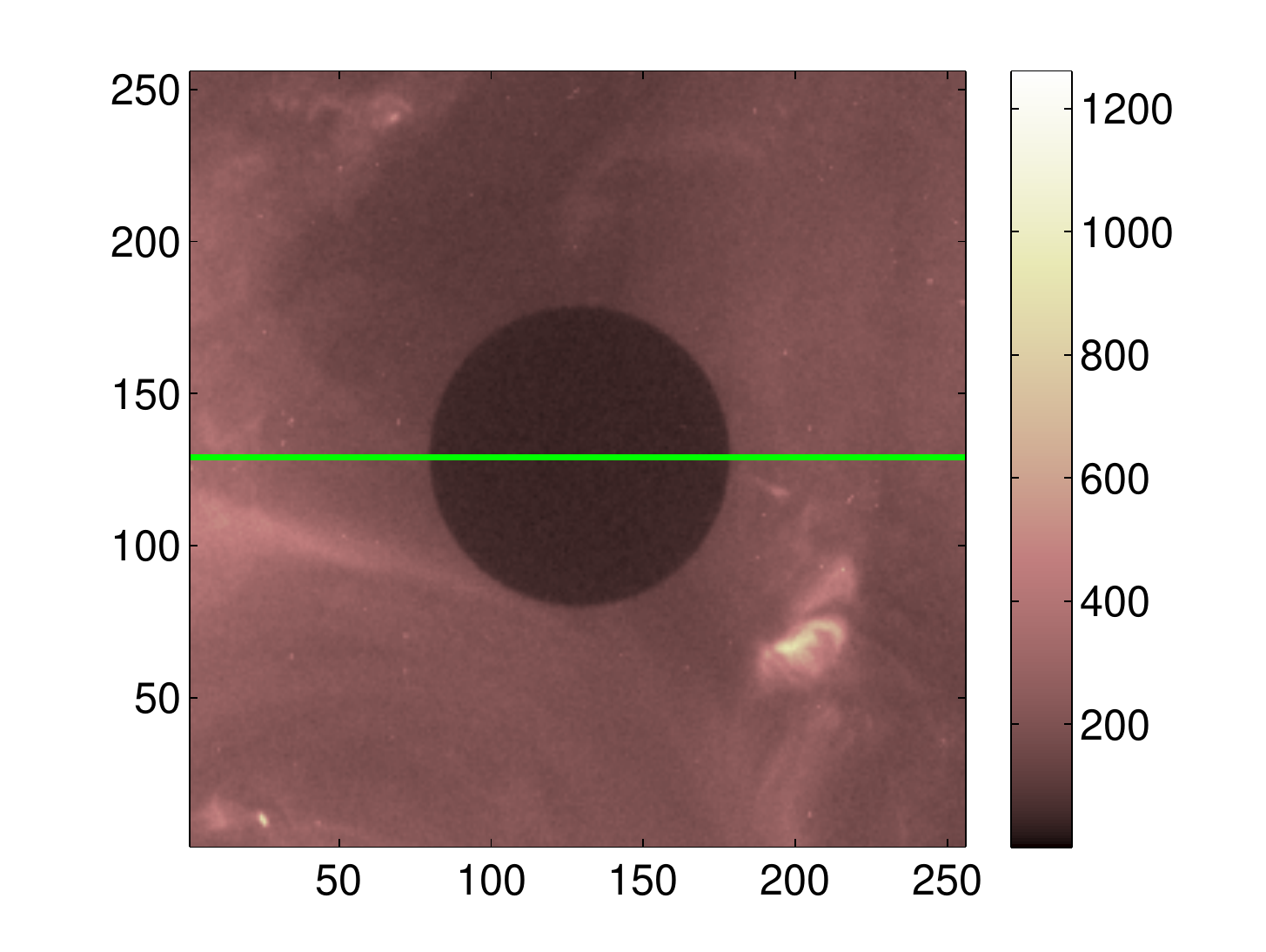} &
		\includegraphics[height=4cm]{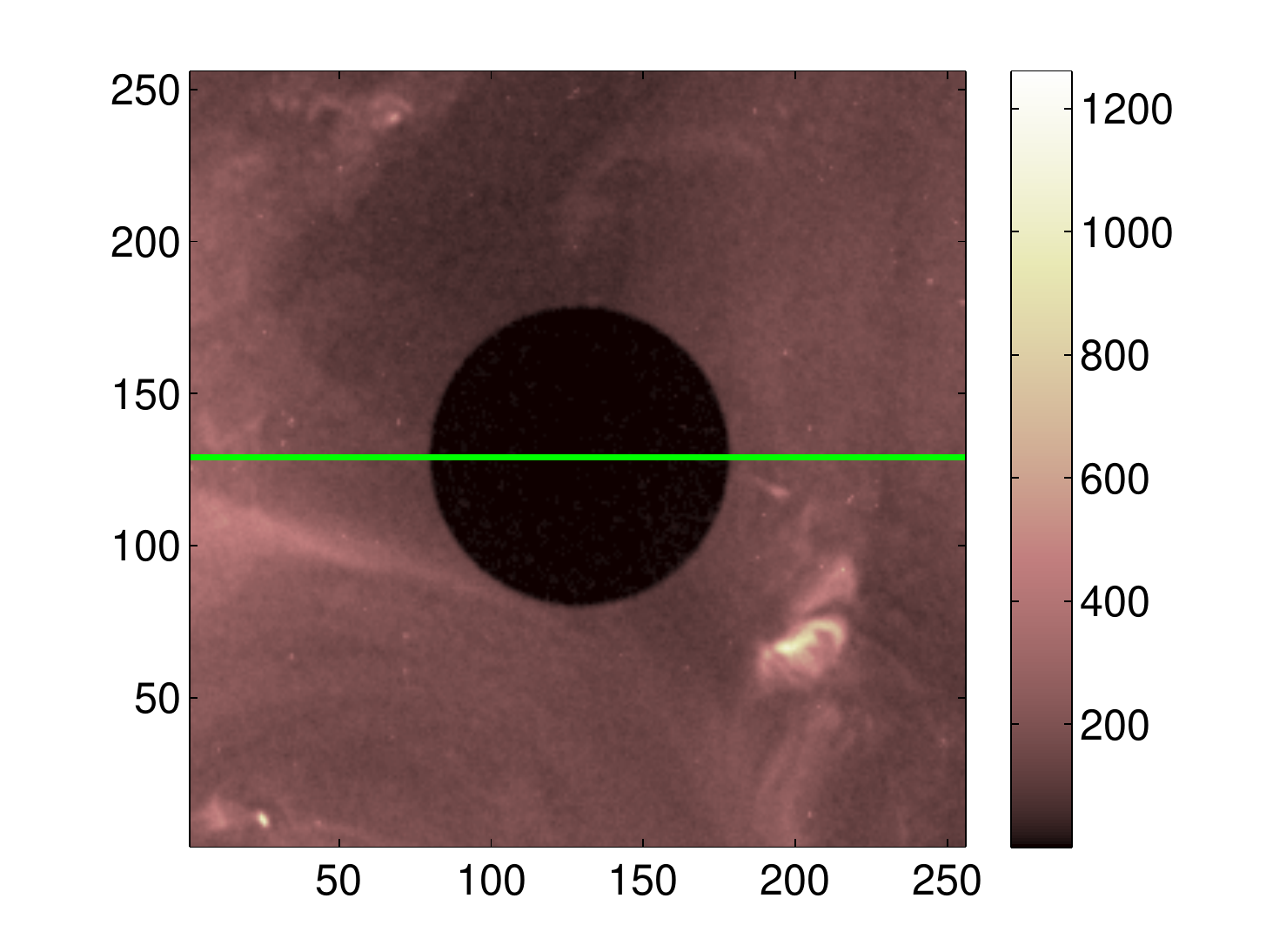} &
		\includegraphics[height=4cm]{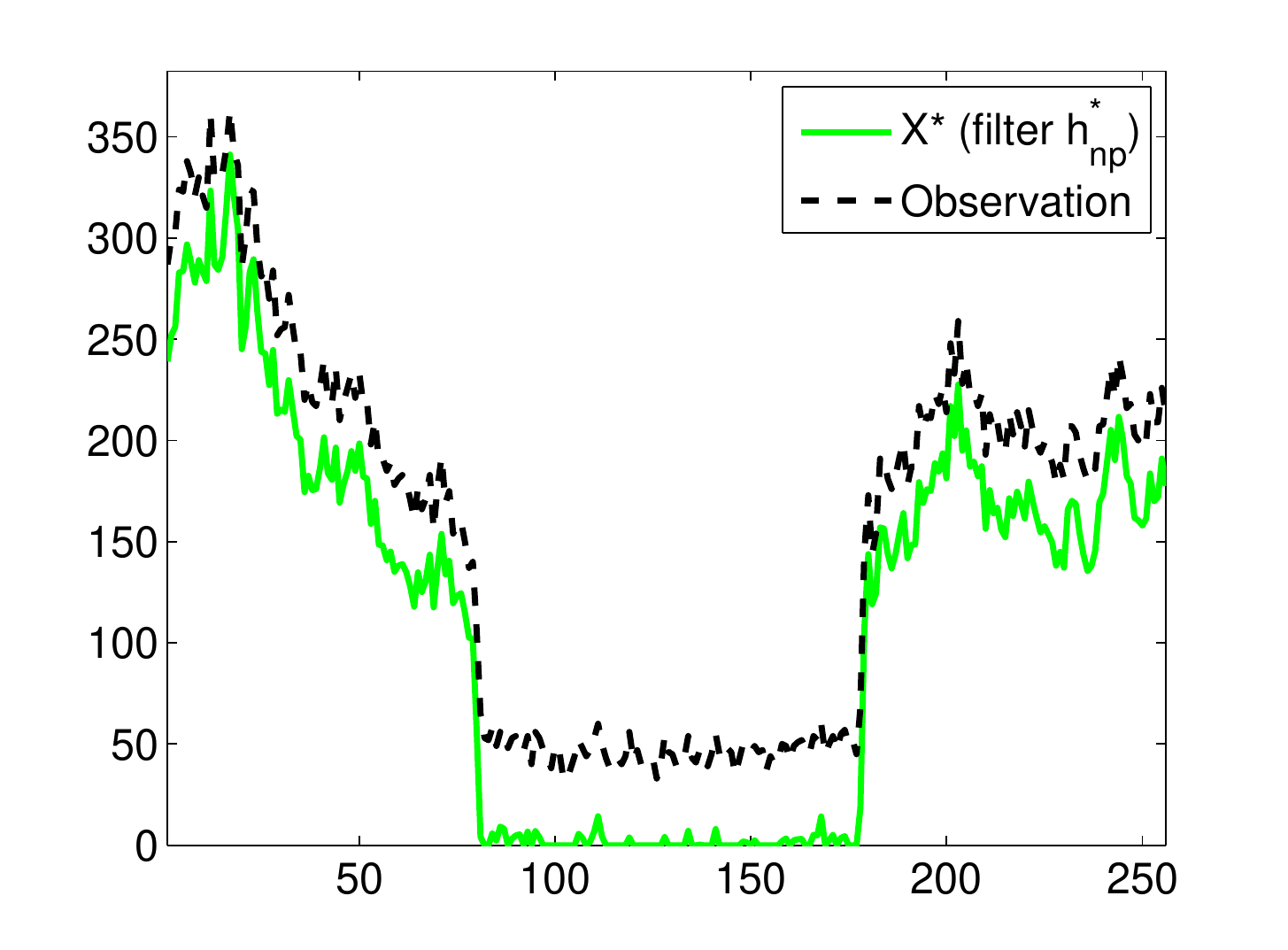}
	\end{tabular} 
\caption{Image reconstruction results for the Venus transit observed by SDO/AIA at 00:02 UT on June 6th 2012. Results are shown for the non-parametric filter, \ie $\bs h_{\rm np}^*$. (left) Observed image, (center) 2-D reconstruction result, and (right, best viewed in color) 1-D profile along $y = 129$. The figures on the left and on the center contain a green line indicating the 1-D profiles shown on the right.}
 \label{fig:ExpResults_Venus_Image}
\end{figure}

\begin{figure}[ht]
\centering
	\begin{tabular}{c c c}
		\includegraphics[height=4cm]{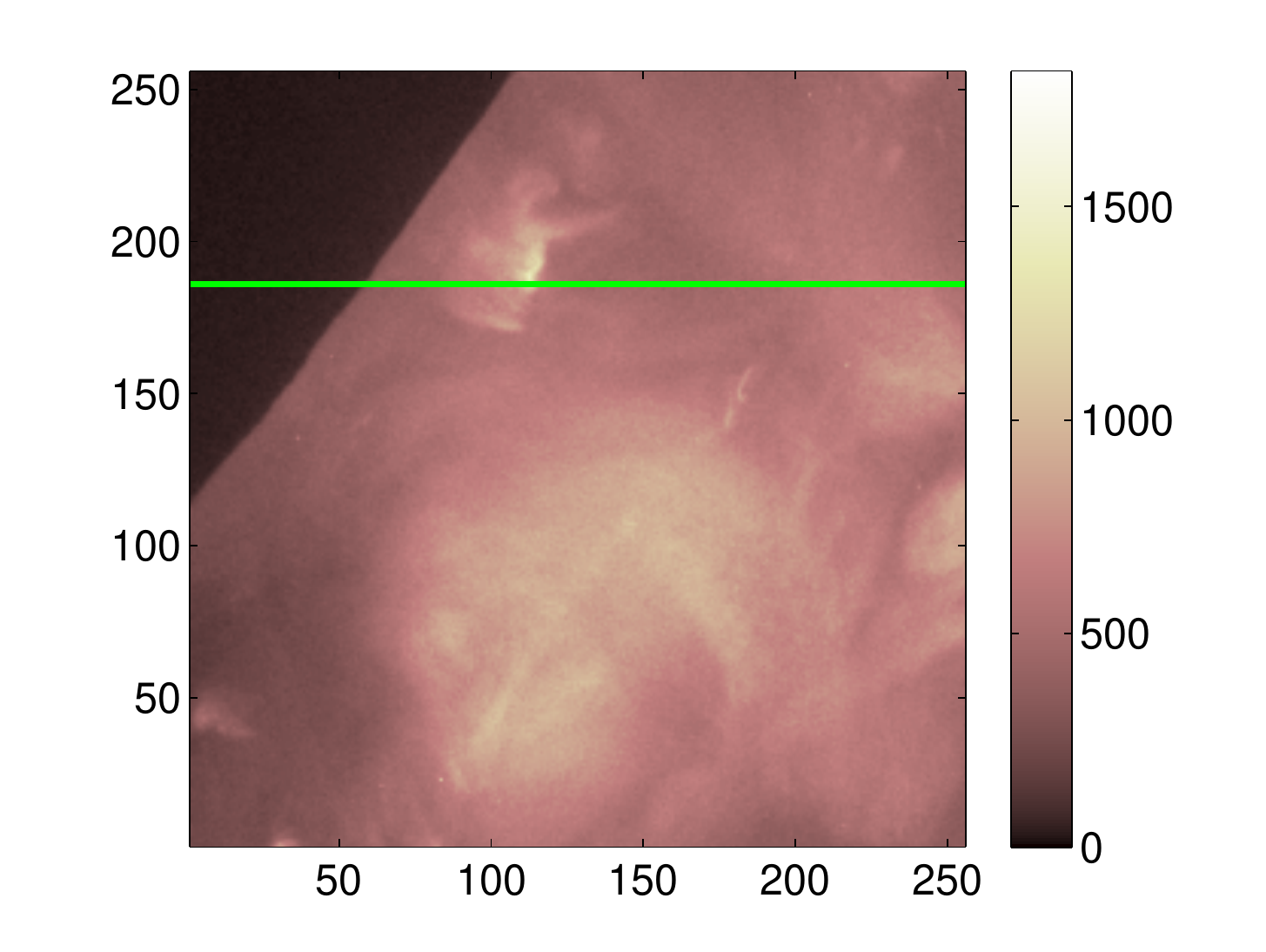} &
		\includegraphics[height=4cm]{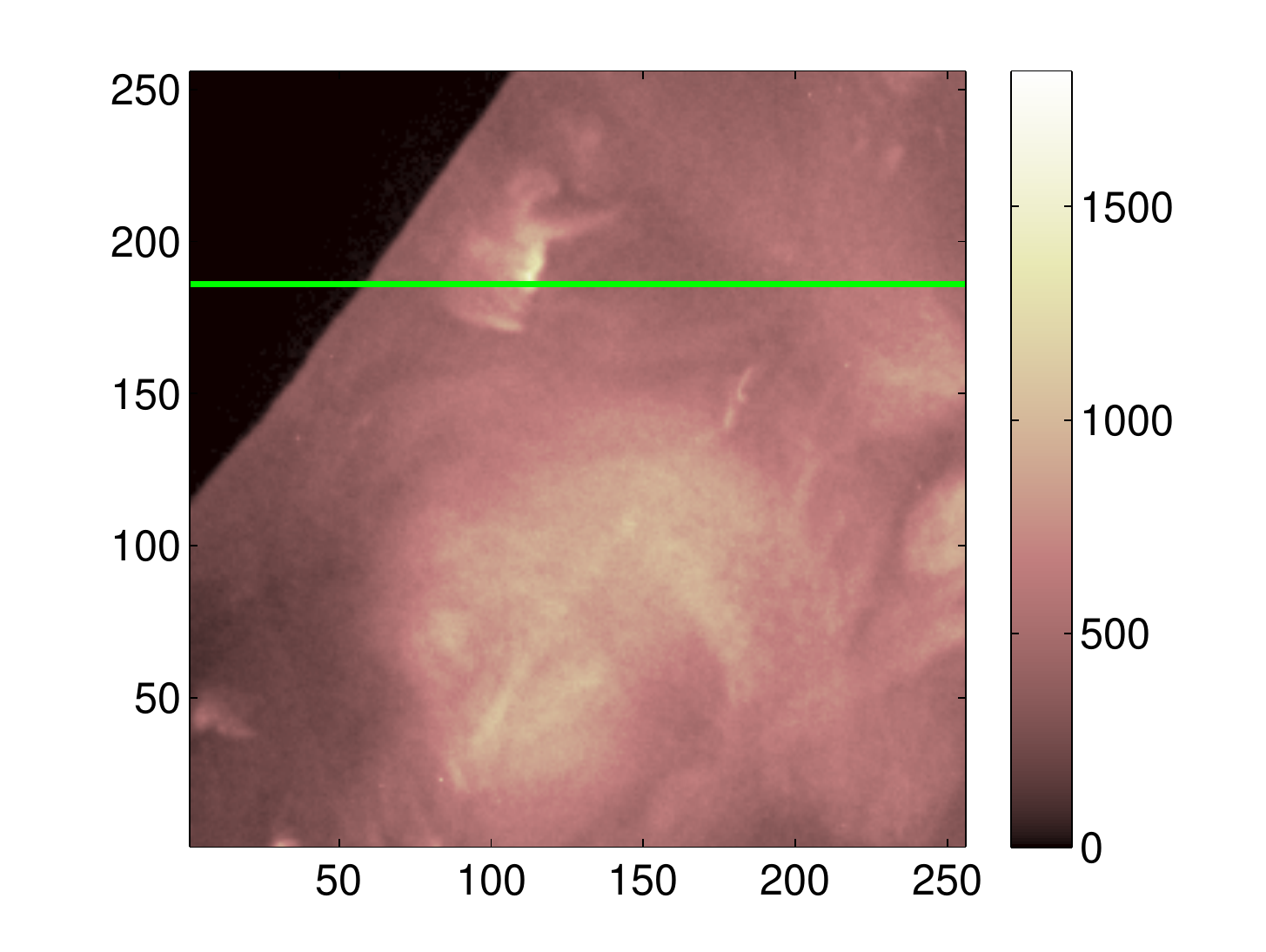} &
		\includegraphics[height=4cm]{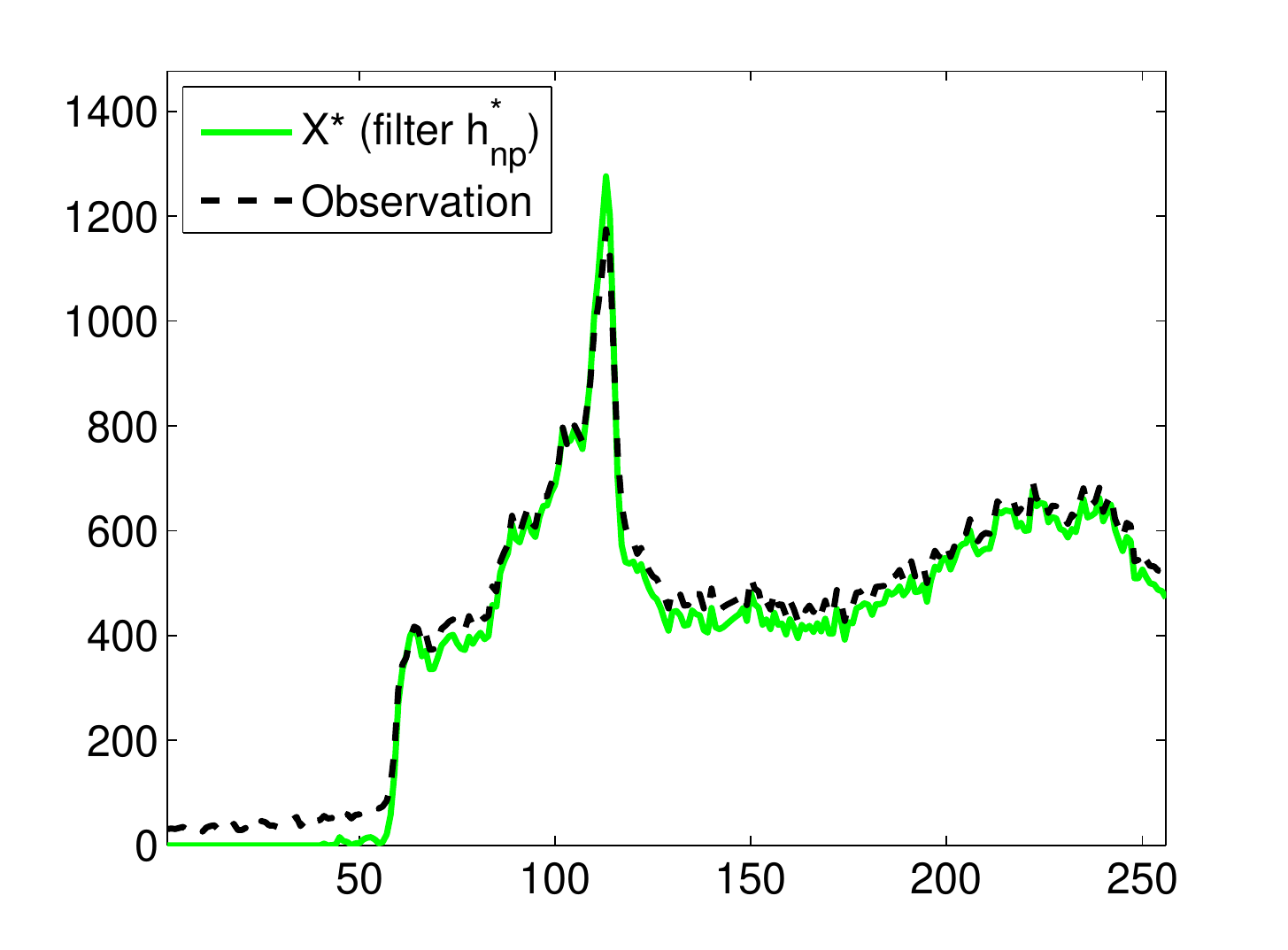} 
	\end{tabular}
\caption{Image reconstruction results for the Moon transit observed by SDO/AIA at 13:00 UT on March 4th 2011. Results are shown for the non-parametric filter, \ie $\bs h_{\rm np}^*$. (left) Observed image, (center) 2-D reconstruction result, and (right, best viewed in color) 1-D profile along $y = 186$. The figures on the left and on the center contain a green line indicating the 1-D profiles shown on the right.}
 \label{fig:ExpResults_MoonAIA_Image}
\end{figure}

The estimated filters from Figure~\ref{fig:ExpResults_AIA_Filter_severalSigma}-(center), Figure~\ref{fig:ExpResults_AIA_Filter_Poduval}-(left) and Figure~\ref{fig:ExpResults_AIA_Filter_Poduval}-(right) were validated using the non-blind deconvolution formulation from Section~\ref{sec:NBID}. 
Notice that, similarly to what has been done for the blind deconvolution, the value of $\rho$ has been selected adaptively by optimizing the residual whiteness defined in~(\ref{eq:MR}). We also observed empirically that the value of $\rho$ must be kept smaller than the one obtained in Algorithm~\ref{alg:alternatemin_ext}. The validation was done on transit images using the modified observations with $\mu = 43.3$ DN. A first validation was performed on the Venus transit image taken by SDO/AIA at 00:02 UT on June 6th 2012 (see Figure~\ref{fig:ExpResults_Venus_Image}-(left)), a patch that has not been used before for estimating $\bs h^*$. A second validation of the filters was done on the Moon transit image taken by SDO/AIA at 13:00 UT on March 4th 2011 (see Figure~\ref{fig:ExpResults_MoonAIA_Image}-(left)). 

We quantify the apparent disk emissions by summing the pixel values inside the disk. Table~\ref{tab:ExpResults_AIA_DiskIntensity} displays the disk intensity ratio for the images deconvolved using the following filters: (1) the parametric PSF estimated by Poduval et al.~\cite{2013ApJ...765..144P}, \ie $\bs h_{\rm p_1}$; (2) the non-parametric PSF of Figure~\ref{fig:ExpResults_AIA_Filter_severalSigma}-(center), \ie $\bs h_{\rm np}^*$; and (3) the parametric/non-parametric PSF of Figure~\ref{fig:ExpResults_AIA_Filter_Poduval}-(right), \ie $\bs h_{\rm p_2-np}^*$.

\begin{table}[ht]
	\centering
	\begin{tabular}{ c  c  c }
		\hline \hline
		Filter & $\cl S_X/ \cl S_Y$ Venus & $\cl S_X/ \cl S_Y$ Moon \\
		\hline
		$\bs h_{\rm p_1}$ & $0.74 \times 10^{-2}$ & $1.30 \times 10^{-2}$ \\
		\hline
		$\bs h_{\rm np}^*$ & $3.65 \times 10^{-2}$ & $3.64 \times 10^{-2}$ \\ 
		\hline
		$\bs h^*_{{\rm p_2-np}}$ & $0.28 \times 10^{-2}$ & $0.65 \times 10^{-2}$ \\
		\hline
	\end{tabular}
	\caption{Disk intensity ratio for the different filters.}
  \label{tab:ExpResults_AIA_DiskIntensity}
\end{table}

We observe that, for both transits, the parametric/non-parametric model reaches lower disk apparent emissions. We also note that, compared to the non-parametric filter estimation, the parametric model provide slightly better results in terms of reducing the disk apparent emissions. However, the non-parametric scheme presents the advantage of being generally applicable for any optical instrument without the need of an exact model of the imaging process.

Figures~\ref{fig:ExpResults_Venus_Image}~and~\ref{fig:ExpResults_MoonAIA_Image} illustrate the image reconstruction results when using the non-parametric PSF, \ie $\bs h^*_{\rm np}$. The figure on the left presents the observed patch, the one on the center depicts the \mbox{2-D} estimated image and the one on the right presents a 1-D profile that allows the observation of the disk of Venus and the Moon, respectively. Due to lack of space, only the non-parametric PSF validation is illustrated.

Let us note that, if the long-range effect $\bs \mu$ is not removed from the observations, the parametric PSFs taken with a larger support of $2048 \times 2048$ pixels are not able to eliminate the offset and, hence, are not able to remove the apparent emissions inside the disk of Venus.

\begin{figure}[ht]
\centering
	\begin{tabular}{c c c}
		\includegraphics[height=4cm]{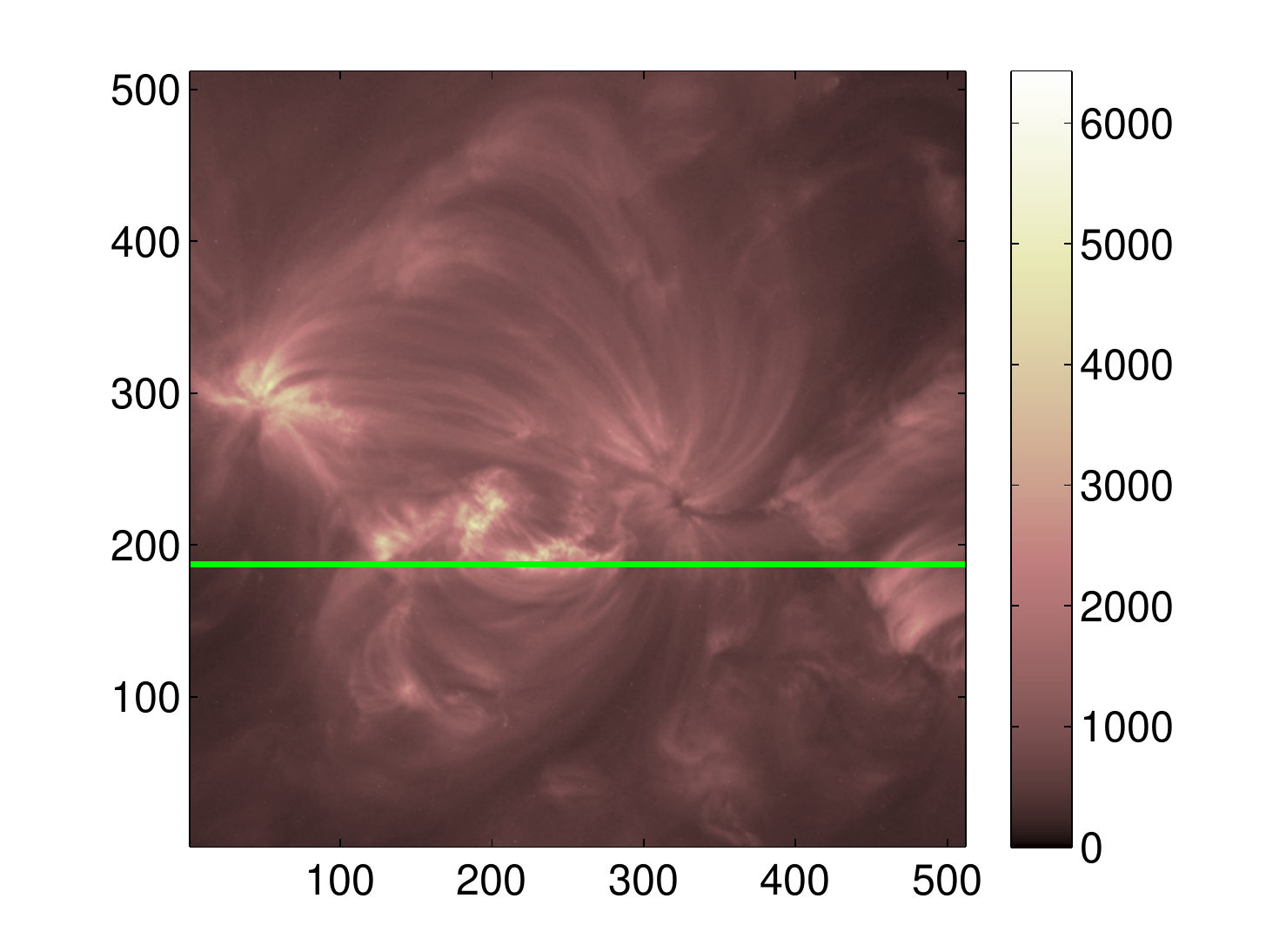} &
		\includegraphics[height=4cm]{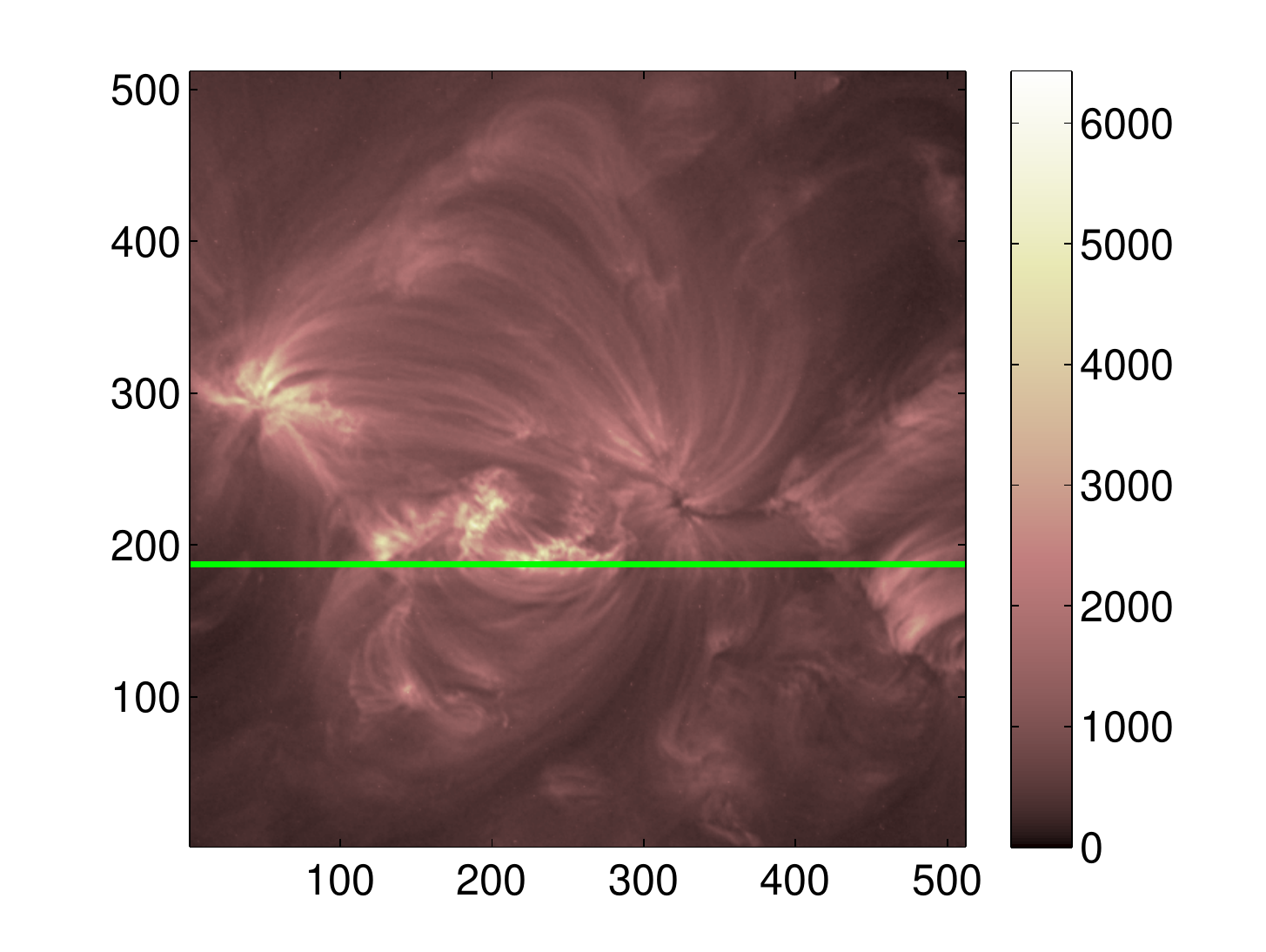} &
		\includegraphics[height=4cm]{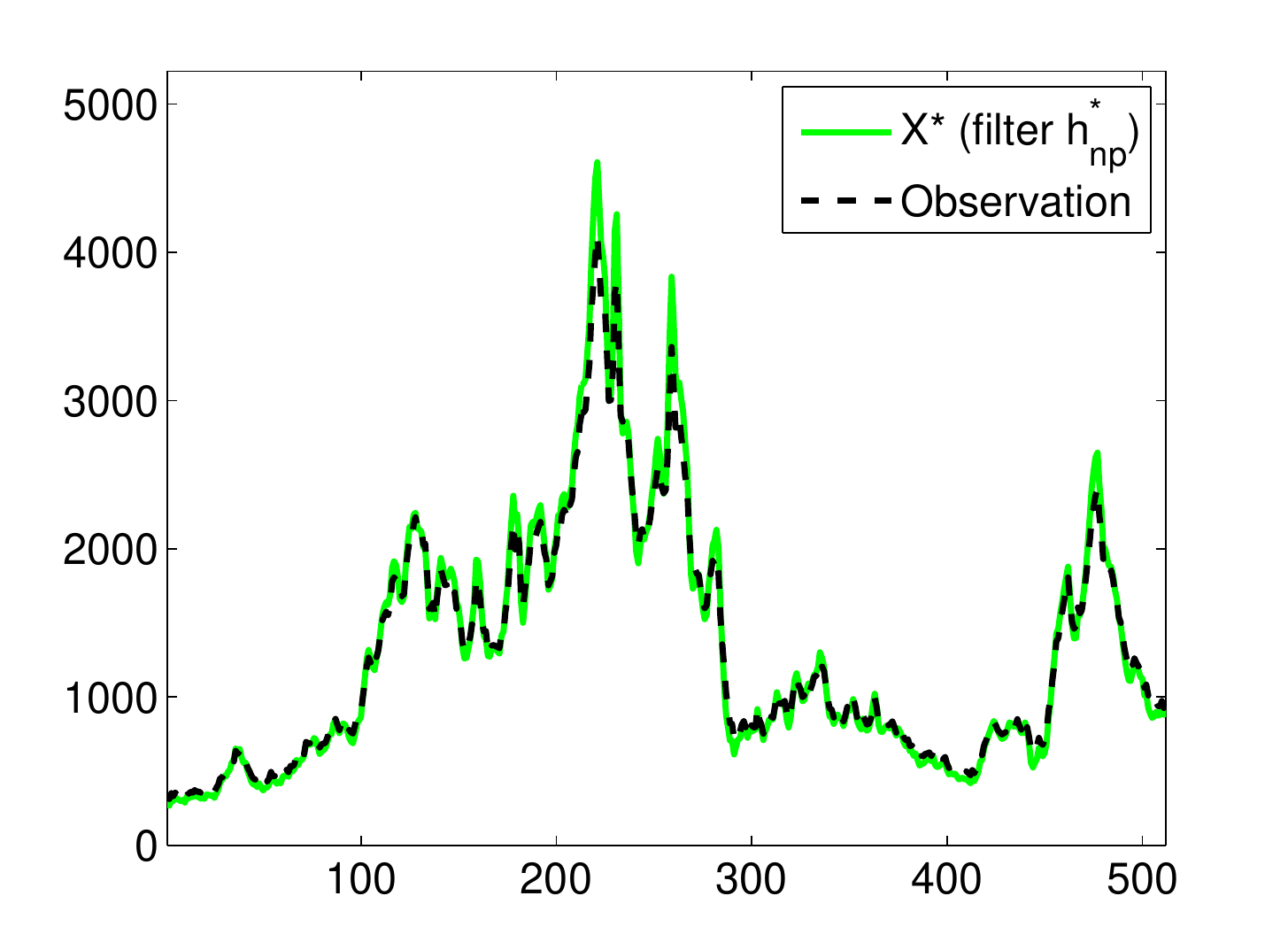}
	\end{tabular}
\caption{Reconstruction results for a SDO/AIA image containing an active region. The image has been captured at 10:00 UT on August 8th 2011. Results are shown for the non-parametric filter, \ie $\bs h_{\rm np}^*$. (left) Observed image, (center) 2-D reconstruction result, and (right, best viewed in color) 1-D profile along $y = 187$. The figures on the left and on the center contain a green line indicating the 1-D profiles shown on the right.}
 \label{fig:ExpResults_AR}
\end{figure}

Finally, the estimated filters were also used to deconvolve non-transit images. For this, the non-blind deconvolution formulation from Section~\ref{sec:NBID}, using an adaptive $\rho$, was applied to the modified observations with $\mu = 43.3$ DN. The results are shown for a non-transit image taken by SDO/AIA at 10:00 UT on August 8th 2011. We selected a portion of the original image of $512 \times 512$ pixels around the active region (see Figure~\ref{fig:ExpResults_AR}-(left)). Figure~\ref{fig:ExpResults_AR}-(center) depicts the 2-D estimated image using the non-parametric PSF, \ie $\bs h_{\rm np}^*$, and Figure~\ref{fig:ExpResults_AR}-(right) shows a 1-D profile along $y=187$.

\begin{figure}[ht]
\centering
\includegraphics[height=3.4cm]{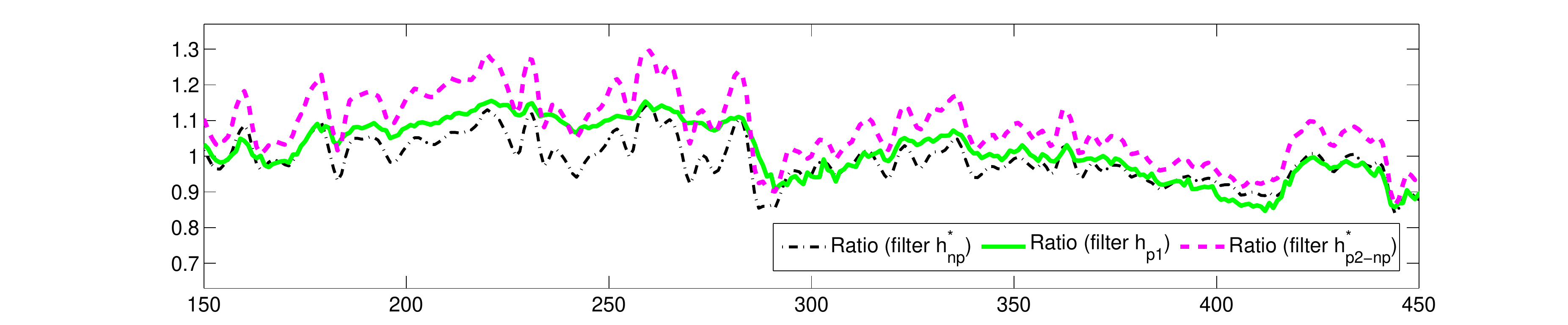}
\caption{Reconstruction results for the image in Figure~\ref{fig:ExpResults_AR}-(left). The figure (best viewed in color) shows the ratio (computed using a pixel by pixel division) between the deconvolved and the observed images along $y=187$, for the different considered filters: $\bs h_{\rm np}^*$, $\bs h_{\rm p_1}$ and $\bs h_{\rm p_2-np}^*$. The horizontal axis is in correspondence with the horizontal axis of Figure~\ref{fig:ExpResults_AR}-(right).}
 \label{fig:ExpResults_AR_Ratio}
\end{figure}

We observe that the non-parametric PSF enhances the image, providing brighter active regions and coronal loops, and darker regions of lower intensity than in the original image. To compare those results with the other filters mentioned above, we take the ratio between the deconvolved images and the observation (computing a pixel by pixel division), as shown in Figure~\ref{fig:ExpResults_AR_Ratio}. We notice that the deconvolution resulting from the combined parametric/non-parametric PSF presents a higher correction from the observations than the ones obtained using the other filters. In dark regions, the non-parametric PSF provides results similar to those obtained by the parametric PSF, however, in brighter areas the non-parametric PSF seems to be able to recover more details.

\subsubsection{SECCHI/EUVI - Moon transit}

We consider three $2048 \times 2048$ images from the transit ($P = 3$) recorded by the 17.1~nm channel of EUVI. The images are calibrated with the {\tt secchi\_prep.pro} procedure available within the IDL \emph{SolarSoft} library. Following the practice described for the SDO/AIA Venus transit, each image was cropped using a $512 \times 512$ window ($N = 512^2$) centered around the Moon disk. The filter is assumed to have a limited support inside a $129\!\times\!129$ pixel grid ($b = 64$). This allows one to obtain the core of the PSF of around $100\!\times\!100$ pixels, as observed by Shearer et al.~\cite{2012ApJ...749L...8S}, and encompass 99\% of the energy of previously estimated PSFs.

The long-range PSF is modeled by a constant $\mu = 12.5$ DN, estimated by computing the mean intensity value (over five patches) on a disk of radius 10 pixels inside the Moon disk. The estimated noise variance using the RME provided $\sigma_{RME}^2~=~2.21~\textrm{ DN}^2$. Following the same procedure as described before for SDO/AIA, we obtained that the value of $\sigma$ that allows obtaining the whitest residual is $\sigma = 2 \ \sigma_{RME}$ = 4.04 DN.

\begin{figure}[ht]
\centering
	\begin{tabular}{c c c}
		\includegraphics[height=4cm]{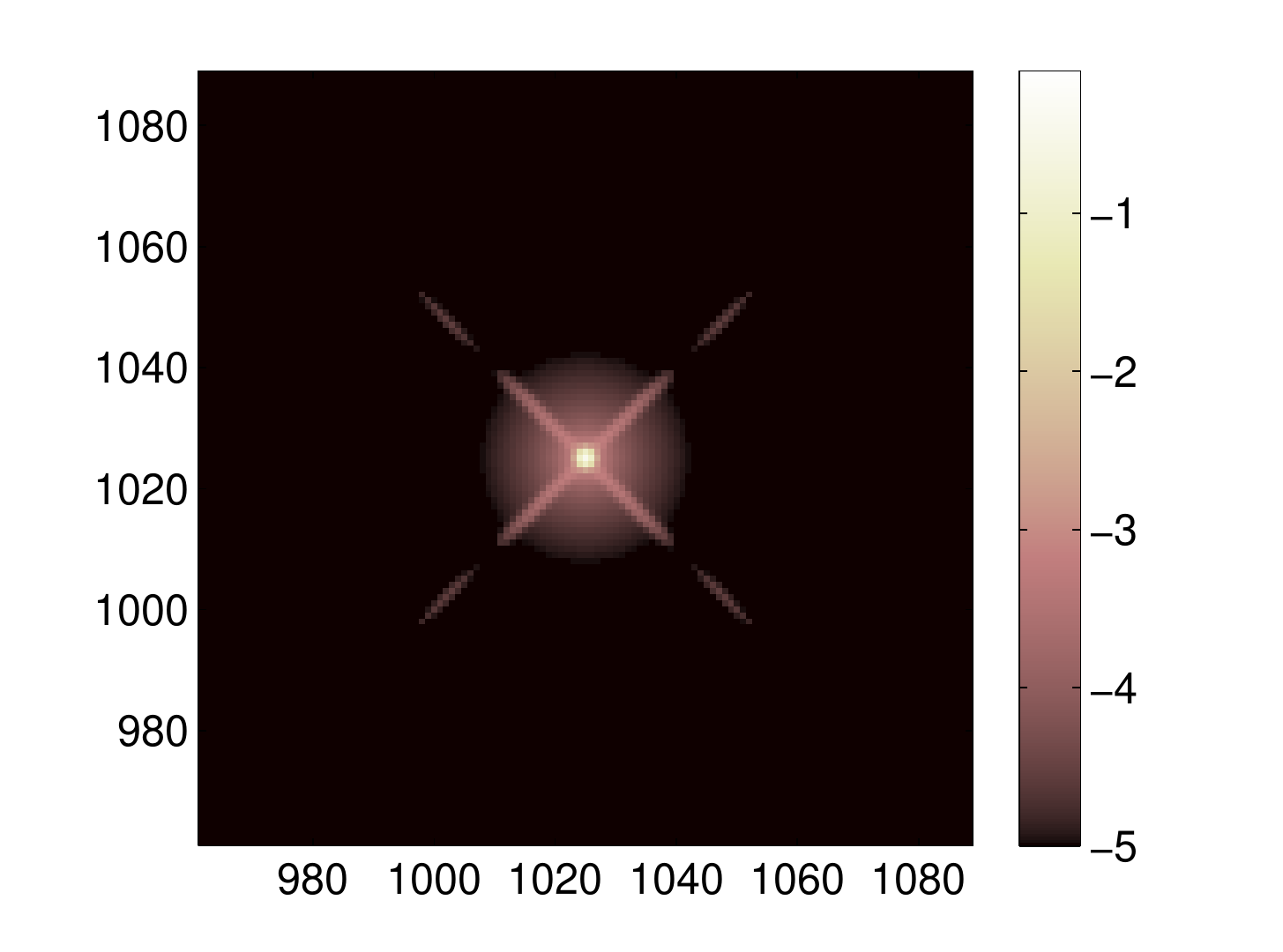} &
		\includegraphics[height=4cm]{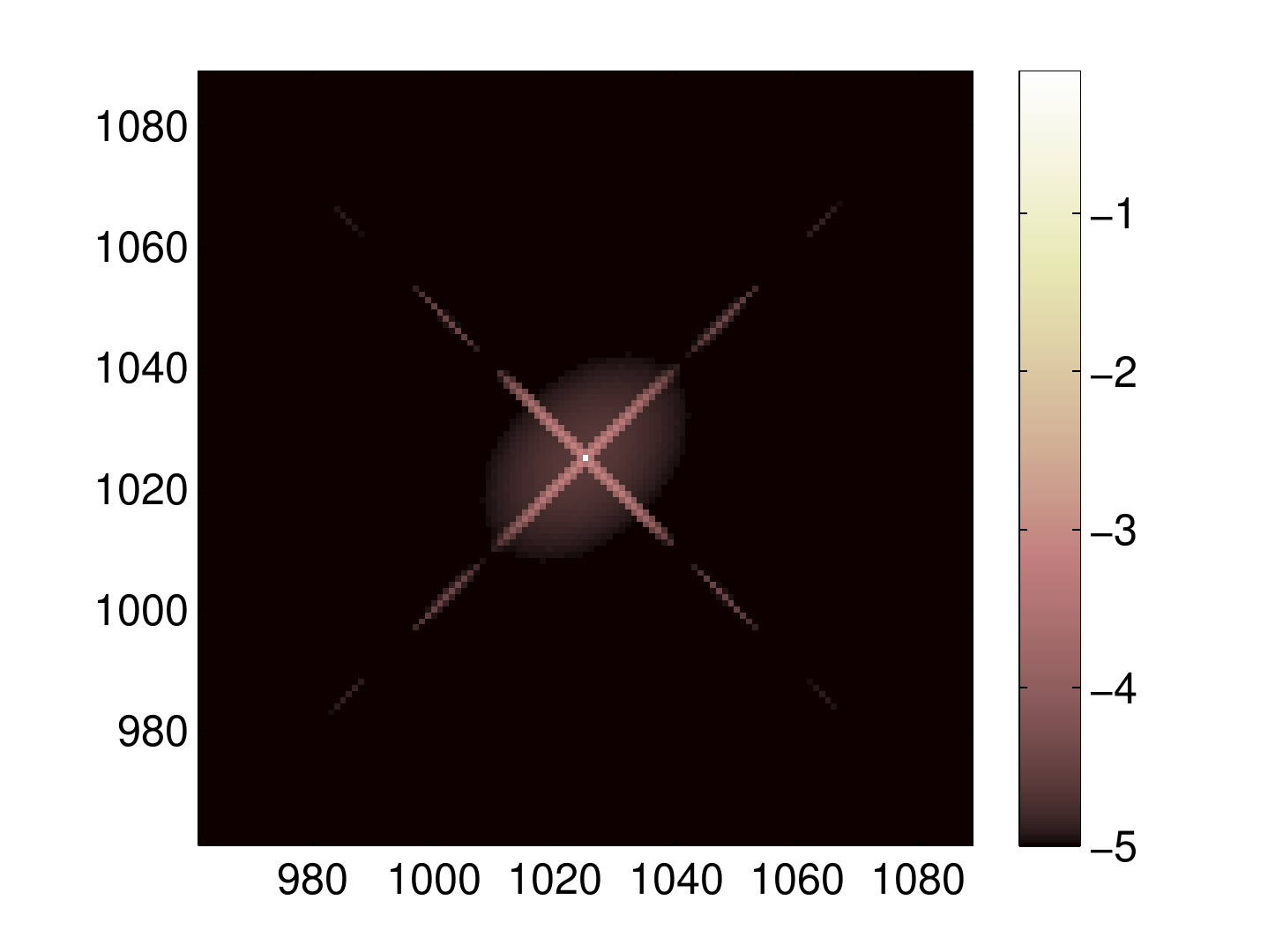} &
		\includegraphics[height=4cm]{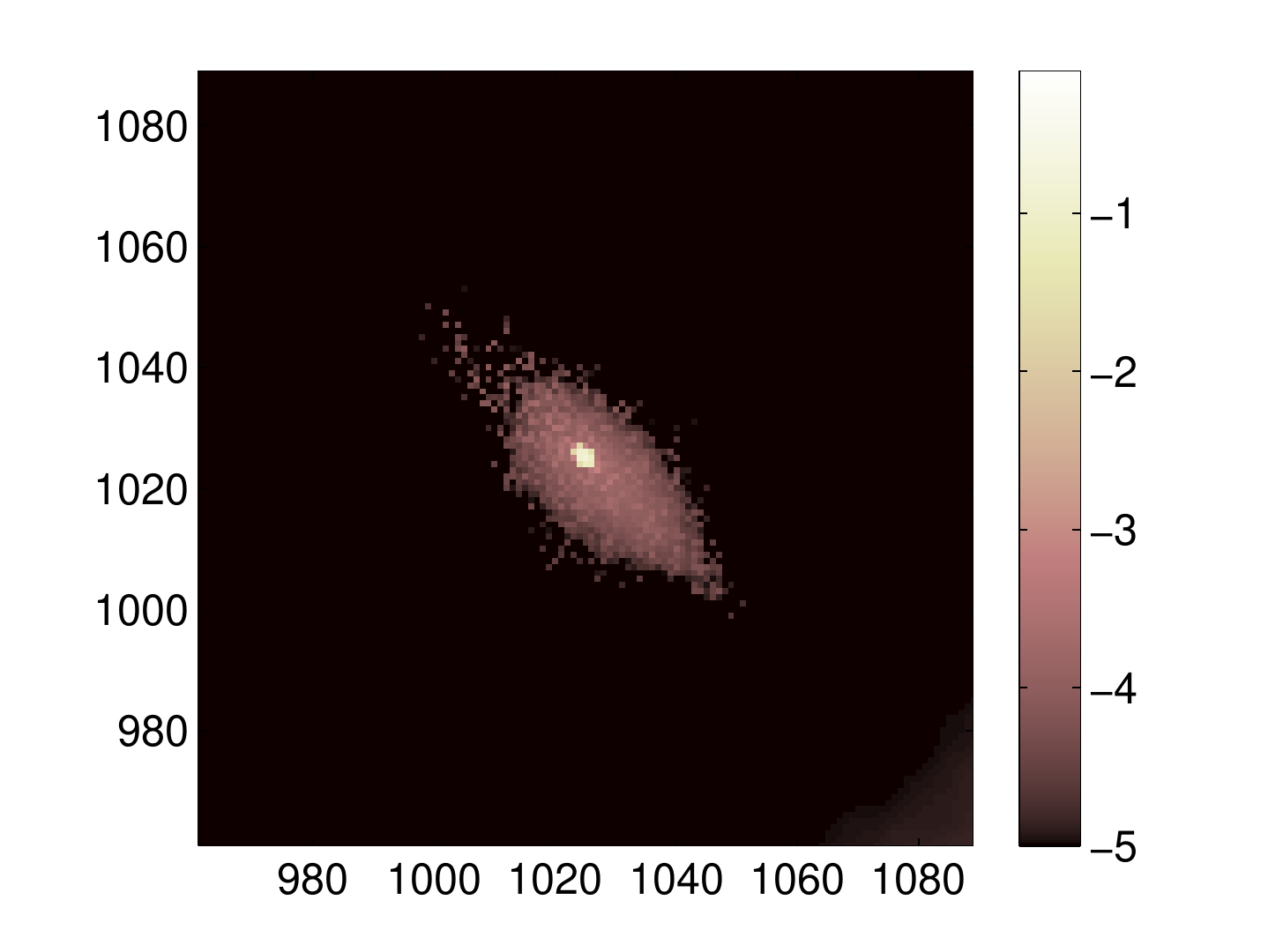} 
	\end{tabular}
\caption{Logarithm of the filters for the SECCHI/EUVI telescope taken inside the set $\Gamma$. (left) Parametric PSF given by the {\tt euvi\_psf.pro} procedure of \emph{SolarSoft}, \ie $\bs h_{{\rm p_1}}$. (center) Parametric PSF given by the {\tt euvi\_deconvolve.pro} procedure of \emph{SolarSoft}, \ie $\bs h_{{\rm p_2}}$. (right) Non-parametric PSF, \ie $\bs h_{\rm np}^*$.}
\label{fig:ExpResults_EUVI_Filter}
\end{figure}

\begin{figure}[ht]
\centering
	\begin{tabular}{c c}
		\includegraphics[height=4cm]{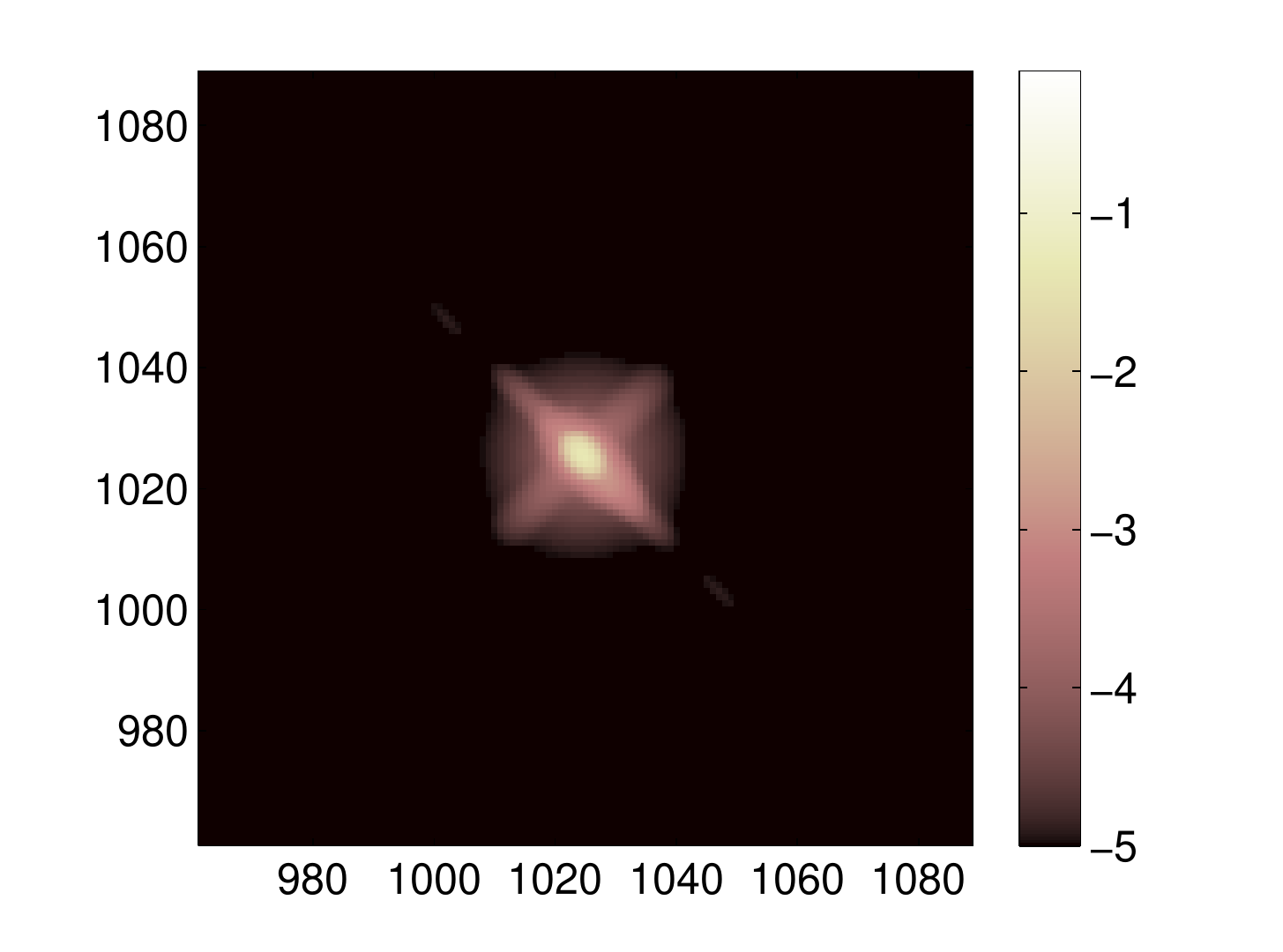} &
		\includegraphics[height=4cm]{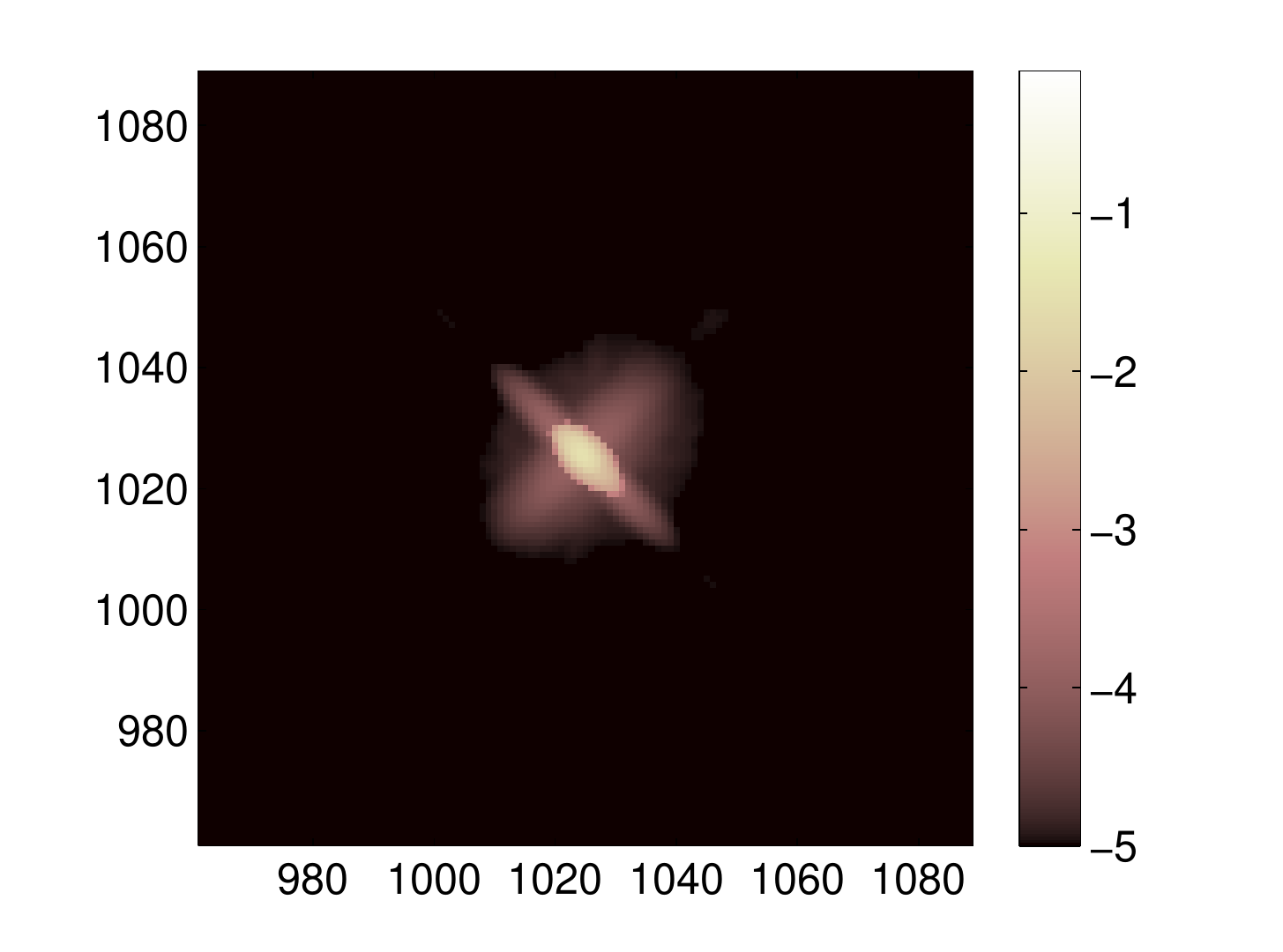} 
	\end{tabular}
\caption{Logarithm of the combined parametric/non-parametric filters for the SECCHI/EUVI telescope taken inside the set $\Gamma$. (left) Using the parametric PSF given by the {\tt euvi\_psf.pro} procedure of \emph{SolarSoft}, \ie $\bs h_{\rm p_1-np}^*$. (right) Using the parametric PSF given by the {\tt euvi\_deconvolve.pro} procedure of \emph{SolarSoft}, \ie $\bs h_{\rm p_2-np}^*$.}
\label{fig:ExpResults_EUVI_Filter_PNP}
\end{figure}

Hereafter, we provide a comparison between the following filters: (1) the parametric PSF given by the {\tt euvi\_psf.pro} procedure of \emph{SolarSoft}, \ie $\bs h_{\rm p_1}$, the standard PSF used for SECCHI/EUVI image analysis; (2) the parametric PSF estimated by Shearer et al.~\cite{2012ApJ...749L...8S}, \ie $\bs h_{\rm p_2}$, and given by the {\tt euvi\_deconvolve.pro} procedure of \emph{SolarSoft}; (3) the non-parametric PSF, \ie $\bs h_{\rm np}^*$; (4) the parametric/non-parametric PSF obtained by incorporating in the acquisition model the parametric PSF given by {\tt euvi\_psf.pro}, \ie $\bs h_{\rm p_1-np}^*$; and (5) the parametric/non-parametric PSF obtained by incorporating in the acquisition model the parametric PSF given by {\tt euvi\_deconvolve.pro}, \ie $\bs h_{\rm p_2-np}^*$. The core of the parametric and non-parametric filters are presented in Figure~\ref{fig:ExpResults_EUVI_Filter} and the core of the resulting parametric/non-parametric filters are depicted in Figure~\ref{fig:ExpResults_EUVI_Filter_PNP}.

We can observe that while the two parametric PSFs favor both diagonals, the one given by the \linebreak {\tt euvi\_deconvolve.pro} presents a slight dominance of that at 45~degrees. The estimated non-parametric PSF clearly favors one of the two diagonals. Nevertheless, it presents an orientation at $-45$~degrees. The source of this difference in orientation is unknown and it should be further investigated. This, however, is beyond the scope of the present paper. Both parametric/non-parametric PSFs present a similar behavior, favoring both diagonals but with a slight dominance of that at $-45$~degrees.

\begin{figure}[ht]
\centering
	\begin{tabular}{c c c}
		\includegraphics[height=4cm]{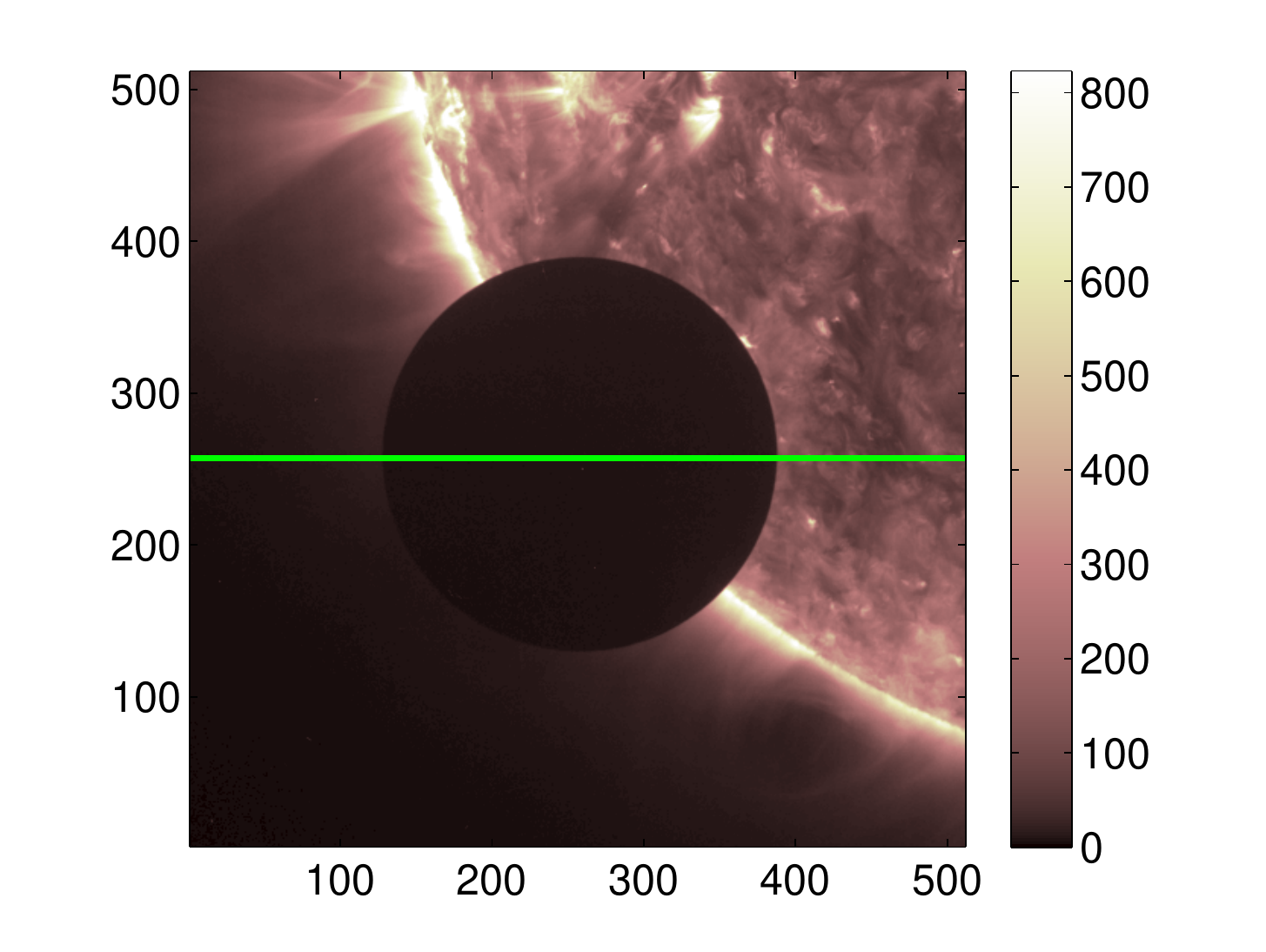} &
		\includegraphics[height=4cm]{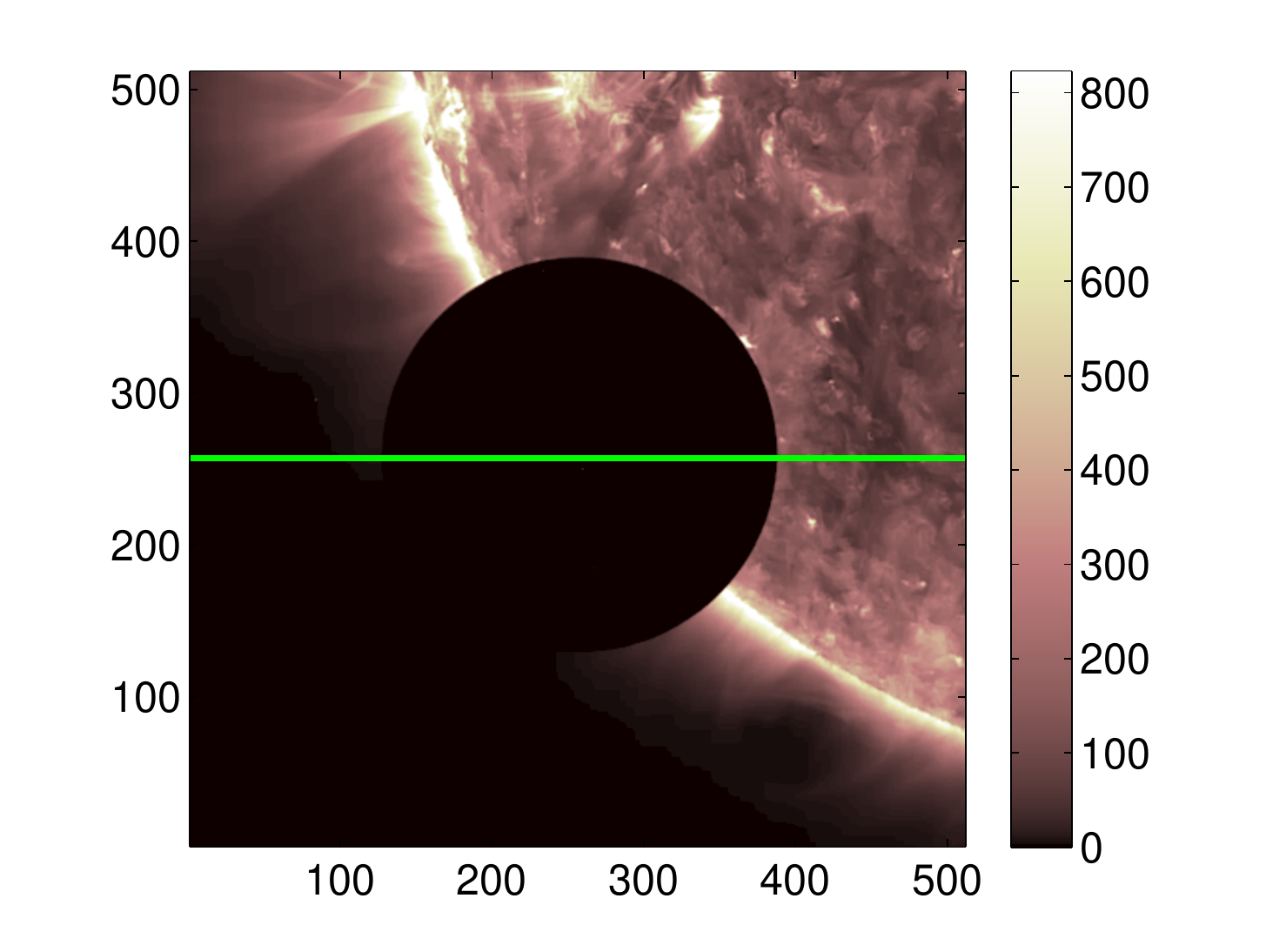} &
		\includegraphics[height=4cm]{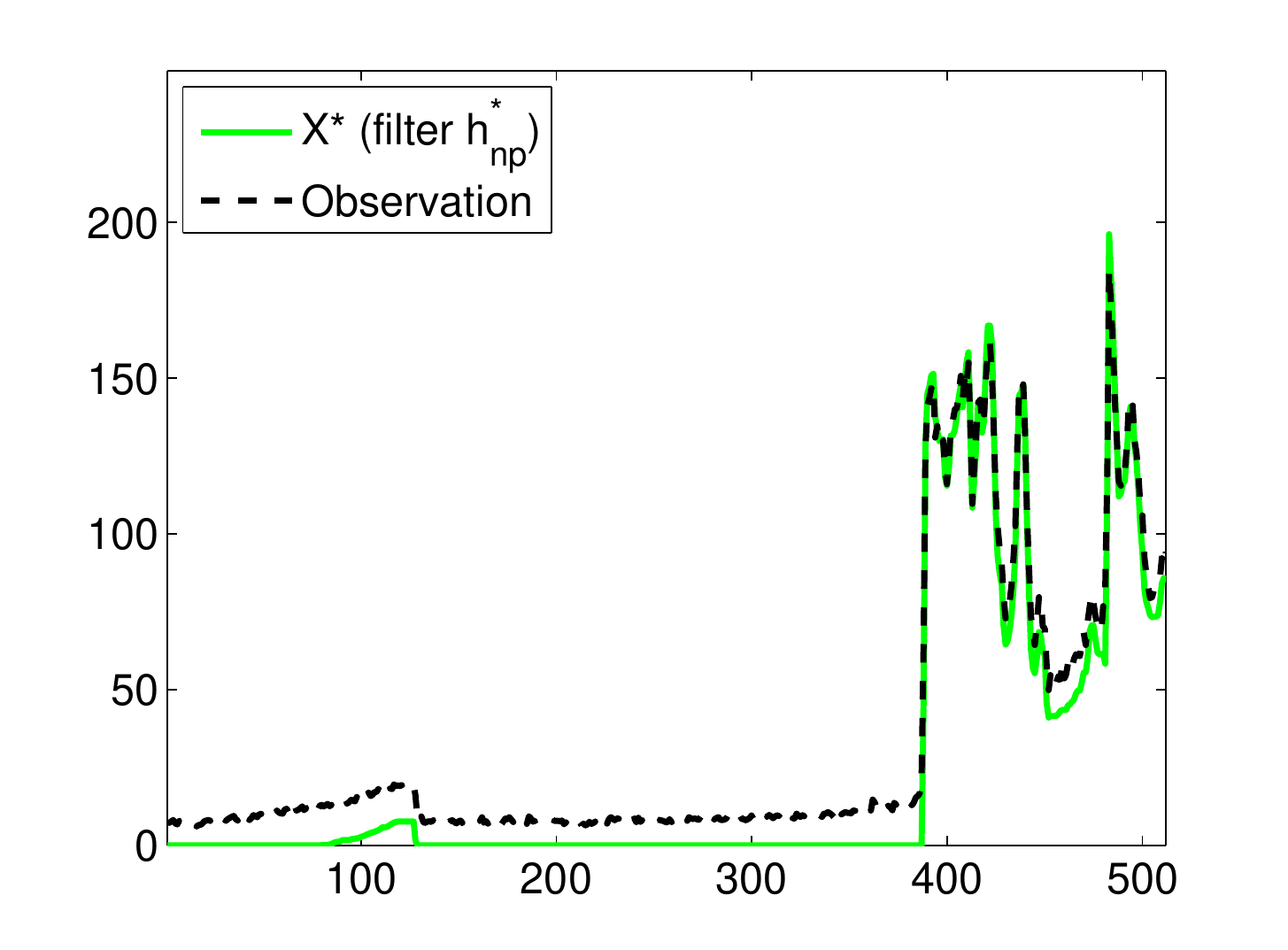} 
	\end{tabular}
\caption{Image reconstruction results for the Moon transit observed by SECCHI/EUVI at 08:02 UT on February 25th 2007. Results are shown for the non-parametric filter, \ie $\bs h_{\rm np}^*$. (left) Observed image, (center) 2-D reconstruction result, and (right, best viewed in color) 1-D profile along $y~=~257$. The figures on the left and on the center contain a green line indicating the 1-D profiles shown on the right.}
 \label{fig:ExpResults_Moon_Image}
\end{figure}

The filters from Figure~\ref{fig:ExpResults_EUVI_Filter} and Figure~\ref{fig:ExpResults_EUVI_Filter_PNP} were validated using the non-blind deconvolution described in Section~\ref{sec:NBID}. Similarly to what has been done for SDO/AIA, the value of $\rho$ has been selected adaptively by optimizing the value of the residual whiteness. Let us note that, as observed empirically, the value of $\rho$ must be kept smaller than the one obtained in Algorithm~\ref{alg:alternatemin_ext}. The filter validation was performed on the Moon transit image taken by SECCHI/EUVI at 08:02 UT on February 25th 2007 (see Figure~\ref{fig:ExpResults_Moon_Image}\mbox{-(left)}). Let us note that this patch has not been used before for estimating $\bs h^*$ and that the non-blind deconvolution is applied to the modified observation using $\mu = 12.5$ DN. We quantify the apparent Moon emissions by summing the pixel values inside the disk. Table~\ref{tab:ExpResults_Moon_DiskIntensity} displays the disk intensity ratio for the image deconvolved using the different filters presented above. 

\begin{table}[ht]
	\centering
	\begin{tabular}{ c  c }
		\hline \hline
		Filter & $\cl S_X/ \cl S_Y$ \\
		\hline
		$\bs h_{\rm p_1}$ & 5.72 $\times 10^{-3}$ \\
		\hline
		$\bs h_{\rm p_2}$ & 5.80 $\times 10^{-3}$ \\
		\hline
		$\bs h^*_{\rm np}$ & 2.02 $\times 10^{-3}$  \\
		\hline
		$\bs h^*_{\rm p_1-np}$ & 1.81 $\times 10^{-3}$ \\
		\hline
		$\bs h^*_{\rm p_2-np}$ & 1.29 $\times 10^{-3}$ \\
		\hline
	\end{tabular}
	\caption{Moon disk intensity ratio for the different filters.}
  \label{tab:ExpResults_Moon_DiskIntensity}
\end{table}

We notice that, as for the Venus transit images, when the long-range effect $\mu$ is not removed from the observations, the parametric PSFs considered in a larger support of 1024 $\times$ 1024 pixels are not able to remove the offset and to recover zero emissions inside the lunar disk. If the long-range effect is considered through the parameter $\mu$, the non-parametric PSF presents a better behavior and is able to reduce more of the disk emissions as compared to the parametric PSFs (see Table \ref{tab:ExpResults_Moon_DiskIntensity}). We also observe that the parametric/non-parametric PSFs reach lower disk apparent emissions.

In Figure~\ref{fig:ExpResults_Moon_Image} we illustrate the reconstruction results when using the non-parametric PSF, \ie $\bs h^*_{\rm np}$. The figure on the left presents the observed patch, the one on the center depicts the 2-D estimated image and the one on the right depicts a 1-D profile of along $y = 257$ (to allow the observation of the lunar disk). Due to lack of space, only the non-parametric PSF validation is illustrated.

We can observe that, in Figure~\ref{fig:ExpResults_Moon_Image}, part of the off-limb portion of the image is forced to zero. Since this part of the image is fainter than the lunar disk, the proposed non-blind deconvolution method in Section~\ref{sec:NBID} is not able to preserve such low intensities. However, finding a deconvolution method that would preserve these low intensities in the off-limb region (\eg by selecting other sparsity priors) is beyond the scope of this work.

\begin{figure}[ht]
\centering
	\begin{tabular}{c c c}
		\includegraphics[height=4cm]{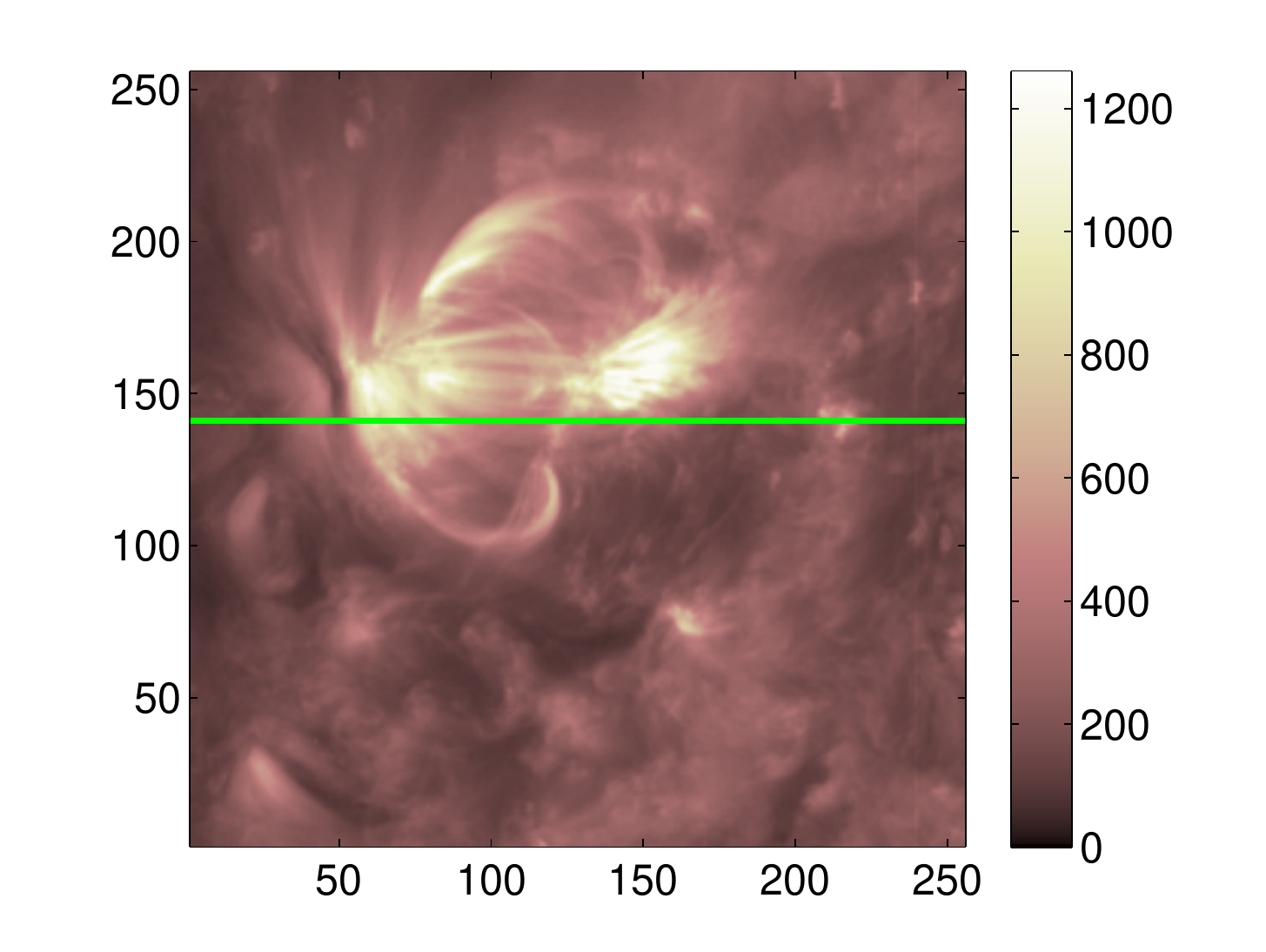} &
		\includegraphics[height=4cm]{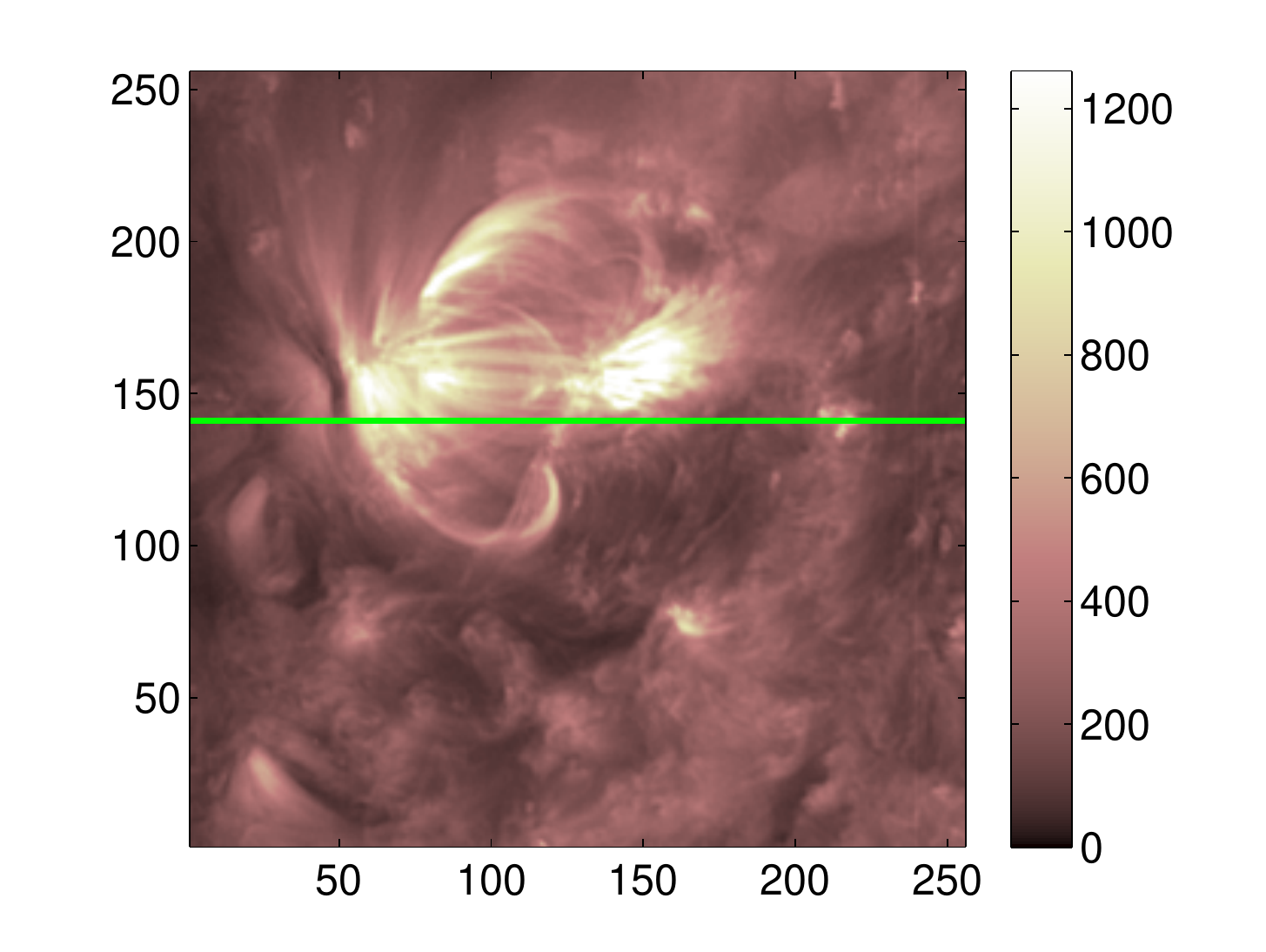}&
		\includegraphics[height=4cm]{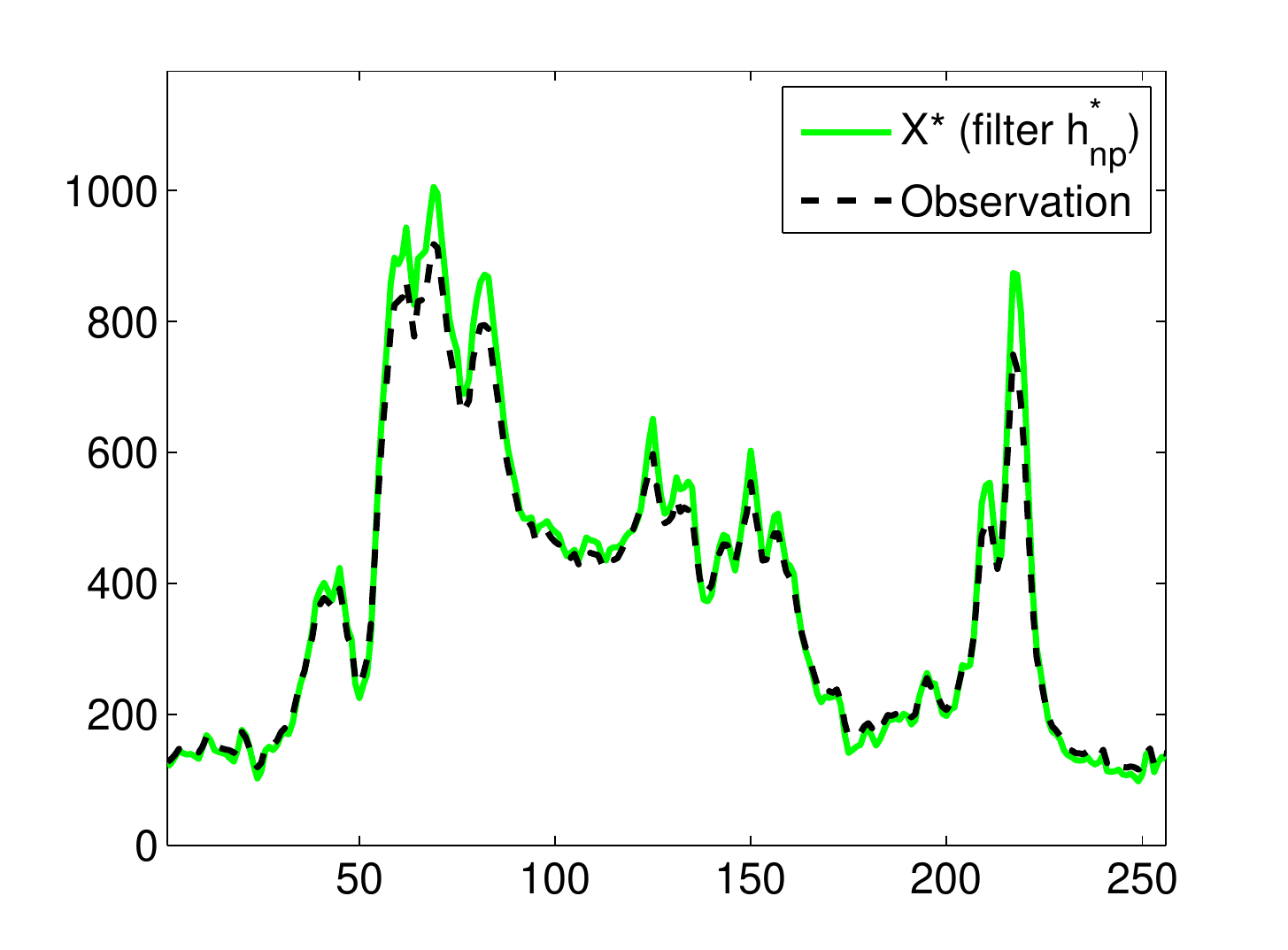}
	\end{tabular}
\caption{Reconstruction results for a SECCHI/EUVI image containing an active region. The image has been captured at 04:02 UT on February 25th 2007. Results are shown for the non-parametric filter, \ie $\bs h_{\rm np}^*$. (left) Observed image, (center) 2-D reconstruction result, and (right, best viewed in color) 1-D profile along $y = 141$. The figures on the left and on the center contain a green line indicating the 1-D profiles shown on the right.}
 \label{fig:ExpResults_AR_EUVI}
\end{figure}

The estimated filters were also used to deconvolve non-transit images. For this, the non-blind deconvolution formulation from Section~\ref{sec:NBID}, using an adaptive $\rho$, was applied to the modified observations with $\mu =12.5$~DN. The results are shown for a non-transit image taken by SECCHI/EUVI at 04:02 UT on February 25th 2007. We selected a portion of the original image of $256 \times 256$ pixels around the active region (see Figure~\ref{fig:ExpResults_AR_EUVI}-(left)). Figure~\ref{fig:ExpResults_AR_EUVI}-(center) depicts the 2-D estimated image using the non-parametric PSF, \ie $\bs h_{\rm np}^*$, and Figure~\ref{fig:ExpResults_AR_EUVI}-(right) shows a 1-D profile along $y=141$.

\begin{figure}[ht]
\centering
	\includegraphics[height=3.5cm]{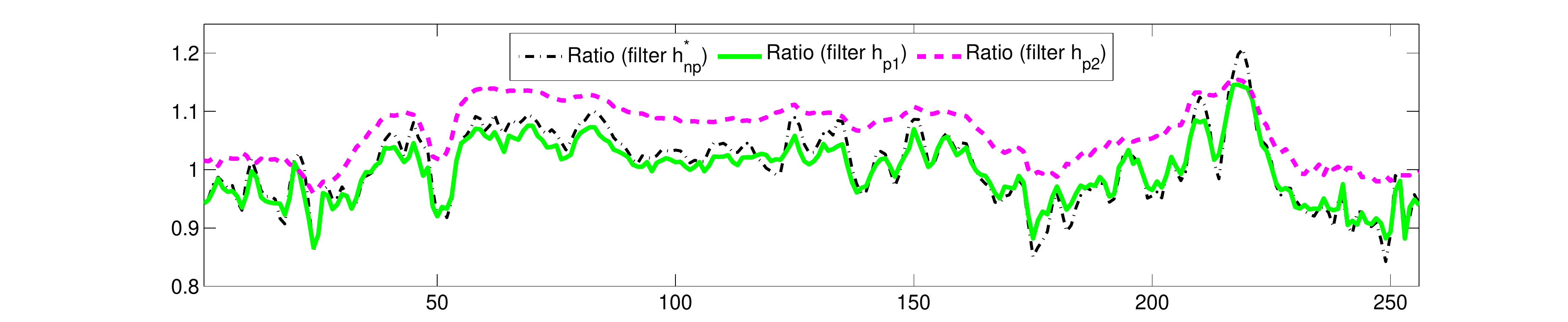} \\
	\includegraphics[height=3.5cm]{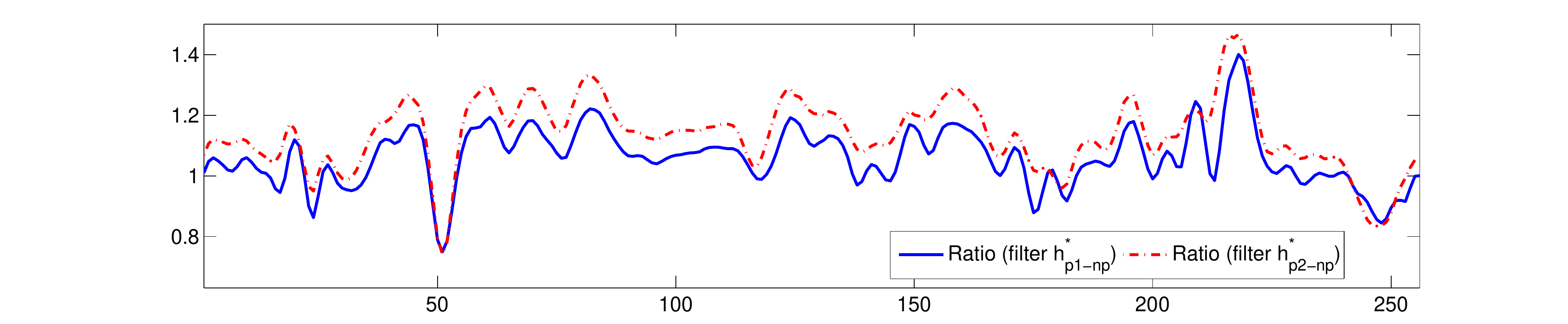}
\caption{Reconstruction results for the image in Figure~\ref{fig:ExpResults_AR_EUVI}-(left). The figure shows the ratio (computed using a pixel by pixel division) between the deconvolved and the observed images along $y=141$. (top, best viewed in color) Ratio for filters $\bs h_{\rm np}^*$, $\bs h_{{\rm p}_1}$, and $\bs h_{{\rm p}_2}$. (bottom, best viewed in color) Ratio for filters $\bs h_{\rm p_1-np}^*$ and $\bs h_{\rm p_2-np}^*$. The horizontal axis is in correspondence with the horizontal axis of Figure~\ref{fig:ExpResults_AR_EUVI}-(right).}
 \label{fig:ExpResults_AR_EUVI_Ratio}
\end{figure}

Figure~\ref{fig:ExpResults_AR_EUVI} shows that the estimated non-parametric PSF is able to enhance the image and provide more details. Similarly to SDO/AIA, we compare these results with the ones from the other PSFs by taking the ratio between the deconvolved images and the observation (computing a pixel by pixel division). Results are depicted in Figure~\ref{fig:ExpResults_AR_EUVI_Ratio}. We observe that the non-parametric PSF provides similar results to those obtained by the parametric PSF given by the {\tt euvi\_psf.pro} procedure. These two PSFs are able to provide higher details than the parametric PSF given by the {\tt euvi\_deconvolve.pro} procedure. The deconvolution resulting from the combined parametric/non-parametric PSFs present a higher correction from the observations than the ones obtained using the other three filters. This corresponds to what was observed for the Moon transit images in Table~\ref{tab:ExpResults_Moon_DiskIntensity}.

\section{Discussion and perspectives}
\label{sec:discussion}

We proposed a blind deconvolution method that allows one to recover the PSF of a given astronomical instrument provided the latter captures a celestial body transit (as observed in Figure~\ref{fig:Transit}). The proposed non-parametric approach is comparable to previous parametric ones with respect to the quality of the PSF core. It presents, however, some limitations in terms of the considered noise model and the estimation of the PSF long-range effects. Fortunately, the optimization techniques we use in this work are flexible enough to open perspectives of improvement in future works.

\subsection{Noise model}

The incidence of photon flux on a EUV telescope is converted to digital numbers (DN) through a series of steps, each potentially introducing some noise. The beam of photons impinges the optical system where the PSF acts as a blurring operator. Simultaneously, a spectral selection is performed on the signal before it reaches the CCD detector. The latter has an heterogeneous response across its surface. Finally, the camera electronics convert electrons into DN, adding the read-out noise.
The pixels in the resulting image can be modeled as the realization from a random variable $Y$, whose noise part can be decomposed into additive, Poissonian, and multiplicative degradations. The expectation ($\EE$) and the variance ($\VV$) of $Y$ then verify:
\begin{equation}
  \label{eq:noise_model}
  \EE[Y] = x
  \ \ \textrm{and} \ \ 
  \VV[Y] = \sigma^2 + \beta x + \alpha x^2,
\end{equation}
where $\sigma$ is the standard deviation of the additive component and $\alpha,\beta$ are parameters.

Shearer et al.~\cite{2012ApJ...749L...8S} handled Poisson corrupted data through a Variance Stabilization Transform (VST) which provides an approximated Gaussian noise distributed data. In order to generalize our AWGN model to models such as~(\ref{eq:noise_model}), a VST could be included in the $\ell_2$-fidelity term in~(\ref{eq:BID2}) and~(\ref{eq:NBID}), on both the observations and the convolution result~\cite{2012ApJ...749L...8S}. This can be done provided that we know the conversion between DN and the photon counts (\eg from instrument specifications). The algorithm described in Section~\ref{sec:AM} is adaptable to this stabilized fidelity term through specific proximal operators or gradient descent~\cite{4738431,anthoine2012some}. Notice, however, that such a stabilization cannot be applied only on the observations as a mere preprocessing of the data, while using afterwards other deconvolution methods assuming additive Gaussian noise corruption. Indeed, the VST being a non-linear process of the form \linebreak $x \in \bb R_+ \mapsto \sqrt{\alpha x + \beta}$ for some $\alpha,\beta \in \bb R_+$~\cite{ansc}, it breaks the convolutive model (\ref{eq:discrete_model_vector}) on which those deconvolution methods rely. In other words, the VST of the convolution of $\bs h$ by $\bs x_j$ in (\ref{eq:discrete_model_vector}) is  not equivalent to (and hardly approximable by) the convolution of the variance stabilized vectors, which breaks the applicability of those techniques.  As an alternative to using a VST, the true Poisson distribution could also be used to design a specific fidelity term, which is based on the negative log-likelihood of the posterior distribution induced by the observation model, \ie the KL divergence of this distribution~\cite{Fish:95,anthoine2012some,0266-5611-29-6-065017}. However, it is unclear how to adapt the method proposed by Attouch et al.~\cite{Attouch:2010:PAM:1836121.1836131} to the resulting framework.

\subsection{Estimation of long-range effects}

Our paper aimed at preventing strong assumptions on the shape of the central part of the PSF.  We provide a general deconvolution method for different types of telescopes, the generality of our method being of particular interest for telescopes exhibiting optical aberrations. This strategy differs from the one adopted by most parametric PSF estimations. Nonetheless, additional convolutive filter regularizations could be included in our scheme by, for instance,  promoting the sparsity (in synthesis or in analysis) of this filter in an appropriate basis (\eg wavelet basis), or by enforcing a certain decaying law of its amplitude in function of the radial distance. Both kind of priors can be expressed as convex costs (\eg with a $\ell_1$-norm or a weighted $\ell_\infty$-ball constraint) with closed-form (or \emph{simple}) proximal operators. These adaptations can thus be integrated in Algorithm~\ref{alg:alternatemin} with additional efforts for limiting the computational time of the more complex deconvolution procedure.  Moreover, such additional priors can help in enlarging the support $\Gamma$ where the filter is truly estimated, hence reaching an estimation of the long-range PSF. 

Regarding the approximation of the long-range PSF impact by a constant, we notice that, instead of estimating the constant $\mu$ in a pre-processing step (see Section \ref{sec:FilterPrior}), in future work we could let  $\mu$ be a free parameter of the deconvolution problem (\ref{eq:BID2}). The value of $\mu$ can then be optimized jointly with the image and the filter as this does not break the convexity of the sub-problems detailed in Section~\ref{sec:AM}. 

Finally, inspired by the works of Shearer et al.~\cite{2012ApJ...749L...8S,Shearerphdthesis}, a convolutive combination of a known parametric filter with an unknown non-parametric one was considered in this work to further stabilize the non-convexity of the blind deconvolution problem. However, this construction is limited since, from the convolution theorem, no correction of the parametric PSF can be made in the part of its spectrum where it vanishes. Future works should therefore consider additive correction of the parametric PSF as suggested by Poduval et al.~\cite{2013ApJ...765..144P}, a modification that Algorithm~\ref{alg:alternatemin} could also include.

\section{Conclusion}

We have demonstrated how a non-parametric blind deconvolution technique is able to estimate the core of the PSF of an optical instrument with high quality. The quality of the estimated PSF core is comparable with the one provided by parametric models based on the optical characterization of the imaging instrument. We also demonstrate that, if the parametric PSF is incorporated in the acquisition model, the blind deconvolution approach is able to provide a `corrected' PSF such that most of the apparent emissions inside the disk of a celestial body during a solar transit can be removed.
Let us note that non-parametric techniques cannot outperform the accuracy of parametric methods, however in situations where the telescope imaging model cannot be obtained due to some instrument's properties (\eg PICARD/SODISM~\cite{2014SoPh..289.1043M}), the non-parametric method presents a great advantage. Moreover, the proposed method is not specific to a given instrument but can be applied to any optical instrument provided that we have strong knowledge on some image pixels values and their exact location, such as the information available during the transit of the Moon or a planet. We have also shown that the use of the Proximal Alternating Minimization technique proposed by Attouch et al.~\cite{Attouch:2010:PAM:1836121.1836131}, allows to efficiently solve the non-convex problem of blind deconvolution with theoretical convergence guarantees. Furthermore, we show the importance of considering multiple observations for the same filter in order to provide a better conditioning to the filter estimation problem.

\section*{Acknowledgements}
Part of this work is funded by the AOC SPW Project (SKYWIN - 6894). LJ
is supported by the Belgian FRS-FNRS fund. VD acknowledges support
from the Belgian Federal Science Policy Office through the ESA-PRODEX
program, grant No. 4000103240. The authors would like to thank
Craig DeForest and Jean-Fran\c{c}ois Hochedez for inspiring discussions and Jean-Pierre Wuelser for having kindly provided us Moon trajectory data in SECCHI/EUVI images. 
Thanks are extended to the two anonymous referees for their comments, 
which substantially improved the quality of the paper. Also, we would like to thank the anonymous reviewer who provided the MATLAB code that computes only the mesh diffraction component for the PSF of SDO/AIA.

\appendix

\section{On the approximation of the long-range PSF impact by a constant}
\label{sec:long_range_psf}

In this work, in order to compensate the restriction of the estimated filter to its core, we assume that, for moderate solar intensities, the long-range effect of the PSF can be approximated by a constant inside a limited patch. This assumption is compatible with the observations made on the convolution of a solar image with a parametric filter that only contains the long-range effect, \ie a filter built by taking a standard parametric PSF containing long-range patterns and setting to zero the central pixels inside a window of size $(2b+1)\times(2b+1)$ with $b = 64$. 

\begin{figure}[ht]
\centering
	\includegraphics[height=3.75cm]{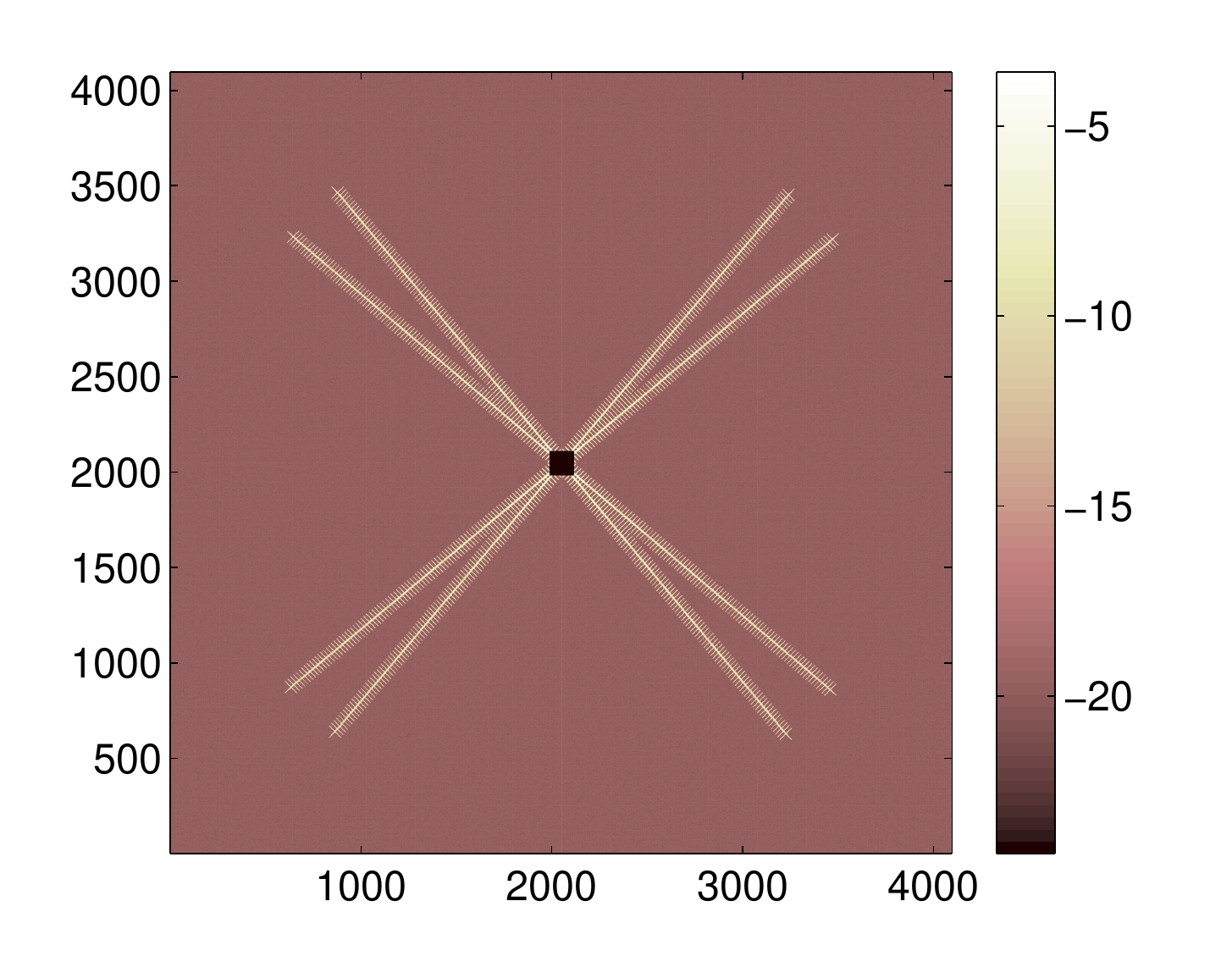} 
	\includegraphics[height=3.75cm]{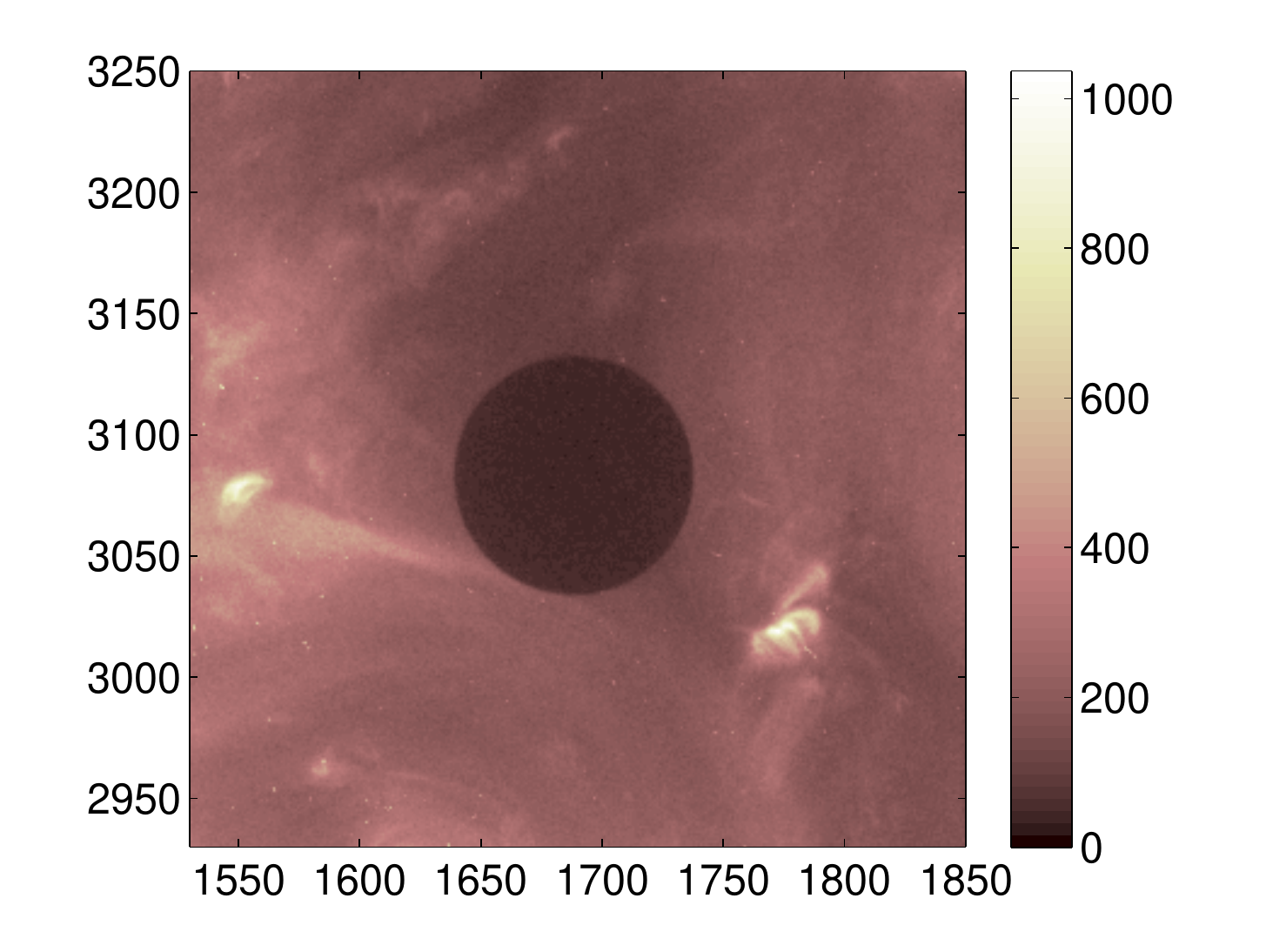} 
	\includegraphics[height=3.75cm]{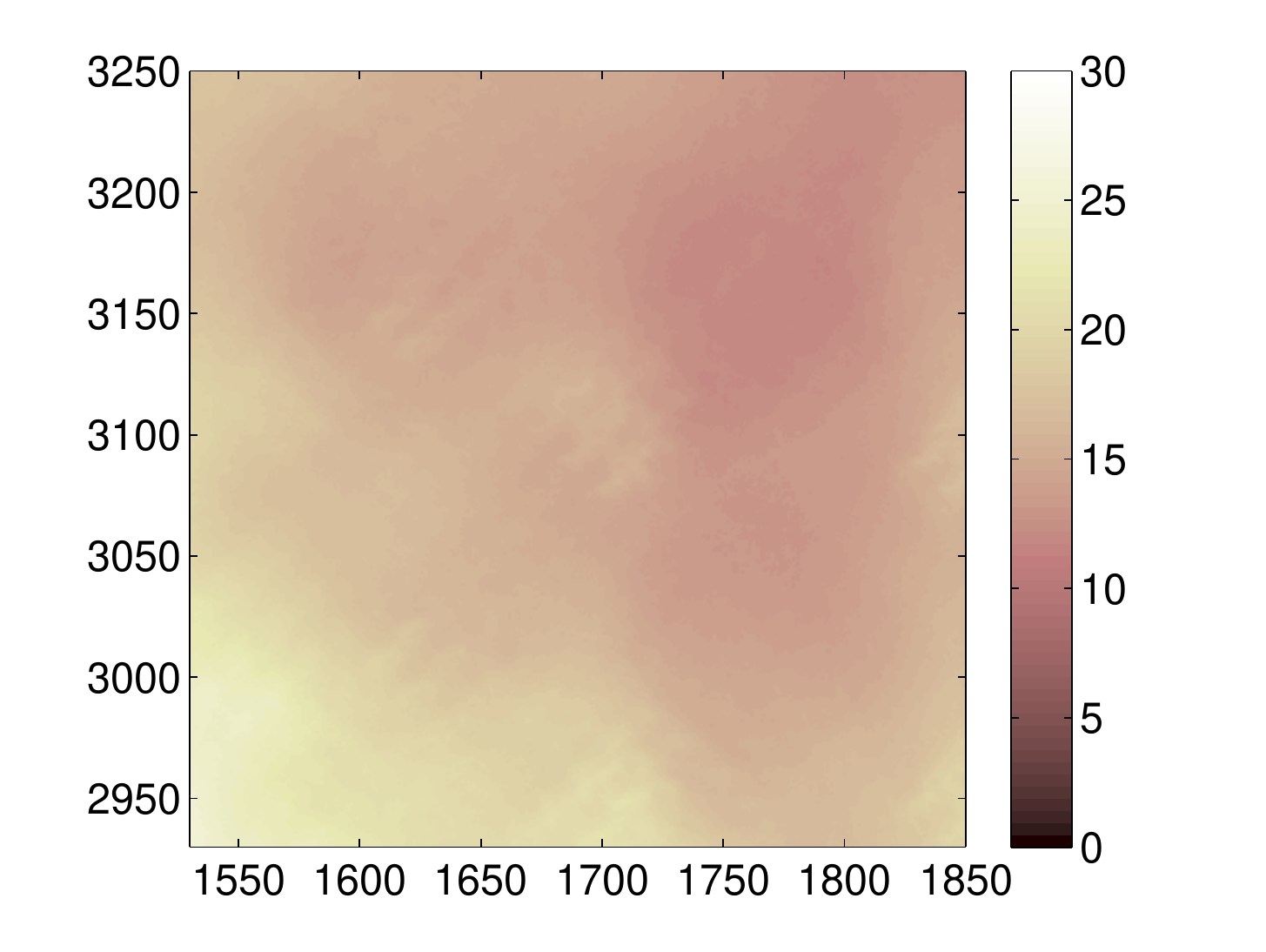} 
\caption{Validating the approximation of the long-range PSF impact by a constant inside the disk of Venus. (left) Logarithm of the filter used for the validation: parametric filter estimated by \cite{aia_psf} with the central pixels set to zero inside a window of size $(2b+1)\times(2b+1)$, with $b = 64$. (center) Observed image inside a window containing the disk of Venus. (right) Convolved image inside a window containing the disk of Venus. The intensity within the window is almost constant.}
\label{fig:constant_mu_AIA}
\end{figure}

For SDO/AIA, we use the parametric filter estimated by \cite{aia_psf}. The validation was performed using a $4096 \times 4096$ level 1 image from the Venus transit recorded by the 19.3~nm channel of AIA at 00:00:01 UT on June 6th 2012. These results are presented in Figure~\ref{fig:constant_mu_AIA}.

\begin{figure}[ht]
\centering
	\includegraphics[height=3.75cm]{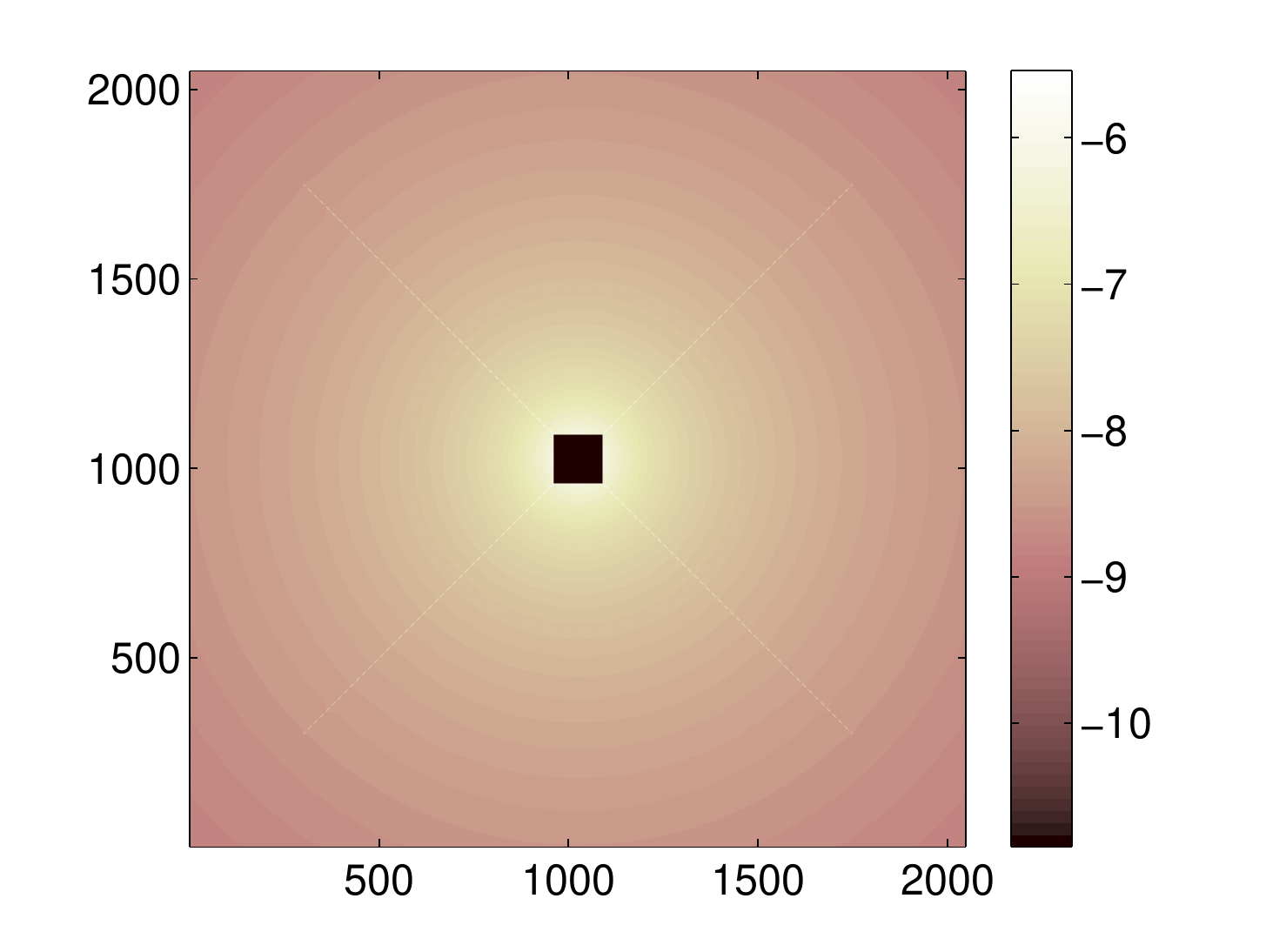} 
	\includegraphics[height=3.75cm]{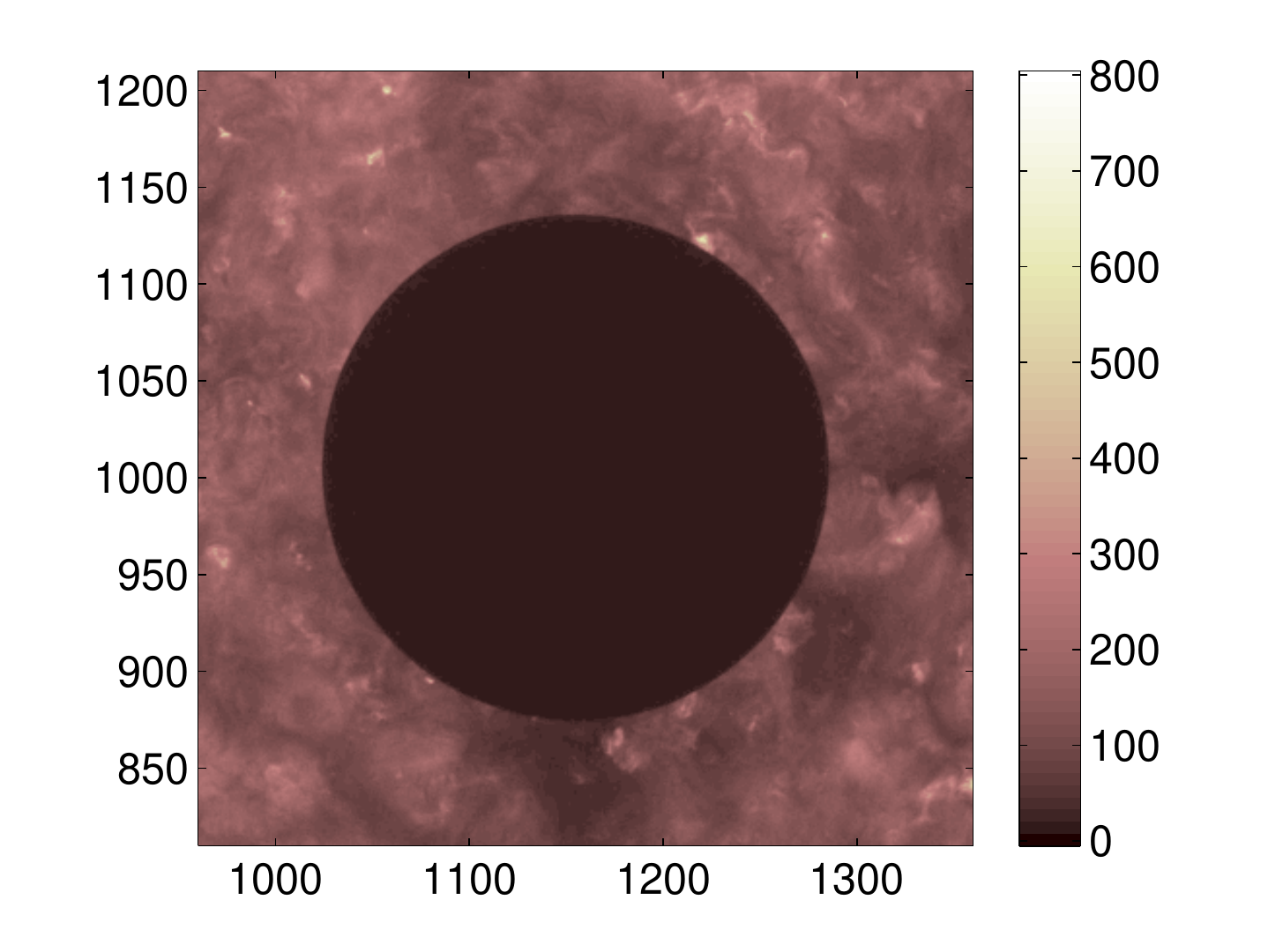} 
	\includegraphics[height=3.75cm]{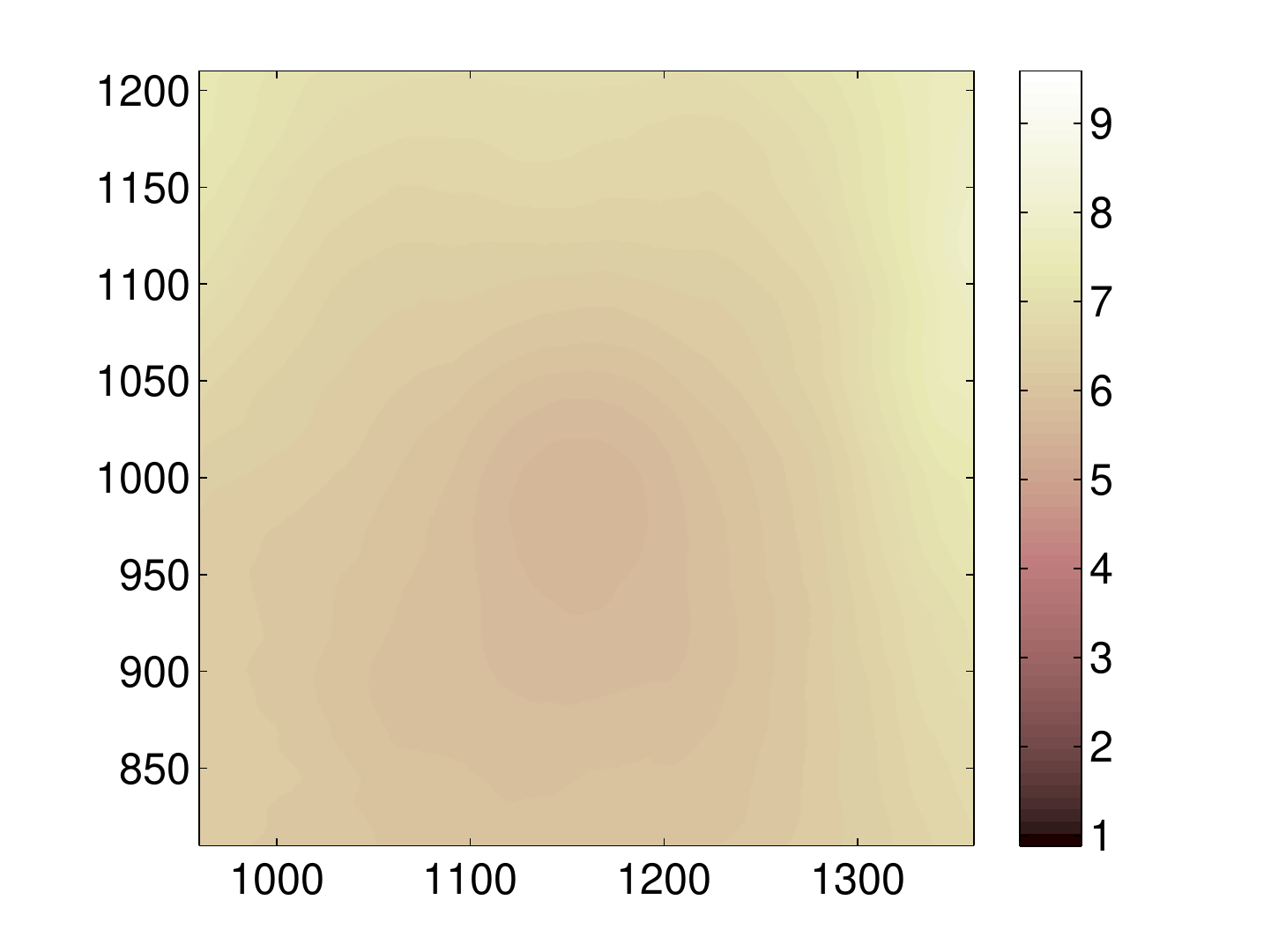} 
\caption{Validating the approximation of the long-range PSF impact by a constant inside the lunar disk. (left) Logarithm of the filter used for the validation: parametric filter given by the {\tt euvi\_psf.pro} procedure of \emph{SolarSoft} with the central pixels set to zero inside a window of size $(2b+1)\times(2b+1)$, with $b = 64$. (center) Observed image inside a window containing the lunar disk. (right) Convolved image inside a window containing the lunar disk. The window intensity is almost constant.}
\label{fig:constant_mu_EUVI}
\end{figure}

For SECCHI/EUVI, we use the parametric filter given by the {\tt euvi\_psf.pro} procedure of \emph{SolarSoft}. The validation was performed using a $2048 \times 2048$ image from the Moon transit recorded by the 17.1~nm channel of EUVI at 14:00:00 UT on February 25th 2007. The convolution results are presented in Figure~\ref{fig:constant_mu_EUVI}.

The resulting low frequency images in Figure~\ref{fig:constant_mu_AIA}-(right) and Figure~\ref{fig:constant_mu_EUVI}-(right) validate the use of a constant $\mu$ to approximate the long-range effect. Let us note that this is just a first approach to estimate the long-range effect and that more sophisticated methods can be used as explained in Section~\ref{sec:discussion}.

\section{Algorithm initialization}
\label{sec:init}

Algorithm \ref{alg:alternatemin} requires an initial value of the image ($\bs X_0$), the filter ($\bs h_0$) and the regularization parameter ($\rho_0$). In this section we describe in details how we proceed with this initialization.

\subsection{Image and filter}
\label{sec:init_ImageFilter}

For the first value of $\rho$, the alternated algorithm (Algorithm \ref{alg:alternatemin}) is initialized with the trivial solution, where the image is given by the modified observations $\bs Z$ and the filter is given by the delta function $\bs \delta_0$. For the subsequent iterations on $\rho$, the image and the filter are initialized using the value of the previous iteration.

\subsection{Regularization parameter}
\label{sec:param}

The regularization parameter has an important role in solving the first step in Algorithm \ref{alg:alternatemin}, since it determines the trade-off between the image regularization and the fidelity to the observations. Several works have studied the choice of this parameter when solving general ill-posed inverse problems. In basis pursuit denoising, \cite{donoho01091994} proposed a universal value for $\rho$ given by $\sigma \sqrt{2 \textrm{log}(N)}$. Since this value increases with the amount of samples ($N$), it tends to provide overly smoothed images. Another state-of-the-art choice, proposed by \cite{Donoho95adaptingto}, is based on the minimization of Stein's Unbiased Risk Estimate (SURE) \cite{stein1981}. This method provides accurate denoising results, however it requires complete knowledge of the degradation model which makes it unsuitable for our blind deconvolution problem. In this work, we use the Bayesian approach proposed by \cite{862633}, to estimate the parameter $\rho$ using the noise characteristics and the probability distribution function of the wavelet coefficients: 
\begin{equation}
\label{eq:rhoMAP}
	\rho = \frac{\sqrt{2} \ \sigma^2}{\tau},
\end{equation}
where $\tau$ is the standard deviation of the wavelet coefficients of the signal to reconstruct. When we assume that they follow a Laplacian probability distribution with zero mean, the value of $\tau$ can be estimated as: $\textstyle\frac{\sqrt{2}}{PQ}\sum_{j = 1}^P\|(\bs A)_j\|_1$, where $\bs A = \bs S_\Theta \bs \Psi^\mathtt{T} \bs X$ is the matrix containing the image wavelet coefficients inside the set $\Theta$. 

Since the Ground Truth signal $\bs X$ is usually not available, the value of $\tau$ in (\ref{eq:rhoMAP}) is determined using the modified observations $\bs Z$, \ie $\tau = \tau_Z = \textstyle\frac{\sqrt{2}}{PQ}\sum_{j=1}^P\|(\bs A')_j\|_1$, with $\bs A' = \bs S_\Theta \bs \Psi^\mathtt{T} \bs Z$. Since this value is higher than the actual $\rho$, it will lead to over-regularized images. Therefore, we propose to refine the value of $\rho$ by an iterative update based on the discrepancy principle between the actual noise energy in (\ref{eq:epsilon}) and the energy of the residual image (\ref{eq:residual}). By performing a maximum of five updates of the parameter $\rho$, we assure an appropriate image regularization. The iterative estimation of $\rho$ is summarized in Algorithm \ref{alg:alternatemin_ext}.

\begin{algorithm}[ht]
\newcommand{\sComment}[1]{\vspace{1mm}\Statex\quad\ \emph{\underline{#1}}:\vspace{1mm}}
  \caption{Iterative estimation of $\rho$ 
    \label{alg:alternatemin_ext}}
  \begin{algorithmic}[1]
    \Require{\parbox[t]{10cm}{$\bs X^{(1)}\!=\!\bs Z$; $\bs h^{(1)}\!=\!\bs \delta_0$; $\rho^{(1)}\!=\!(\sqrt{2} \sigma^2/\tau_Z)$; $\varepsilon\!=\!\sigma \sqrt{NP\!+\!2\sqrt{NP}}$; MaxIter = 5}\vspace{2mm}}
		\For{$l = 1$ to MaxIter}
			\sComment{Images and Filter estimation}
				\State Estimate $\bs X^{(l+1)}$ and $\bs h^{(l+1)}$ using Algorithm \ref{alg:alternatemin} with $\bs X_0 = \bs X^{(l)}$, $\bs h_0 = \bs h^{(l)}$ and $\rho_0 = \rho^{(l)}$
			\sComment{Compute residual and whiteness measure}
				\State $\bs R^{(l)} = \bs Y - \bs\Phi(\bs h^{(l+1)}) \ \bs X^{(l+1)}$
				\State Compute $\cl M(\bs R^{(l)})$ using (\ref{eq:MRP})
           	\sComment{Parameter update}
				\State $\varepsilon^{(l)} = \|\bs R^{(l)}\|_F$
				\State $\rho^{(l+1)} = \rho^{(l)} \left( \varepsilon/\varepsilon^{(l)} \right)$
			 \sComment{Stop when residual is spectrally whiter}
				\If{ $ \cl M^{(l+1)} < \cl M^{(l)}$ }\quad  \rm break.
				\EndIf
		\EndFor
		\State Return $\bs X^* = \bs X^{(l+1)}$ and $\bs h^* = \bs h^{(l+1)}$
  \end{algorithmic}
\end{algorithm}

\section{Numerical reconstruction algorithms}
\label{sec:appendix_CP_APG}

For the sake of completeness, in this section, we briefly describe the proximal algorithms that we use to solve the two subproblems of Algorithm~\ref{alg:alternatemin}. We refer the reader to \cite{Chambolle2011, CombettesPesquet2011, OPT-003} for a comprehensive understanding.

Proximal algorithms rely on a key element in convex signal analysis, the proximal operator
\begin{equation}
\label{eq:prox}
	\textrm{prox}_{\nu\varphi} \bs z := \textrm{argmin}_{\bs u\in\bb R^D} \ \varphi(\bs u) + \frac{1}{2\nu}\|\bs u - \bs z\|^2,
\end{equation}
which is uniquely defined for any function $\varphi \in \Gamma_0(\bb R^D)$, for some $D\in \bb N$~\cite{CombettesPesquet2011}. Proximal algorithms allow the minimization of non-smooth functions such as the $\ell_1$-norm and the indicator functions present in the image and filter subproblems.

\subsection{First subproblem: Image estimation}
\label{sec:appendix_CP}

In the first subproblem, we are interested in finding the image candidate that minimizes
\begin{equation}
\label{eq:step1}
	\bs X^{(k+1)} = \textrm{argmin}_{\tilde{\bs X} \in \bb R^{N \times P}} \ \rho \ \| \ \bs S_\Theta \ \bs \Psi^\mathtt{T} \ \tilde{\bs X} \ \|_1 + \textstyle\frac{1}{2} \| \ \bs Z - \bs \Phi (\bs h^{(k)}) \ \tilde{\bs X} \ \|^2_F + \frac{\lambda_x^{(k)}}{2} \| \ \tilde{\bs X} - \bs X^{(k)} \ \|_F^2 + \imath_{\cl P_0}(\tilde{\bs X}),
\end{equation}
a problem that contains a sum of four functions belonging to $\Gamma^0(\bb R^D)$, for some dimensions $D~\in~\{QP, NP\}$. For this, we use the Chambolle-Pock (CP) primal-dual algorithm~\cite{Chambolle2011}, which is commonly used for solving the minimization of two functions in $\Gamma^0$. In order to take into account the sum of more than two functions, the algorithm is extended as in \cite{Gonzalez2014421}. If we set $F_1 (\bs V_1) = \rho \| \bs V_1 \|_1$ for $\bs V_1 \in \bb R^{Q \times P}$, $F_2 (\bs V_2) = \textstyle\frac{1}{2} \| \bs Z - \bs V_2 \|^2_F$ for $\bs V_2 \in \bb R^{N \times P}$, $F_3(\bs V_3) = \frac{\lambda_x^{(k)}}{2} \| \bs V_3 - \bs X^{(k)} \|_F^2$ for $\bs V_3 \in \bb R^{N \times P}$ and $H(\bs U) = \imath_{\cl P_0}(\bs U)$ for $\bs U \in \bb R^{N \times P}$, the CP iterations are:
\begin{eqnarray}
\label{eq:CP_Algorithm_PS}
	\nonumber \bs V_1^{(t+1)}\hspace{-2.5mm}&=\textrm{prox}_{\nu F_1^\star} \left(\bs V_1^{{(t)}} + \nu \bs S_\Theta \bs \Psi^\mathtt{T} \bar{\bs U}^{(t)} \right),\hspace{5cm}\\
	\nonumber \bs V_2^{(t+1)}\hspace{-2.5mm}&=\textrm{prox}_{\nu F_2^\star} \left(\bs V_2^{{(t)}} + \nu \bs \Phi (\bs h^{(k)}) \bar{\bs U}^{(t)} \right),\hspace{4.8cm}\\
	\nonumber \bs V_3^{(t+1)}\hspace{-2.5mm}&=\textrm{prox}_{\nu F_3^\star} \left(\bs V_3^{{(t)}} + \nu \bar{\bs U}^{(t)} \right),\hspace{5.9cm}\\
	\nonumber \bs U^{{(t+1)}}\hspace{-2.5mm}&= \textrm{prox}_{\frac{\alpha}{3} H} \left\{\bs U^{{(t)}} - \frac{\beta}{3} \left[ \bs \Psi \bs S_\Theta^\mathtt{T} \bs V_1^{{(t+1)}} + (\bs \Phi (\bs h^{(k)}))^\mathtt{T} \bs V_2^{{(t+1)}} + \bs V_3^{{(t+1)}} \right]\right\},\\
	\bar{\bs U}^{(t+1)}\hspace{-2.5mm}&= 2\,\bs U^{(t+1)} - \bs U^{(t)},\hspace{7.2cm}  
\end{eqnarray}
with $\bs U^{(t)}$ tending to a minimizer $\bs X^{(k+1)}$ of (\ref{eq:step1}) for $t \to + \infty$. The functions $F_i^\star$ are the convex conjugates of functions $F_i$, with $i = \{1,2,3\}$, and their proximal operators are computed via the proximal operator of $F_i$ in (\ref{eq:prox}) by means of the conjugation property~\cite{CombettesPesquet2011}: $\textrm{prox}_{\nu F_i^\star} \bs z = \bs z - \nu \textrm{ prox}_{\frac{1}{\nu} F_i} \left( \frac{1}{\nu}\bs z\right)$. The step sizes $\nu$ and $\beta$ are adaptively selected using the procedure described in \cite{Gonzalez2014421}.

\subsection{Second subproblem: Filter estimation}
\label{sec:appendix_APG}

In the second subproblem, we are interested in finding the filter candidate that minimizes
\begin{equation}
\label{eq:step2}
	\bs h^{(k+1)} = \textrm{argmin}_{\tilde{\bs h} \in \bb R^{N}} \ \textstyle\frac{1}{2} \sum_{j=1}^{P} \| \ \bs z_j - \bs \Phi (\bs x_j^{(k+1)}) \ \tilde{\bs h} \ \|^2_2 + \frac{\lambda_h^{(k)}}{2} \| \ \tilde{\bs h} - \bs h^{(k)} \ \|_F^2 + \imath_{\cl D}(\tilde{\bs h}),
\end{equation}
with $\bs \Phi (\bs x_j^{(k+1)}) \in \bb R^{N \times N}$ a circulant matrix of kernel $\bs x_j^{(k+1)}$. This problem has the following shape
\begin{equation}
	\textrm{argmin}_{\bs u \in \bb R^{N}} F(\bs u) + G(\bs u),
\end{equation}
with $F (\bs u) = \textstyle\frac{1}{2} \sum_{j=1}^{P} \| \ \bs z_j - \bs \Phi (\bs x_j^{(k+1)}) \ \bs u \ \|^2_2 + \frac{\lambda_h^{(k)}}{2} \| \bs u - \bs h^{(k)} \|_2^2$ and $G(\bs u) = \imath_{\cl D}(\bs u)$. Since $F : \bb R^N \rightarrow \bb R$ is convex and differentiable with gradient $\nabla F$, and $G : \bb R^N$ to $\bb R \cup \{+\infty\}$ belongs to $\Gamma^0 (\bb R^N)$, the problem can be solved using the Accelerated Proximal Gradient (APG) Method \cite{OPT-003}. 
The APG iterations are:
\begin{eqnarray}
\label{eq:APG_Algorithm}
	\nonumber \bs v^{(t+1)}\hspace{-2mm}&= \bs u^{(t)} + \beta^{(t)} \left(\bs u^{(t)} - \bs  u^{(t-1)}\right), \hspace{0.9cm}\\
	\nonumber \bs u^{{(t+1)}}\hspace{-2mm}&= \textrm{prox}_{\nu G} \left(\bs v^{(t+1)} - \nu \nabla F(\bs v^{(t+1)}) \right),\\
	\beta^{(t)} &= \frac{t}{t+3}, \hspace{4.3cm}  
\end{eqnarray}
with $\bs u^{(t)}$ tending to a minimizer $\bs h^{(k+1)}$ of (\ref{eq:step2}) for $t\to + \infty$. The step size $\nu$ is adaptively selected using the line search of \cite{BeckTeboulle2010}.

\section{Convolutions with unknown boundary conditions}
\label{sec:appendix_ubc}

As explained in Section \ref{sec:Model}, our blind deconvolution approach is realized in a restricted FoV of size $n\times n$. For all our computations, as those realized in Algorithm~\ref{alg:alternatemin}, fast convolution methods exploiting the FFT must be performed. Therefore, it is important to properly handle the frontiers of our images and avoid both implicit FFT frontier periodizations and the influence of unobserved image values integrated by the filter extension. Inspired by the work of \cite{6502713}, we implement their \emph{\textbf{unknown} boundary conditions}, where instead of expanding the observation using zero-padding as in \cite{Anconelli2006,2012A&A...539A.133P}, the boundaries are considered unknown in the convolution process and then, by properly selecting the pixels inside the known boundaries, the observed image is obtained. In a nutshell, the unknown boundary conditions method proceeds by considering every convolution in a bigger space obtained by expanding the original one by the radius of the filter in all directions, \ie by a border of width $b$. Later, the optimization methods can freely set the values in this border. We only impose that the correct convolution values are those selected in the former domain.

Mathematically, we consider a larger image $\bar{\bs x}_j \in \Rbb^{M}$ and a larger filter $\bar{\bs h} \in \Rbb^M$, with $M = N + 2b$. The observation is obtained by selecting the pixels inside the boundaries of the convolved image using a selection operator $\bs S_N \in \Rbb^{N \times M}$. The problem in (\ref{eq:discrete_model_vector}) can be rewritten as:
\begin{equation}
  \bs Y = \bs S_N \ \bs \Phi (\bar{\bs h}) \ \bar{\bs X} + \bs N.
\end{equation}

The regularized blind deconvolution problem in (\ref{eq:BID2}) transforms as follows:
\begin{eqnarray}
\label{eq:BID_selection}
	\textrm{min}_{\tilde{\bs X}, \tilde{\bs h}} & \ \rho \ \| \ \bs S_\Theta \ \bs \Psi^\mathtt{T} \ \tilde{\bs X} \ \|_1 + \textstyle\frac{1}{2} \| \ \bs Z - \ \bs S_N \ \bs \Phi (\tilde{\bs h}) \ \tilde{\bs X} \ \|^2_F \\ \nonumber
		\textrm{s.t.} & \ (\tilde{\bs x}_j)_i = 0 \ \textrm{if} \ i \in \Omega_j; \ (\tilde{\bs x}_j)_i \geq 0 \ \textrm{otherwise} \ \ \ \\ \nonumber
		& \ \tilde{\bs h} \in \cl PS; \ \textrm{supp} \ \tilde{\bs h} \in \Gamma  \quad \quad \quad \quad \quad \quad \quad \ \ \
\end{eqnarray}
with $\tilde{\bs X} \in \Rbb^{M \times P}$ and $\tilde{\bs h} \in \Rbb^M$. The actual image $\bs X^*$ and filter $\bs h^*$ are obtained by applying the operator $\bs S_N$ to both the estimated image $\bar{\bs X}^*$ and filter $\bar{\bs h}^*$, respectively, \ie $\bs X^* = \bs S_N \ \bar{\bs X}^*$, $\bs h^* = \bs S_N \ \bar{\bs h}^*$.

The unknown boundaries are also taken into consideration in the set $\Theta$ by excluding from this set the wavelet coefficients that are affected by the image boundaries.

\def\aj{AJ}%
\def\actaa{Acta Astron.}%
\def\araa{ARA\&A}%
\def\apj{ApJ}%
\def\apjl{ApJ}%
\def\apjs{ApJS}%
\def\ao{Appl.~Opt.}%
\def\apss{Ap\&SS}%
\def\aap{A\&A}%
\def\aapr{A\&A~Rev.}%
\def\aaps{A\&AS}%
\def\azh{AZh}%
\def\baas{BAAS}%
\def\bac{Bull. astr. Inst. Czechosl.}%
\def\caa{Chinese Astron. Astrophys.}%
\def\cjaa{Chinese J. Astron. Astrophys.}%
\def\icarus{Icarus}%
\def\jcap{J. Cosmology Astropart. Phys.}%
\def\jrasc{JRASC}%
\def\mnras{MNRAS}%
\def\memras{MmRAS}%
\def\na{New A}%
\def\nar{New A Rev.}%
\def\pasa{PASA}%
\def\pra{Phys.~Rev.~A}%
\def\prb{Phys.~Rev.~B}%
\def\prc{Phys.~Rev.~C}%
\def\prd{Phys.~Rev.~D}%
\def\pre{Phys.~Rev.~E}%
\def\prl{Phys.~Rev.~Lett.}%
\def\pasp{PASP}%
\def\pasj{PASJ}%
\def\qjras{QJRAS}%
\def\rmxaa{Rev. Mexicana Astron. Astrofis.}%
\def\skytel{S\&T}%
\def\solphys{Sol.~Phys.}%
\def\sovast{Soviet~Ast.}%
\def\ssr{Space~Sci.~Rev.}%
\def\zap{ZAp}%
\def\nat{Nature}%
\def\iaucirc{IAU~Circ.}%
\def\aplett{Astrophys.~Lett.}%
\def\apspr{Astrophys.~Space~Phys.~Res.}%
\def\bain{Bull.~Astron.~Inst.~Netherlands}%
\def\fcp{Fund.~Cosmic~Phys.}%
\def\gca{Geochim.~Cosmochim.~Acta}%
\def\grl{Geophys.~Res.~Lett.}%
\def\jcp{J.~Chem.~Phys.}%
\def\jgr{J.~Geophys.~Res.}%
\def\jqsrt{J.~Quant.~Spec.~Radiat.~Transf.}%
\def\memsai{Mem.~Soc.~Astron.~Italiana}%
\def\nphysa{Nucl.~Phys.~A}%
\def\physrep{Phys.~Rep.}%
\def\physscr{Phys.~Scr}%
\def\planss{Planet.~Space~Sci.}%
\def\procspie{Proc.~SPIE}%

\newcommand{\etalchar}[1]{$^{#1}$}

\end{document}